\documentclass{article}

\usepackage{microtype}
\usepackage{graphicx}
\usepackage{subfigure}
\usepackage{booktabs} 

\usepackage{hyperref}

\usepackage[accepted]{icml2025}

\usepackage{amsmath}
\usepackage{amssymb}
\usepackage{mathtools}
\usepackage{amsthm}
\usepackage{mathabx}
\usepackage{bbm}
\usepackage{mathrsfs}

\usepackage{relsize}
\usepackage{exscale}

\usepackage{algorithm}
\usepackage{algorithmicx} 
\usepackage{algpseudocode} 

\usepackage[capitalize,noabbrev]{cleveref}

\theoremstyle{plain}
\newtheorem{theorem}{Theorem}[section]
\newtheorem{proposition}[theorem]{Proposition}
\newtheorem{lemma}[theorem]{Lemma}
\newtheorem{corollary}[theorem]{Corollary}
\theoremstyle{definition}
\newtheorem{definition}[theorem]{Definition}
\newtheorem{assumption}[theorem]{Assumption}
\theoremstyle{remark}
\newtheorem{remark}[theorem]{Remark}

\usepackage[textsize=tiny]{todonotes}

\usepackage{enumerate}
\usepackage{enumitem}

\usepackage{caption}
\usepackage{graphicx}
\usepackage{subcaption}
\usepackage{multirow}

\usepackage{makecell}

\usepackage{graphicx}
\usepackage{mwe}

\usepackage{tikz}
\usetikzlibrary{shapes,arrows,positioning,fit,decorations.pathreplacing,calc}

\usepackage{blkarray}
\usepackage{bm}
\usepackage{todonotes}

\usepackage{abbreviation_definitions}

\usepackage[toc,page,header]{appendix}
\usepackage{minitoc}

\icmltitlerunning{Improved Online Confidence Bounds for Multinomial Logistic Bandits}

\doparttoc 
\faketableofcontents 

\begin{document}

\twocolumn[
\icmltitle{Improved Online Confidence Bounds for Multinomial Logistic Bandits}




\begin{icmlauthorlist}
\icmlauthor{Joongkyu Lee}{sch}
\icmlauthor{Min-hwan Oh}{sch}
\end{icmlauthorlist}

\icmlaffiliation{sch}{Seoul National University, Seoul, Korea}

\icmlcorrespondingauthor{Min-hwan Oh}{minoh@snu.ac.kr}

\icmlkeywords{Machine Learning, ICML}

\vskip 0.3in
]



\printAffiliationsAndNotice{}  

\begin{abstract}
In this paper, we propose an improved online confidence bound for multinomial logistic (MNL) models and apply this result to MNL bandits, achieving variance-dependent optimal regret. 
Recently, \citet{lee2024nearly} established an online confidence bound for MNL models and achieved nearly minimax-optimal regret in MNL bandits. 
However, their results still depend on the norm-boundedness of the unknown parameter $B$ and the maximum size of possible outcomes $K$.
To address this, we first derive an online confidence bound of $\mathcal{O}\left(\sqrt{d \log t} + B \sqrt{d} \right)$, which is a significant improvement over the previous bound of $\mathcal{O} (B \sqrt{d} \log t \log K )$ \citep{lee2024nearly}.
This is mainly achieved by establishing tighter self-concordant properties of the MNL loss and applying Ville’s inequality to bound the estimation error. 
Using this new online confidence bound, we propose a constant-time algorithm, \AlgName{}, which achieves a variance-dependent regret bound of $\mathcal{O} \Big( d \log T 
\sqrt{ \sum_{t=1}^T \sigma_t^2 
} \Big) $ for sufficiently large $T$, where $\sigma_t^2$ denotes the variance of the rewards at round $t$, $d$ is the dimension of the contexts, and $T$ is the total number of rounds.
Furthermore, we introduce a Maximum Likelihood Estimation (MLE)-based algorithm, \AlgNameMLE{}, which achieves an anytime $\operatorname{poly}(B)$-free regret of $\mathcal{O} \Big( d \log (BT) 
\sqrt{ \sum_{t=1}^T \sigma_t^2 
} \Big) $.
\end{abstract}

\section{Introduction}
\label{sec:Introduction}
The multinomial logistic (MNL) bandit framework~\citep{rusmevichientong2010dynamic, saure2013optimal, agrawal2017thompson, agrawal2019mnl, oh2019thompson, oh2021multinomial, perivier2022dynamic, agrawal2023tractable,  lee2024nearly} provides a principled approach to tackling sequential assortment selection problems.
At every round $t$, an agent offers an assortment of at most $K$ items among total $N$ items and receives feedback \textit{only} for the chosen decisions.
The user choice probability follows an MNL model~\citep{mcfadden1977modelling}.
This framework is widely deployed in industry, with applications ranging from news recommendation systems to online retail, where assortment selections are optimized based on user-choice feedback from the offered options.
In such applications, the agent often has access to item features and, potentially, contextual information about the user. 
This setup is referred to as the \textit{contextual} MNL bandit problem~\citep{agrawal2019mnl, agrawal2017thompson, ou2018multinomial, chen2020dynamic, oh2019thompson, oh2021multinomial, perivier2022dynamic, agrawal2023tractable, lee2024nearly}.

Recently, in contextual MNL bandits,~\citet{lee2024nearly} proposed a constant-time algorithm and 
obtained a regret of $\BigO(B^{3/2} d \log K (\log T)^{3/2}  \sqrt{T} )$.
Although this result is nearly minimax optimal when ignoring $B$ and logarithmic terms, it still depends on the maximum assortment size $K$ and the norm-boundedness of the parameter $B$.
Intuitively, a larger $K$ may provide more information~\citep{lee2024nearly}, suggesting that the regret should not scale with any factor involving $K$. 
Moreover, while $\operatorname{poly}(B)$-free regret bound has been established for generalized linear model (GLM) bandits~\citep{lee2024unified} using the MLE, it remains unclear whether such a bound can be obtained while maintaining a constant per-round computational cost.

Our main goal is to design a constant-time algorithm that achieves improved regret with respect to $\operatorname{poly}(B)$ and $K$.
The main challenge in achieving such regret lies in deriving a tight confidence bound.
Currently, the best-known online confidence bound is $\BigO(B \sqrt{d} \log t \log K)$~\citep{lee2024nearly}, which  explicitly depends on both $B$ and $\log K$.
This dependency poses a significant bottleneck for obtaining improved regret.
Furthermore, to the best of our knowledge, there is no variance-dependent regret in contextual MNL bandits.
Hence, the following research questions arise:
\begin{itemize}
    \item \textit{Can we derive a \underline{$B, K$-improved} confidence bound for \underline{online parameter estimation} in MNL models?}

    \item \textit{Can we design a \underline{constant-time} algorithm that achieves \underline{$B, K$-improved (or free)} and \underline{variance-dependent} regret in contextual MNL bandits?}
\end{itemize}
In the first part of our main results (Section~\ref{subsec:K-free_online}), we construct a $K$-free online confidence bound with improved dependence on $B$, which depends on a \textit{update condition parameter} $\alpha$.
This significantly improves upon previous results in online parameter estimation in MNL models~\citep{zhang2024online, lee2024nearly}.
To achieve this, we first establish self-concordant-like properties with respect to the $\ell_\infty$-norm (instead of the traditional $\ell_2$-norm) (Propositions~\ref{prop:self_concordant} to \ref{prop:hessian_usedful}).
This improvement enhances the existing self-concordant properties of MNL models~\citep{tran2015composite}, which is of independent interest.
Then, unlike \citet{zhang2024online, lee2024nearly}, we apply Ville’s inequality~\citep{ville1939etude} to bound estimation errors, thereby avoiding the need for a smoothing technique~\citep{foster2018logistic}, which would otherwise lead to a $\BigO (\sqrt{\log t \log K} )$ looser confidence bound.

In the second part (Section~\ref{subsec:alg_online}), we propose a constant-time algorithm, called \AlgName{}, that achieves $B$-improved, $K$-free, and variance-dependent regret.
This algorithm updates the parameter only within a specific space constructed during the \textit{adaptive warm-up} rounds.
With high probability, this space contains the true parameter $ \wb^\star $, while also shrinking sufficiently relative to the current feature set $ \Xcal_t $.
This is the key to keeping the update condition parameter $\alpha$ of the online confidence bound small (or constant).
Note that since the parameter is updated in a fully online manner, the computational cost per round of \AlgName{} remains constant throughout all rounds.
Furthermore, we introduce a novel regret decomposition, which ultimately allows us to achieve a variance-dependent regret bound of $\mathcal{O}\Big( \left(d \log T + Bd \sqrt{ \log T} \right) \sqrt{\sum_{t=1}^T \sigma_t^2 }\Big)$, where $\sigma_t^2$ denotes the variance of the rewards at round $t$.

In the final part (Section~\ref{subsec:regret_MLE}), as an independent contribution and inspired by \citet{lee2024unified}, we propose an MLE-based algorithm, \AlgNameMLE{}, that leverages an MLE confidence bound and achieves completely $\operatorname{poly}(B), K$-free regret with only $\log B$ dependence.
However, note that the per-round computational cost of \AlgNameMLE{} increases linearly with $t$ due to the use of the MLE, whereas the per-round computational cost of \AlgName{} remains constant.

Our main contributions are summarized as follows:
\begin{itemize}
    \item 
    \textbf{Sharper online confidence bound for the MNL models}:
    We first establish a confidence bound for online parameter updates in MNL models, which depends on the update condition parameter $\alpha$ (defined later).
    In Theorem~\ref{thm:online_confidence_set}, when the parameter is updated over the entire space $\BB^d(B)$, as is common in prior works~\citep{faury2022jointly, zhang2024online, lee2024nearly}, we achieve a confidence bound of $\BigO(B \sqrt{d \log t} + B^{3/2} \sqrt{d} + B^2)$, significantly improving upon the previous bound of $\BigO(B \sqrt{d} \log t \log K + B^{3/2} \sqrt{d \log K})$~\citep{lee2024nearly}.
    More importantly, when the parameter is updated within a specific space where the update condition parameter $\alpha$ is bounded by a constant, we achieve a confidence bound of $\BigO(\sqrt{d \log t} + B\sqrt{d})$, which is completely independent of $\operatorname{poly}(B)$ and $K$.
    \item 
    \textbf{New $B$-improved, $K$-free, variance-dependent regret bound}:
    To apply our new online confidence bound to MNL bandits and achieve a tighter regret in terms of $\operatorname{poly}(B)$ and $K$, we propose an algorithm called \AlgName{}. 
    In addition, through a novel regret decomposition, we derive a variance-dependent optimal regret of $\BigO\Big( \left(d \log T + B d \sqrt{ \log T} \right) \sqrt{\sum_{t=1}^T \sigma_t^2 }\Big)$, where $\sigma_t^2 \leq 1$ represents the variance of the rewards at round $t$.
    For sufficiently large $T$, we obtain a $\BigOTilde\Big( d \sqrt{\sum_{t=1}^T \sigma_t^2 }\Big)$ regret.
    To the best of our knowledge, this is the first $B,K$-free and variance-dependent optimal regret bound in contextual MNL bandits.
    \item \textbf{Completely  $\operatorname{\textbf{poly}}(B),K$-free confidence and regret bound using MLE}: 
    We propose an MLE-based algorithm, \AlgNameMLE{}, which achieves $\operatorname{poly}(B), K$-free variance-dependent optimal regret of $\BigO(d \log (B T) 
            \sqrt{ \sum_{t=1}^T \sigma_t^2 
            })$ by leveraging a $B$-free MLE confidence bound.
\end{itemize}

\begin{table*}[t]
\caption{Comparisons of regret bounds in recent works on contextual logistic and MNL bandits with $T$ rounds, the maximum size of assortments $K$, $d$-dimensional feature vectors, the norm-boundedness of the unknown parameter $B$, problem-dependent constants $1/\kappa = \BigO(K^2 e^{3B})$ and $\kappa_t^\star \!:=\! \sum_{i \in S_t^\star} p_t(i | S_t^\star, \mathbf{w}^\star)p_t(0 | S_t^\star, \mathbf{w}^\star) \leq 1$,
and the variance of the rewards $\sigma_t^2 \leq 1$ at round $t$ (formally defined in~\eqref{eq:sigma_def}).
For the computational cost (abbreviated as ``Comput.''), we consider only the dependence on $t$.
The term ``Intractable'' refers to computational runtimes that are non-polynomial.
}
\resizebox{\textwidth}{!}{
\centering
\begin{tabular}{cllcccc}
\toprule
\multicolumn{1}{l}{}         &         Algorithm        & Regret & Rewards &  Comput. per Round \\
\midrule
\multirow{7}{*}{\begin{tabular}[c]{@{}c@{}} Logistic \\ Bandits\end{tabular}}
                             & \makecell[l]{\citet{abeille2021instance}\\\textcolor{gray}{\footnotesize{(\texttt{OFULog})}} }              & $\BigO \Big( \textcolor{red}{B^{3/2}} d  \log T  \sqrt{ \sum_{t=1}^T \kappa^\star_t } \Big)$    &  Uniform  &   \textcolor{red}{Intractable}   \\
                             & \makecell[l]{\citet{abeille2021instance} \\ \textcolor{gray}{\footnotesize{(\texttt{OFULog-r)}}} }         & $\BigO \Big( \textcolor{red}{B^{5/2}} d  \log T  \sqrt{ \sum_{t=1}^T \kappa^\star_t } \Big)$    &  Uniform  &   \textcolor{red}{$\BigO( t )$}  \\
                             & \makecell[l]{\citet{faury2022jointly} \\ \textcolor{gray}{\footnotesize{(\texttt{ada-OFU-ECOLog)}}} }          & $\BigO \Big( \textcolor{red}{B} d  \log T  \sqrt{ \sum_{t=1}^T \kappa^\star_t } \Big)$    &  Uniform  &   \textcolor{red}{$\BigO(\log t )$}   \\
                             & \makecell[l]{\citet{lee2024unified} \\ \textcolor{gray}{\footnotesize{(\texttt{OFUGLB)}}} }               & $\BigO \Big(  d  \log ( \textcolor{blue}{B} T)  \sqrt{ \sum_{t=1}^T \kappa^\star_t } \Big)$    &  Uniform  &   \textcolor{red}{$\BigO( t )$}   \\
\midrule
\multirow{10}{*}{\begin{tabular}[c]{@{}c@{}} MNL \\ Bandits\end{tabular}}
                             & \makecell[l]{\citet{chen2020dynamic} \\\textcolor{gray}{\footnotesize{(\texttt{MLE-UCB})}} }                              & $\BigO \Big( \textcolor{red}{B} d \log ( \textcolor{red}{K} T) \sqrt{T} \Big)$    &  Uniform/Non-Uniform  &   \textcolor{red}{Intractable}        \\
                             & \makecell[l]{\citet{oh2021multinomial} \\\textcolor{gray}{\footnotesize{(\texttt{UCB-MNL})}} }                          & $\BigO \Big( \textcolor{red}{\frac{1}{\kappa}} d \log T \sqrt{T} \Big) = \BigO \Big(\textcolor{red}{K^2 e^{B} } d \log T \sqrt{T} \Big)$ &   Uniform/Non-Uniform  &     \textcolor{red}{$\BigO( t )$} \\
                             & \makecell[l]{\citet{perivier2022dynamic} \\\textcolor{gray}{\footnotesize{(\texttt{OFU-MNL})}} }                              & $\BigO \Big( \textcolor{red}{ B K} d \log ( \textcolor{red}{K} T) \sqrt{ \sum_{t=1}^T \kappa^\star_t } \Big)$    &    Uniform    &    \textcolor{red}{Intractable}        \\
                             & \makecell[l]{\citet{lee2024nearly} \\\textcolor{gray}{\footnotesize{(\texttt{OFU-MNL+})}} }      & $\BigO \Big( \textcolor{red}{B^{3/2}} d \log \textcolor{red}{K} (\log \textcolor{red}{T})^{3/2}  \sqrt{T} \Big)$    &    Uniform/Non-Uniform    &    \textcolor{blue}{$\BigO(1)$}          \\
                             & \makecell[l]{\textbf{This work} \\\textcolor{gray}{\footnotesize{(\AlgName{}, Theorem~\ref{thm:regret_main})}  } }                  &  $\BigO \Big( \left(d  \log T + \textcolor{blue}{B}d \sqrt{ \log T} \right) \sqrt{ \sum_{t=1}^T \sigma_t^2 } \Big)$     &      Uniform/Non-Uniform  &  \textcolor{blue}{$\BigO(1)$}        \\
                             & \makecell[l]{\textbf{This work} \\\textcolor{gray}{\footnotesize{(\AlgNameMLE{}, Theorem~\ref{thm:MLE})}} }  &  $\BigO \Big( d  \log ( \textcolor{blue}{B} T) \sqrt{ \sum_{t=1 }^T \sigma_t^2 } \Big)$   &  Uniform/Non-Uniform  &     $\textcolor{red}{\BigO(t)}$    \\
\bottomrule
\end{tabular}
}
\label{tab:regrets}
\end{table*}
\section{Related Work}
\label{sec:Related}
\textbf{Logistic bandits.}
The logistic bandit problem~\citep{dong2019performance, faury2020improved, abeille2021instance, faury2022jointly, lee2024improved, lee2024unified} is a special case of the MNL bandit problem. 
In this setting, the agent offers only a single item (i.e., $K=1$) and receives $0$-$1$ binary feedback, restricting the problem to the uniform rewards setting.
As summarized in Table~\ref{tab:regrets}, recent works have successfully eliminated the harmful dependency on $1/\kappa$ (which can be exponentially large) in the leading term, achieving instance-dependent regret (i.e., 
 $\kappa^\star_t$-dependent regret). 
However, most of these approaches still suffer from an unnecessary dependency on the norm-boundedness of the unknown parameter, $\operatorname{poly}(B)$.
While a recent work by \citet{lee2024unified} successfully eliminated the $\operatorname{poly}(B)$ factors, their approach incurs a per-round computational cost that grows linearly with $t$.
Thus, the question of whether it is possible to design a $B$-free, computationally efficient algorithm remains open.

\textbf{MNL bandits.}
The MNL bandits~\citep{agrawal2019mnl, agrawal2017thompson, ou2018multinomial, chen2020dynamic, oh2019thompson, oh2021multinomial, perivier2022dynamic, agrawal2023tractable, lee2024nearly} address more sophisticated problems compared to logistic bandits, as they involve selecting a set of items (thus highlighting their combinatorial nature) and consider non-uniform rewards rather than binary feedback.
Recently, \citet{lee2024nearly} made significant progress by resolving the long-standing open problem of establishing the minimax optimal regret (ignoring factors of $B$ and logarithmic terms) with computational efficiency. 
However, as shown in Table~\ref{tab:regrets}, all existing regret bounds increase with $B$ and $K$. 
Furthermore, the tightest regret bound by \citet{lee2024nearly} includes an additional $(\log T)^{1/2}$ term, arising from a loose confidence bound.
To address these limitations, in this paper, we construct the sharper online confidence bound to date and, leveraging this, achieve (asymptotically) $B,K$-free regret while maintaining computational efficiency.

\textbf{RL with MNL models.}
There has been growing interest in incorporating MNL models into reinforcement learning (RL).
One line of work extends MNL bandits to the RL setting. 
Recently,~\citet{lee2025combinatorial} proposed a new framework, called \textit{combinatorial RL with preference feedback}, in which the agent selects a subset of items in each round to maximize long-term cumulative reward based on MNL-modeled preferences,
and established the minimax-optimal regret bound in linear MDPs~\citep{jin2020provably}.

Another direction focuses on RL with MNL-based transition models. 
\citet{hwang2022model} introduced \textit{MNL-MDPs}, a class of MDPs where the transition probabilities are parameterized by an MNL model.
Building on this, \citet{cho2024randomized} and \citet{li2024provably} concurrently improved the dependency on $1/\kappa = \BigO(K^2 e^{3B})$ in their regret bounds.
\citet{park2024infinite} further extended this direction to the infinite-horizon setting.

\section{Preliminaries}
\label{sec:preliminaries}

\textbf{Notations.}
For a positive integer $n$, we define $[n]$ as the set $\{1, 2, \ldots, n \}$. 
The $\ell_2$- and $\ell_\infty$-norm of a vector $x$ is denoted by $\|x\|_2$ and $\|x \|_{\infty}$, respectively.
For a positive semi-definite matrix $A$ and a vector $x$, we use $\|x \|_A$ to represent $\sqrt{x^\top A x}$.
For any two symmetric matrices $A$ and $B$ of the same dimensions, $A \succeq B$ indicates that $A-B$  is a positive semi-definite matrix.
Finally, we define $\mathcal{S}$ as the set of candidate assortments with a size constraint of at most $K$, i.e., $ \mathcal{S} = \{S \subseteq [N]: |S| \leq K \}$.

\subsection{Problem Setting}
\label{subsec:problem_setting}

We consider the contextual MNL bandit problem, where an agent selects assortments (sets of items) and receives feedback based on user choices.
Specifically, at each round $t$, the agent receives a feature vector $x_{ti} \in \RR^d$ and a reward $r_{ti}$ for every item $i \in [N]$.
Note that the feature set  $\Xcal_t := \{x_{ti}\}_{i=1}^N$ and rewards $\{r_{ti}\}_{i=1}^N$ can be \textit{arbitrarily} chosen by an adversary.
The agent then offers an assortment $S_t = \{ i_1, \dots, i_{l} \} \in \mathcal{S}$, where $l \leq K$.
After presenting the assortment, the agent observes the user's purchase decision $c_t \in S_t\cup \{0\}$, where $\{0\}$ represents the ``outside option'', indicating that the user did not choose any item from $S_t$.
The user choices are modeled using the Multinomial Logistic (MNL) framework~\citep{mcfadden1977modelling},  where the probability of selecting an item $i \in S_t \cup \{ 0 \}$ is defined as:
\begin{align*}
    \label{eq:mnl_model}
    p_t(i | S_t, \wb^\star) 
    &:= \frac{\exp(x_{t i}^\top  \wb^\star )}{ 1 \!+\!\sum_{j \in S_t }\exp( x_{tj}^\top \wb^\star )},
\end{align*}
where $\wb^\star \in \RR^d$ is an \textit{unknown} parameter and $x_{t0} = \mathbf{0}$.

The choice response for each item $i \in S_t \cup \{ 0\}$ is defined as $y_{ti} := \mathbbm{1}(c_t = i) \in \{0,1\}$.
Hence, the choice feedback vector $\yb_t := (y_{t0}, y_{ti_1}, \dots y_{ti_{l}}) $ is sampled from the multinomial (MNL) distribution 
$ \yb_{t} \sim \operatorname{MNL} \{ 1, ( p_t(0 | S_t, \wb^\star), \dots, p_t(i_l | S_t, \wb^\star)  )\}$,
where the parameter $1$ indicates that $\yb_t$ is a single-trial sample, meaning $y_{t0} + \sum_{k=1}^l y_{ti_k} = 1$. 
Then, the expected revenue of an assortment $S$ is defined as:
\begin{align*}
    R_{t}(S, \wb^\star):=  \sum_{i \in S} p_t(i | S, \wb^\star) r_{ti} 
    = \frac{ \sum_{i \in S }\exp(x_{ti}^\top  \wb^\star )r_{ti} }{ 1 \!+\!\sum_{j \in S }\exp( x_{tj}^\top \wb^\star )}.
\end{align*}
We denote $S_{t}^\star$ as the optimal assortment at time $t$, i.e.,  $S_{t}^\star := \argmax_{S \in \mathcal{S}}  R_{t}(S, \wb^\star)$.
The goal of the agent is to minimize the cumulative regret over the $T$ rounds:
\begin{align*}
    \Regret (\wb^\star )
    := \sum_{t=1}^T   R_{t}(S_{t}^\star, \wb^\star) 
    - R_{t}(S_{t}, \wb^\star). 
\end{align*}
When $K=1$ and $r_{t1}=1$, the MNL bandit reduces to the binary logistic bandit with $R_t(S=\{x\}, \wb^\star) = \sigma\left( x^\top \wb^\star \right) = 1/ (1 + \exp (-x^\top \wb^\star) )$, where $\sigma(\cdot)$ is the sigmoid function.

We will work under the standard boundedness assumption.
\begin{assumption}[Bounded assumption] 
\label{assum:bounded_assumption}
We assume that, for all $t \geq 1$, $i \in [N]$, $\| x_{ti} \|_2 \leq 1$ and $ r_{ti}  \in [0,1]$.
There exists a \textit{known} constant such that $\| \wb^\star \|_2  \leq \BoundParam$, 
\end{assumption}
Following the previous contextual MNL bandit literature~\citep{ oh2021multinomial, perivier2022dynamic, zhang2024online, lee2024nearly}, we  introduce the problem-dependent constant:
\begin{definition}  
\label{def:kappa}
Let $\mathcal{W} = \{ \wb \in \RR^d \mid \| \wb \|_2 \leq \BoundParam \}$.
There exists $\kappa >0$ such that, for any $i \in S$, $S \in \mathcal{S}$, and $t \in [T]$, we have $\min_{\wb \in \mathcal{W}} p_t(i | S, \wb) p_t(0 | S, \wb) \geq \kappa$.
\end{definition}
A small $\kappa$ signifies a greater deviation from the linear model. 
Notably,  $1/\kappa$ can be exponentially large, growing on the order of $\BigO(K^2 e^{3B})$.
Therefore, it is crucial to ensure that our regret bound does not depend on  $1/\kappa$.

\section{Main Results}
\label{sec:main}
%
%
\subsection{Sharper Online Confidence Bound for MNL Model}
\label{subsec:K-free_online}
Instead of performing Maximum Likelihood Estimation (MLE) as done in previous studies~\citep{chen2020dynamic,oh2021multinomial, perivier2022dynamic}, we follow the approach of~\citet{zhang2024online, lee2024nearly}  and adopt the online mirror descent (OMD) algorithm for parameter estimation. 
To begin, we define the multinomial logistic loss function for round $t$ as:
\begin{equation}
    \label{eq:loss}
    \ell_t(\wb) := - \sum_{i \in S_t} y_{ti} \log p_t(i | S_t, \wb).    
\end{equation}
In this paper, we present a \textit{general} description of online parameter estimation. 
We consider a (possibly) \textit{time-varying} compact convex search space
$\Wcal_t \subseteq \RR^d$
and allow for \textit{occasional updates} to the parameter rather than requiring updates at every round.
We denote $\Tcal \subseteq [T]$ as all the update rounds.
At the update round $t \in \Tcal$, the true parameter $\wb^\star$ is estimated as follows:
\begin{align*}
    \wb_t' &= \argmin_{\wb \in \Wcal_t} \| \wb - \wb_t \|_{H_t}, 
    \tag{projection onto $\Wcal_t$}
    \\
    \wb_{t+1} 
    &= \argmin_{\wb \in \Wcal_t }   \, \langle \nabla \ell_t (\wb_{ t }' ), \wb \rangle
    + \frac{1}{2 \eta} \| \wb - \wb_t' \|_{\tilde{H}_{t}}^2,
    \numberthis \label{eq:online_update}
\end{align*} 
where $\eta > 0$ is the step-size parameter, and $\Wcal_t \subseteq \RR^d$ is the compact convex set, which will be specified later.
The matrix $\tilde{H}_{t}$ is defined as $\tilde{H}_{t} := H_{t} + \eta \nabla^2 \ell_{t}(\wb_t')$,
where
$H_{t} := \lambda \Ib_d + \sum_{s \in \Tcal \setminus \{t, \dots, T \} } \nabla^2 \ell_s (\wb_{s+1})$
with $\lambda >0$.

If no update is performed,  $\wb_t$, $\tilde{H}_{t}$ and $H_{t} $ remain unchanged.
Formally, let $t' \in \Tcal$ denote the last update round prior to $t$ (i.e., $t' < t$).
Then, we have
$\wb_{t'+1} = \dots = \wb_t $, $H_{t'+1} = \dots = H_{t}$,
and $\tilde{H}_{t'+1} = \dots = \tilde{H}_{t} $.

In the optimization problem~\eqref{eq:online_update}, we first solve the unconstrained optimization problem in closed form, obtaining $\wb^{\prime}_{t+1}$. 
Then, we 
project  $\wb^{\prime}_{t+1}$ back into the feasible set.
\begin{align*}
    \wb^{\prime}_{t+1} &= \wb_t' - \eta \tilde{H}_t^{-1} \nabla \ell_t(\wb_t'),
    \\
    \wb_{t+1} &= \argmin_{\wb \in \Wcal_t } \| \wb - \wb^{\prime}_{t+1} \|_{\tilde{H}_t}, 
    \numberthis
    \label{eq:PGD}
    .
\end{align*}

This estimator is efficient in both computation and storage. 
\begin{remark} [Computational cost] \label{remark:compute_online}
    For a general convex set $\Wcal_t$, the projection optimization problem (e.g., Equation~\eqref{eq:PGD}) can be solved up to  $\epsilon >0 $  accuracy using
    the Projected Gradient Descent algorithm (e.g., Algorithm 2 in~\citep{hazan2016introduction}), requiring  computational cost of $\BigO(K d^3 \log (1/\epsilon) )$.
    As a special case, if $\Wcal_t$ is an ellipsoid,  the optimization problem can be solved in a single projection step (via a closed‐form projection), which needs only $\mathcal{O}(Kd^3)$ cost.
\end{remark} 
In terms of storage, the estimator avoids retaining all historical data, as $\tilde{H}_t$, and $H_t$  can be updated incrementally, requiring only $\mathcal{O}(d^2)$ storage.

Our first main contribution is the development of an improved online confidence bound for MNL bandits, which depends on the update condition parameter $\alpha$.
The proof is deferred to Appendix~\ref{app_sec:proof_thm_confidence}.
\begin{theorem} [Improved online confidence bound]
\label{thm:online_confidence_set}
    Let $\delta \in (0, 1]$ and $\Tcal \subseteq [T]$ denote the set of update rounds.
    For all $t \in \Tcal$, we assume the following update conditions hold:
    \begin{align*}
        \sup_{ \wb \in \Wcal_t}  
            |x_{ti}^\top (\wb  - \wb^\star) |
             \leq \alpha,
        \quad \forall i  \in S_{t},
    \end{align*}
    where  $\Wcal_t$  is a compact convex set,
    and $\alpha > 0$.
    We set $\eta = (1+ 3\sqrt{2} \alpha)/2$ and 
    $ \lambda = \max \{ 72 \eta \alpha, 144 \eta d, 2 \}$.
    Then, under Assumption~\ref{assum:bounded_assumption}, 
    with probability at least $1- \delta$, we have:
    \begin{align*}
        &\| \wb_t - \wb^\star \|_{H_t} 
        = \BigO\left(
        \alpha
        \sqrt{
            d \log (t/\delta)
            }
            + B \sqrt{\lambda}
        \right).
    \end{align*}
\end{theorem}
\begin{remark} [Condition of Theorem~\ref{thm:online_confidence_set}]
\label{remark:thm_online_condition}
Note that the condition in Theorem~\ref{thm:online_confidence_set} is easy to satisfy and has already been addressed in prior works~\citep{faury2022jointly, zhang2024online, lee2024nearly}.
Specifically, if the parameter is updated at every round (i.e., $\Tcal = [T]$) over the entire parameter space (i.e., $\Wcal_t  = \Wcal$), as is common in previous works~\citep{faury2022jointly, zhang2024online, lee2024nearly}, it follows directly that $\alpha = B$.
\end{remark}
\textbf{Discussion of Theorem~\ref{thm:online_confidence_set}.} 
When the parameter is updated at every round (so $\alpha = 2 B$) and $\lambda$ is set to $\lambda = \Theta( B d + B^2)$, we obtain a completely $K$-free confidence bound of $\BigO(B\sqrt{d \log t} + B^{3/2} \sqrt{d} + B^2)$.
Compared to the recently established confidence bound  $\BigO( B \sqrt{d} \log t \log K  
+ B^{3/2} \sqrt{d \log K })$~\citep{lee2024nearly}, our bound is tighter by a factor of $\sqrt{\log t} \log K$ in the leading term.

More importantly, and perhaps more interestingly, if we can construct $\Wcal_t$ such that $\alpha$ remains small (or constant) and updates occur \textit{only} when this condition is met, we achieve a confidence bound of $\BigO(\sqrt{d \log t} + B \sqrt{d})$.
For sufficiently large $t$, i.e., $t \geq \BigO(e^{B^2})$, this further simplifies to $\BigO(\sqrt{d \log t})$, representing a significant improvement over the previous bound $\BigO( B \sqrt{d} \log t \log K)$~\citep{lee2024nearly}, with no dependence on $B$ or $K$.

\textbf{Proof sketch of Theorem~\ref{thm:online_confidence_set}.}
We provide a proof sketch and highlight the technical novelties of Theorem~\ref{thm:online_confidence_set}.

Following the previous works~\citep{zhang2024online, lee2024nearly}, we first bound the estimation error between $\wb_{t+1}$ and $\wb^\star$ as follows:
\begin{align*}
    \| \wb_{t+1} - \wb^\star \|_{H_{t+1}}^2
    &\!\lesssim 
    \eta \!\!\!
        \sum_{s \in \Tcal_{t+ 1}} 
        \!\!\!
        \left(\ell_{s}(\wb^\star)
        - \ell_{s}(\wb_{s+1})
        \right)
    \!+\! B^2 \lambda
    ,
    \numberthis
    \label{eq:proof_sketch_w_gap}
\end{align*}
where $\Tcal_{t+1} \subseteq \Tcal$ is the set of update rounds prior to $t+1$.

\textbf{1) $B, K$-independent step size $\eta$.}  In~\citet{zhang2024online, lee2024nearly}, $\eta$ is set as 
$\eta \simeq \log K + B$, based on Lemma 4 from~\citet{jezequel2021mixability}. 
To eliminate the dependency on  $B$ and $\log K$, 
we establish 
Proposition~\ref{prop:self_concordant}, which shows that the MNL loss is $3\sqrt{2}$-self-concordant with respect to the 
$\ell_{\infty}$-norm (rather than the 
$\ell_2$-norm, as shown in~\citet{tran2015composite}), which may be of independent interest.
This result enables us to set 
$\eta \simeq \alpha$~(Proposition~\ref{lemma:second_order_loss}), making it independent of $B$ and $K$.

\textbf{2) Intermediary term.} 
Inspired by~\citet{zhang2024online, lee2024nearly}, we introduce an intermediary parameter:
\begin{align*}
    \tilde{\zb}_s := \sigmab_s^+ \left( \EE_{\wb \sim P_s} \left[\sigmab_s\left( (x_{sj}^\top \wb)_{j \in S_s} \right) \right] \right),
\end{align*}
where $\sigmab_s$ is the softmax function,  $\sigmab_s^+$ is its pseudo-inverse, and $P_s$ is a multivariate normal distribution with mean $\wb_s'$ and covariance $c H_s^{-1}$ for some $c > 0$.
Then, we decompose the sum of loss gaps appearing in the first term of Equation~\eqref{eq:proof_sketch_w_gap} as follows:
\begin{align*}
    &\sum_{s \in \Tcal_{t+ 1}} 
       \!\! \left(\ell_{s}(\wb^\star)
        - \ell_{s}(\wb_{s+1})
        \right)
    \\
    &= \underbrace{
    \!\!\sum_{s \in \Tcal_{t+ 1}} \!\!\left( \ell_{s}(\wb^\star)
    -  \bar{\ell}_s(\tilde{\zb}_s)\right)
    }_{(a)}
    + \underbrace{
    \!\!\sum_{s \in \Tcal_{t+ 1}} \!\!
    \left( \bar{\ell}_s(\tilde{\zb}_s)
    - \ell_{s}(\wb_{s+1}) \right).
    }_{(b)}
\end{align*}

\textbf{3) Tighter bound for term $(a)$ via Ville's inequality.}
In \citet{zhang2024online, lee2024nearly}, term $(a)$ is bounded using a Bernstein-type inequality. 
However, the intermediary parameter cannot be used directly, as it is generally unbounded~\citep{foster2018logistic}.
To address this, they employ a \textit{smoothed} version of the parameter, but this leads to a bound of $\BigO(\log K (\log t)^2)$ for term~$(a)$, resulting in a significantly looser confidence bound.

In contrast, we apply \textit{Ville's inequality}~\citep{ville1939etude} without resorting to smoothing. 
To do so, we first show that the following quantity forms a supermartingale:
\begin{align*}
   A_t :=
   \exp \bigg( 
        \sum_{s \in \Tcal_{t+ 1}} \!\left( \ell_{s}(\wb^\star)
        -  \bar{\ell}_s(\tilde{\zb}_s)\right)
        \bigg).
\end{align*}
Then, by Ville's inequality, with probability at least $1-\delta$ (set $\delta \approx \frac{1}{t}$), we can bound term $(a)$ as follows:
\begin{align*}
    \sum_{s \in \Tcal_{t+ 1}} \!\!\left( \ell_{s}(\wb^\star)
    -  \bar{\ell}_s(\tilde{\zb}_s)\right)
    \leq  \log \frac{1}{\delta}
    \approx \log t
    ,
    \numberthis \label{eq:proof_sketch_term_a}
\end{align*}
which is an improvement by a factor of $\BigO(\log K \log t)$ compared to $\BigO(\log K (\log t)^2)$~\citep{zhang2024online, lee2024nearly}.

On the other hand, we can bound term~$(b)$ by applying LemmaF.3 of \citet{lee2024improved} (or Lemma14 of \citet{zhang2024online}):
\begin{align*}
    \sum_{s \in \Tcal_{t+ 1}} \!\!
    \left( \bar{\ell}_s(\tilde{\zb}_s)
    - \ell_{s}(\wb_{s+1}) \right)
    &\lesssim 
    \alpha d \log t.
    \numberthis \label{eq:proof_sketch_term_b}
\end{align*}
Combining~\eqref{eq:proof_sketch_w_gap},~\eqref{eq:proof_sketch_term_a}, and~\eqref{eq:proof_sketch_term_b}, we complete the proof. 

\begin{algorithm*}[tb]
   \caption{\AlgName{}}
   \label{alg:main_online}
    \begin{algorithmic}[1]
       \State {\bfseries Input:} 
       failure level $\delta$, 
       confidence radii $\beta_t(\delta)$ and 
       $\zeta_t(\delta)$.
       \State {\bfseries Initialize:} 
       $\WarmupConfidenceSet_1 (\delta) = \Wcal$,
       $H_1 = \lambda \mathbf{I}_d$,
       $\WarmupHessian_1 = \WarmupRegualizer_1  \mathbf{I}_d$,
       $\wb_1, \WarmupParam_1 \in \mathcal{W}$,
       $\eta := 1$,
       $\WarmupStep:= \frac{1}{2} + 3\sqrt{2}B$,
       $\lambda := 144 d$, 
       $\WarmupRegualizer := \max\{ 12\sqrt{2} \WarmupStep B, 144 \WarmupStep d, 2 \}$,
       $\Threshold_t:= 6\sqrt{2} \zeta_t(\delta)$
       .
       \For{round $t=1, \dots, T$}
            \State Observe feature set $\Xcal_t = \{x_{ti}\}_{i=1}^N$ and rewards $\{r_{ti}\}_{i=1}^N$.
            \If{$ \max_{x \in \Xcal_t} \|x \|_{(\WarmupHessian_t)^{-1}}^{2} \geq 1/\Threshold_t^2$}   \Comment{\textit{Adaptive warm-up}}
                \State {Offer $S_t = \{ i_t \}$, where $x_{t i_t} = \argmax_{x \in \Xcal_t} \|x \|_{(\WarmupHessian_t)^{-1}}^2 $,
                and observe $\yb_t$.}  
                \label{eq:alg_warm_assortment}
                \State Update 
                $(\WarmupParam_{t+1}, \WarmupHessian_{t+1})
                \leftarrow \AlgNameOnline{} (\Wcal, \ell_t, \WarmupHessian_t, \WarmupParam_{t}, \WarmupStep)$
                by Algorithm~\ref{alg:online_update}.
                \label{eq:alg_warm_update}
                \State Calculate $\WarmupConfidenceSet_{t+1} (\delta) \leftarrow 
                \left\{
                    \wb \in \RR^d
                    \mid
                    \| \wb - \WarmupParam_{t+1} \|_{\WarmupHessian_{t+1}} 
                    \leq \zeta_{t+1}(\delta)
                \right\}$.
                \label{eq:alg_warm_CS}
                \State 
                Update
                $H_{t+1} \leftarrow H_t$ and
                 $\wb_{t+1} \leftarrow \wb_t$.
                 \label{eq:alg_warm_unchanged}
            \Else  \Comment{\textit{Planning \& Learning}}
                \State Offer $S_t = \argmax_{S \in \mathcal{S}} \tilde{R}_t (S)$ and observe $\yb_t$.
                \label{eq:alg_learning_assortment}
                \State  Update 
                $(\wb_{t+1}, H_{t+1})
                \leftarrow \AlgNameOnline{} (\WarmupConfidenceSet_{t} (\delta), \ell_t, H_t, \wb_t, \eta)$
                by Algorithm~\ref{alg:online_update}.
                \label{eq:alg_learning_update}
                \State 
                Update
                $\WarmupHessian_{t+1} \leftarrow \WarmupHessian_t$,
                $\WarmupParam_{t+1} \leftarrow \WarmupParam_t$,
                and $\WarmupConfidenceSet_{t+1} (\delta) \leftarrow \WarmupConfidenceSet_{t} (\delta) $.
                \label{eq:alg_learning_unchaged}
            \EndIf
       \EndFor
    \end{algorithmic}
\end{algorithm*}
\begin{algorithm}[tb]
   \caption{\AlgNameOnline{}, \textbf{R}estricted \textbf{S}pace \textbf{OMD} }
   \label{alg:online_update}
    \begin{algorithmic}[1]
       \State {\bfseries Input:} convex set $\Wcal_t$,  
       $\ell_t$,
       $H_t$,
       $\wb_t$,
       $\eta$.
            \State\hspace{\algorithmicindent} Update $\tilde{H}_{t} \leftarrow H_{t} + \eta \nabla^2 \ell_{t}(\wb_t)$.
            \State\hspace{\algorithmicindent} Calculate $\wb_{t+1}$ by Equation~\eqref{eq:online_update}.
            \State\hspace{\algorithmicindent} Update $H_{t+1} \leftarrow H_t + \nabla^2 \ell_t(\wb_{t+1})$.
       \State {\bfseries Return:} $\wb_{t+1}$, $H_{t+1}$.
    \end{algorithmic}
\end{algorithm}

\subsection{Online Update with Adaptive Warm-Up}
\label{subsec:alg_online}
In this subsection, we introduce \AlgName{} (Algorithm~\ref{alg:main_online}),  which employs a novel two-phase online update approach leveraging the improved confidence bound from Theorem~\ref{thm:online_confidence_set} to achieve the tightest regret bound in MNL bandits. 
Note that the feature set $\Xcal_t$ can be arbitrarily given at each round $t$, without imposing any distributional assumptions on the exogenous contexts.

\textbf{Intuition.}
Theorem~\ref{thm:online_confidence_set} indicates that if $\alpha$ is constant, a confidence bound of $\BigO(\sqrt{d \log t })$ can be obtained for sufficiently large $t$.
Our primary objective is to design the search space $\Wcal_t$ to ensure that $\alpha$ remains constant in most rounds. 
To achieve this, we enforce the condition by rejecting, \textit{on-the-fly}, any $\Xcal_t$ that might violate the constancy of $\alpha$. Specifically, it is sufficient to verify the following condition:
\begin{align*}
     \max_{x \in \Xcal_t} \|x \|_{(\WarmupHessian_t)^{-1}}^{2} \geq 1/\Threshold_t^2,
     \numberthis \label{eq:criterion}
\end{align*}
where $\WarmupHessian_t := \WarmupRegualizer \Ib_d + \sum_{s \in \WarmupRounds \setminus \{t, \dots, T \} } \nabla^2 \ell_s (\WarmupParam_{s+1})$ is the warm-up version of $H_t$, i.e., the regularized sum of Hessians corresponding to all assortments played during the adaptive warm-up rounds $\WarmupRounds$.
Here, $\Threshold_t$ is a carefully chosen threshold and $\WarmupRegualizer >0$ is a regularization parameter.

\textbf{Online adaptive warm-up.}
At round $t$, given $\Xcal_t$, if for any feature $x \in \Xcal_t$, the quantity $\|x \|_{(\WarmupHessian_t)^{-1}}^{2}$ is greater than or equal to the threshold $1/\Threshold_t^2$  (as specified in Equation~\eqref{eq:criterion}), we do not update our current estimate $\wb_t$. 
Instead, we update a separate \textit{warm-up parameter} $\WarmupParam_t$  to ensure that the condition in~\eqref{eq:criterion} is more likely to hold in the future.

In such cases, we offer only the single item that maximizes $\|x \|_{(\WarmupHessian_t)^{-1}}^{2}$ (Line~\ref{eq:alg_warm_assortment}). 
Subsequently, we update the warm-up parameter  $\WarmupParam_t$ by invoking \underline{\textbf{R}}estricted \underline{\textbf{S}}pace \underline{\textbf{O}}nline \underline{\textbf{M}}irror \underline{\textbf{D}}escent (\AlgNameOnline{},  Algorithm~\ref{alg:online_update}) as a subroutine (Line~\ref{eq:alg_warm_update}). 
Then, we construct the following parameter set (Line~\ref{eq:alg_warm_CS}):
\begin{align*}
    \WarmupConfidenceSet_{t+1} (\delta) = 
                \left\{
                    \wb \in \RR^d
                    \mid
                    \| \wb - \WarmupParam_{t+1} \|_{\WarmupHessian_{t+1}} 
                    \leq \zeta_{t+1}(\delta)
                \right\},
\end{align*}
where $\zeta_{t+1}(\delta) = \BigO(B \sqrt{d \log (t/\delta)} + B^{3/2}\sqrt{d} + B^2)$.
This ellipsoid is then used in the \AlgNameOnline{} procedure during the \textit{Planning \& Learning} rounds (Line~\ref{eq:alg_learning_assortment}-~\ref{eq:alg_learning_unchaged}).
Note that, since the search space is the entire parameter space $\Wcal$, we can set $\alpha = B$ for the condition of Theorem~\ref{thm:online_confidence_set} to obtain the warm-up confidence bound $\zeta_{t+1}(\delta)$.
The quantities $H_t$ and $\wb_t$ remain unchanged during the warm-up rounds (Line~\ref{eq:alg_warm_unchanged}).

\textbf{Parameter update within restricted space $\WarmupConfidenceSet_{t} (\delta)$.}
When the condition in Equation~\eqref{eq:criterion} does not hold, the parameter $\wb_t$ is updated by searching only within $\WarmupConfidenceSet_{t+1} (\delta)$ using \AlgNameOnline{} as a subroutine (Line~\ref{eq:alg_learning_update}). 
In this scenario, $\alpha$ can be set as a constant (with high probability), leading to a confidence bound of $\BigO(\sqrt{d \log t} + B\sqrt{d})$ (by Theorem~\ref{thm:online_confidence_set}).
\begin{corollary} [Informal, $B$-improved \& $K$-free online confidence bound]
\label{cor:CS_main}
    Let $\delta \in (0,1]$ and $\beta_t(\delta) = \BigO \!\left( 
    \sqrt{\!
        d \log (t/\delta)
        }+ B \sqrt{d}
    \right)$.
    Suppose $\wb^\star \in \WarmupConfidenceSet_t(\delta)$ for all $t \geq 1$.
    Define the following confidence set as follows:
    \begin{align*}
        \Ccal_t (\delta)
        &:= \left\{
            \wb \in \RR^d 
            \mid 
            \|  \wb - \wb_t \|_{H_t}
            \leq \beta_t(\delta)
        \right\}.
    \end{align*}
    Then, we have $\textup{Pr}
        \left[ \forall t \geq 1, \wb^\star \in \Ccal_t(\delta)
        \right] \geq 1- \delta$.
\end{corollary}
\textbf{Efficient assortment selection.} 
Given the confidence set in Corollary~\ref{cor:CS_main}, we calculate the optimistic utility $\UCB_{ti}$ as:
\begin{align*}
    \UCB_{ti} := x_{ti}^\top \wb_t + \beta_t(\delta) \| x_{ti} \|_{H_t^{-1}},
    \quad \, \forall i \in [N]
    .
\end{align*}
If the true parameter $\wb^\star$ lies within the confidence set $\Ccal_t(\delta)$, the value $\UCB_{ti}$ serves as an upper bound for $x_{ti}^\top \wb^\star$.
Using $\UCB_{ti}$, we define the optimistic expected revenue for an assortment $S$ as:
\begin{align}
    \tilde{R}_{t}(S) 
    := \frac{\sum_{i \in S} \exp( \UCB_{ti} ) r_{ti} }{1 + \sum_{j \in S} \exp(\UCB_{tj})}, \label{eq:opt_revenue}
\end{align}
where $r_{ti} \in [0,1]$.
We then offer the assortment $S_t$ that maximizes $\tilde{R}_{t}(S)$, i.e., $S_t = \argmax_{S \in \mathcal{S}} \tilde{R}_{t}(S)$ (Line~\ref{eq:alg_learning_assortment}).
The quantities $\WarmupHessian_t$, $\WarmupParam_t$, and $\WarmupConfidenceSet_t(\delta)$ remain unchanged during the \textit{planning \& learning rounds }(Line~\ref{eq:alg_learning_unchaged}).
Note that the optimization problem in~\eqref{eq:opt_revenue} can be efficiently solved in polynomial time, $\mathcal{O}(\text{poly}(N))$, independent of $t$~\citep{rusmevichientong2010dynamic, davis2014assortment}.
%
%

\textbf{Variance-dependent optimal regret.}
We establish a variance-dependent optimal regret bound through a novel regret decomposition. 
Specifically, we show that the regret is bounded by the sum of covariances between $r_{ti}$ and $\| x_{ti} \|_{H_t^{-1}}$, given $S_t$. 
Thus, with some slight notational abuse (as the expressions do not strictly denote random variables), the regret can be bounded as follows:
\begin{align*}
    \Regret(\wb^\star)
    &\!\lesssim
     \beta_T(\delta) \!
    \sum_{t \notin \WarmupRounds} \operatorname{Cov}_t \left( r_{ti}, \| x_{ti} \|_{H_t^{-1}} \right)
    \\
    &\!\lesssim
    \beta_T(\delta) \!
    \sqrt{ \sum_{t \notin \WarmupRounds}   \VV_t(r_{ti} )}
    \sqrt{
            \sum_{t \notin \WarmupRounds} 
            \VV_t( \| x_{ti} \|_{H_t^{-1}} ) 
        }
    ,
\end{align*}
where $\Cov_t(\cdot, \cdot)$ and $\VV_t(\cdot)$ is the covariance and variance, respectively, given $S_t$.
For simplicity, rewrite $\VV_t(r_{ti} )$ as 
\begin{align*}
    \sigma_t^2 
    \!:= 
    \EE_{i \sim p_t(\cdot | S_t, \wb^\star)} 
        \!\left[
            \left(
                r_{ti} - 
                \EE_{j \sim p_t(\cdot | S_t, \wb^\star)} [r_{tj}]  
            \right)^2
        \right],
    \numberthis \label{eq:sigma_def}
\end{align*}
where $r_{t0}= 0$.
By applying the elliptical potential lemma (Lemma~\ref{lemma:epl_H}) to the sum of the variances of  $\| x_{ti} \|_{H_t^{-1}}$, we derive a variance-dependent regret bound. 
The complete proof is provided in Appendix~\ref{app_sec:proof_thm_regret_main}.
%
\begin{theorem}
\label{thm:regret_main}
    Let $\delta \in (0, 1]$, and assume that Assumption~\ref{assum:bounded_assumption} holds.
    Then, with probability at least $1-\delta$, the regret of~\textup{\AlgName{} }(Algorithm~\ref{alg:main_online}) satisfies
    \begin{align*}
    \Regret(\wb^\star) \!\lesssim&\,
            \left(d \log T 
            + B d \sqrt{ \log T} 
            \right)
            \sqrt{ \sum_{t=1}^T \sigma_t^2 
            }
            \\
            &+ \frac{1}{\kappa} B^3 d^2 
            \left(\log T \right)^2
            + \frac{1}{\kappa} B^4 d \log T
        .
    \end{align*}
\end{theorem}
\textbf{Discussion of Theorem~\ref{thm:regret_main}.} 
For sufficiently large, i.e., $T \geq \BigOTilde(e^{B^2} + \frac{1}{\kappa^2} B^8 d^2)$, \AlgName{} achieves a regret of $\BigO \Big( d \log T 
            \sqrt{ \sum_{t=1}^T \sigma_t^2 
            } \Big) $.
To the best of our knowledge, this is the first variance-dependent and 
$\operatorname{poly}(B), K$-free regret bound in contextual MNL bandits.
Compared to the recent minimax optimal result of $\BigO \left(B^{3/2} d  \log K (\log T)^{3/2} \sqrt{T} \right)$ by  \citet{lee2024nearly}, our method improves the regret by a factor of $\BigO \left( B^{3/2} \log K \sqrt{\log T} \right)$.
Moreover, the $\BigOTilde(\sqrt{T})$ term in \citet{lee2024nearly} is replaced in our result by $\BigOTilde \Big(  \sqrt{  \sum_{t=1}^T \sigma_t^2 } \Big)$.
Since $\sigma_t^2 \leq 1$ always holds, this represents a strict improvement over  $\sqrt{T}$.
\begin{remark} [Computational cost of~\AlgName{}] 
    The proposed algorithm, \AlgName{}, maintains a constant computational cost per round of $\BigO(Kd^3 + \operatorname{poly}(N))$, which is entirely independent of $t$.
    For parameter updates, we utilize the linearized loss, inspired by \citet{zhang2024online}, and work within ellipsoidal search spaces ($\Wcal$ and $\Wcal_t(\delta)$ in both phases. 
    As a result, the update process incurs only a cost of $\BigO(K d^3)$.
    Moreover, the assortment optimization problem can be solved in $\BigO(\operatorname{poly}(N))$~\citep{davis2014assortment}.
\end{remark}
\begin{remark} [Lower bound and optimality]
    For the worst-case regret, we achieve $\BigOTilde(d\sqrt{T})$ (since $\sigma_t = \BigO(1)$), which matches the minimax lower bound of $\Omega (d \sqrt{T})$ established by~\citet{lee2024nearly}.
    When the rewards are uniform, i.e., $r_{ti}= 1$, we obtain $\BigOTilde(d\sqrt{T/K})$, as $\sigma_t^2 \simeq p_t(0|S_t, \wb^\star) \simeq 1/K$.
    This result also matches the uniform reward minimax lower bound of $\Omega (d \sqrt{T/K})$~\citep{lee2024nearly}.
\end{remark}
%
%
\begin{figure*}[t!]
    \centering
        \includegraphics[clip, trim=0cm 0.0cm 0cm 0.0cm, width=\textwidth]{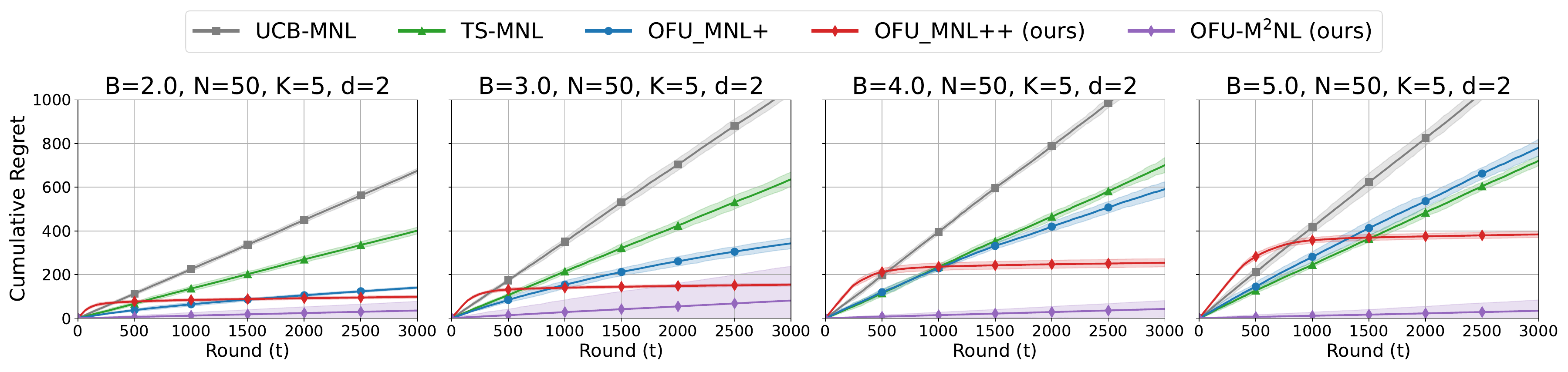}
    \caption{Cumulative regret for varying the norm-boundedness of the unknown parameter $B$.
    } 
    \label{fig:experiment}
\end{figure*}
\begin{figure*}[t!]
    \centering
        \includegraphics[clip, trim=0cm 0.0cm 0cm 0.0cm, width=\textwidth]{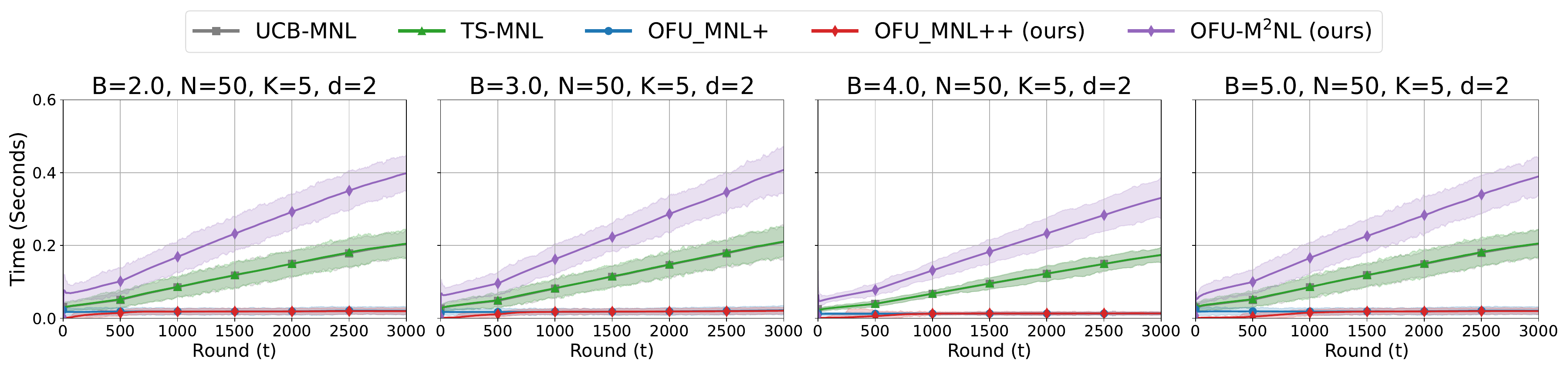}
    \caption{Runtime per round for varying the norm-boundedness of the unknown parameter $B$.
    } 
    \label{fig:experiment_runtime}
\end{figure*}
%
\textbf{Comparison to related works. }
While our approach shares some similarities with previous works~\citep{faury2022jointly, sawarni2024generalized} that also use a similar warm-up phase, there are significant differences.
\begin{remark} [Comparison to~\citet{faury2022jointly}]
    \citet{faury2022jointly}  incurs a $\operatorname{poly}(B)$ dependence in the leading term, whereas our method avoids this entirely by exploiting the self-concordant structure of the MNL loss (see Appendix~\ref{app_subsec:self_concordant}).
    Additionally, their use of MLE in the adaptive warm-up phase results in a per-round computation cost that grows linearly with the number of warm-up rounds. 
    In contrast, our method uses an online update rule, resulting in significantly better computational efficiency.
    Finally, their approach requires prior knowledge of $\kappa$, which is often unknown or hard to estimate in practice.
\end{remark}
\begin{remark} [Comparison to~\citet{sawarni2024generalized}]
    Unlike \citet{sawarni2024generalized}, which requires prior knowledge of  $\kappa$---an impractical assumption in real-world scenarios---our approach does not rely on knowing  $\kappa$ in advance. 
    Additionally, their method fully updates parameters using MLE rather than an online update. 
    As a result, the per-round computation cost of their algorithm scales linearly with $t$, while ours remains constant.
\end{remark}
%


\textbf{Discussion on instance-dependent regret.} 
As a special case, if the rewards are uniform (i.e., $r_{ti}=1$), we can establish an instance-dependent regret bound.
\begin{proposition}
\label{prop:instance_regret_uniform_r}
    Under the same conditions as Theorem~\ref{thm:regret_main} and assuming uniform rewards, for sufficiently large $T$, \textup{\AlgName{}} achieves a regret of 
    $\BigOTilde \Big( d \sqrt{ \sum_{t=1}^T  \kappa^\star_t }\Big)$, 
    where $\kappa_t^\star \!:=\! \sum_{i \in S_t^\star} p_t(i | S_t^\star, \mathbf{w}^\star)p_t(0 | S_t^\star, \mathbf{w}^\star)$.
\end{proposition}
This result improves upon the previous instance-dependent regret of $\BigOTilde \Big( e^{B} d \sqrt{  \sum_{t=1}^T \kappa^\star_t } \Big)$ (Proposition 2 of \citet{lee2024nearly}), by a factor of $e^B$.
The proof and further discussions are provided in Appendix~\ref{app_sec:discussion_instance_regret}.
%

\subsection{MLE-Based Approach}
\label{subsec:regret_MLE}
Inspired by \citet{lee2024unified}, who proposed a $\operatorname{poly}(B)$-free confidence bound using the MLE for generalized linear models (GLM) (but not for MNL models), we introduce an MLE-based algorithm that achieves $\operatorname{poly}(B), K$-free regret.
To this end, we first define the MLE estimator $\MLEParam_t$ as follows:
\begin{align*}
    \MLEParam_t := \argmin_{\wb \in \Wcal} \mathcal{L}_t(\wb), \quad
    \text{where }\,\, \mathcal{L}_t(\wb) = \sum_{s=1}^{t-1} \ell_s (\wb)
    .
\end{align*}
\begin{lemma} [Informal, Improved MLE confidence bound]
\label{lemma:MLE_CS_main}
    Let $\Gcal_t = \int_0^1 (1-v) \nabla^2 \mathcal{L}_t (\MLEParam_t + v (\wb^\star - \MLEParam_t)) \dd v + \frac{1}{8B^2} \Ib_d$.
    Then, for any $t \geq 1$, 
    if Assumption~\ref{assum:bounded_assumption} holds, then with probability at least $1-\delta$,
    we have:
    \begin{align*}
        \left\|
            \wb^\star - \MLEParam_t
        \right\|_{\Gcal_t}
        = \BigO \left(\sqrt{d \log (B t)} \right)
        .
    \end{align*}
\end{lemma}
Note that $\Gcal_t$ is used solely for analytical purposes.
The algorithm and proofs are provided in Appendix~\ref{app_sec:proof_thm:MLE}.
\begin{theorem} 
\label{thm:MLE}
    Let $\delta \in (0, 1]$.
    Then, under Assumption~\ref{assum:bounded_assumption}, with probability at least $1-\delta$, the regret of~\textup{\AlgNameMLE{} }(Algorithm~\ref{alg:MLE}) is bounded as follows:
    \begin{align*}
        \Regret(\wb^\star) \!\lesssim
            d \log (B T) 
            \sqrt{ \sum_{t=1}^T \sigma_t^2 
            }
            + \frac{1}{\kappa} d^2 
            \left(\log (B T)\right)^2.
    \end{align*}
\end{theorem}
\textbf{Discussion of Theorem~\ref{thm:MLE}.} 
Theorem~\ref{thm:MLE} shows that \AlgNameMLE{} enjoys a completely $\operatorname{poly}(B)$-free regret for any $T$, indicating that its regret is tighter than that of \AlgName{} by a factor of $\BigO(\operatorname{poly}(B))$ in the non-leading term.
However, its asymptotic regret still depends on $\log B$, whereas the asymptotic regret of \AlgName{} remains entirely independent of $B$.
Additionally, the per-round computational cost of \AlgNameMLE{} increases linearly with $t$, while that of \AlgName{} remains constant.

\section{Numerical Experiments}
\label{sec:experiments}
We empirically evaluate the performance of our algorithms,~\AlgName{} and~\AlgNameMLE{}, by measuring cumulative regret over $T=3000$ rounds.
The algorithms are tested on $20$ independent instances, and we report the average performance along with a shaded area representing two standard deviations.
In each instance, the true underlying parameter $\wb^\star$ is uniformly sampled from the $d$-dimensional ball $\BB^d(B)$ of radius $B$, and the context features $x_{ti}$ are drawn from a $\BB^d(1)$.
The rewards are sampled from a uniform distribution in each round, i.e., $r_{ti} \sim \operatorname{Unif}(0,1)$.

The baselines are the practical and state-of-the-art algorithms: the UCB-based algorithm, \texttt{UCB-MNL}~\citep{oh2019thompson}, the Thompson Sampling-based algorithm, \texttt{TS-MNL}~\citep{oh2019thompson},
and the constant-time algorithm, \texttt{OFU-MNL+}~\citep{lee2024nearly}.
Figure~\ref{fig:experiment} shows that both of our algorithms significantly outperform the baseline algorithms. 
Although \AlgName{} incurs high regret in the early rounds due to the adaptive warm-up phase (with the number of such rounds depending on $B$), its regret stabilizes after a certain point, exhibiting the lowest slope.
Therefore, we believe that \AlgName{} achieves the best asymptotic performance among all algorithms.
This aligns with our theoretical results, which show that the asymptotic regret of \AlgName{}, $ \BigO(d \log T \sqrt{T}) $, is entirely independent of $ B $ (even in logarithmic terms), whereas other algorithms exhibit $ B $-dependence.
Additionally, \AlgNameMLE{} demonstrates the most robust performance, maintaining its superiority even as $ B $ increases, particularly in the early rounds.
For more details and additional results, refer to Appendix~\ref{app_sec:experimat_details}.

Furthermore, Figure~\ref{fig:experiment_runtime} shows that
the online update methods (\texttt{OFU-MNL+} and \AlgName{}) maintain a constant runtime per round, while the others exhibit a linear increase with $t$ due to their use of MLE-based parameter estimation.
Among them, our MLE-based approach, \AlgNameMLE{}, is the most computationally expensive, as it solves a convex optimization problem to compute the optimistic parameter—unlike the others, which rely on closed-form UCBs (see Line 6 in Algorithm~\ref{alg:MLE}).

\section{Conclusion}
\label{sec:conclusion}
In this work, we construct the sharper online confidence bound for MNL models, with improvements in terms of $\log K$ and $\operatorname{poly}(B)$ dependencies.
Leveraging this result, we propose a constant-time algorithm, \AlgName{}, that achieves $B,K$-free regret in an asymptotic sense.  
Additionally, we introduce a MLE-based algorithm, \AlgNameMLE{}, which ensures $\operatorname{poly}(B),K$-free regret at every round.

\section*{Acknowledgments}
\label{sec:acknowledgments}
We sincerely thank Yu-Jie Zhang and Jungyhun Lee for their valuable discussions.
This work was supported by the National Research Foundation of Korea(NRF) grant funded by the Korea government(MSIT) (No.  RS-2022-NR071853 and RS-2023-00222663) and by AI-Bio Research Grant through Seoul National University.

\section*{Impact Statement}

This paper presents work whose goal is to advance the field of 
Machine Learning. There are many potential societal consequences 
of our work, none which we feel must be specifically highlighted here.



\bibliography{main_bib}
\bibliographystyle{icml2025}

\newpage
\appendix
\onecolumn

\counterwithin{table}{section}
\counterwithin{theorem}{section}
\counterwithin{algorithm}{section}
\counterwithin{figure}{section}
\counterwithin{equation}{section}
\counterwithin{condition}{section}

\addcontentsline{toc}{section}{Appendix} 
\part{Appendix} 
\parttoc 
\section{Notation} 
\label{app_sec:notation}
Let $T$ be the total number of rounds, with $t \in [T]$ representing the current round. 
We denote $N$ as the total number of items, $K$ as the maximum size of assortments, $d$ as the dimension of feature vectors, and $B$ as the upper bound on the norm of the unknown parameter.
For ease of reference, we provide Table~\ref{table_symbols}.
\begin{table}[h!]
\centering
    \caption{Symbols}
    \label{table_symbols}
    \begin{tabular}{ll}
         \toprule
         $x_{ti}$       &      feature vector for item $i$ given at round $t$\\[0.1cm]
         $r_{ti}$       &      reward for item $i$ given at round $t$\\[0.1cm]
         $S_t$       &      assortment chosen by an algorithm at round $t$ \\[0.1cm]   
         $K_t$       & $:= |S_t|$, size of chosen assortment at round $t$   \\[0.1cm] 
         $0$       &      outside option \\[0.1cm]   
         $y_{ti}$       &      choice response for each item $i \in S_t \cup \{0\}$ at round $t$ \\[0.1cm] 
         $ R_{t}(S, \wb^\star)$       &  $:=  \sum_{i \in S} p_t(i | S, \wb^\star) r_{ti} $,    expected revenue of the assortment $S$ at round $t$\\[0.1cm]
         $\ell_t(\wb) $       &   $:= - \sum_{i \in S_t} y_{ti} \log \left( \frac{\exp(x_{ti}^\top  \wb )}{ 1 \!+\!\sum_{j \in S_t }\exp( x_{tj}^\top \wb )}\right)$, loss function at round $t$ \\[0.1cm]
         $\bar{\ell}_t(\zb_t)$       &   $:=  - \sum_{i \in S_t} y_{ti} \log \left( \frac{\exp( z_{ti} )}{ 1 \!+\!\sum_{j \in S_t }\exp( z_{tj} )}\right)$,  loss function at round $t$, $z_{ti} = x_{ti}^\top \wb$\\[0.1cm]
         $\WarmupRounds$     &  set of adaptive warm-up rounds  \\[0.1cm] 
         $\wb_t$       &    online parameter estimate at round $t$ \\[0.1cm] 
         $\wb_t'$       &   projection of $\wb_t$ onto the current search space $\Wcal_t$ \\[0.1cm]
         $\WarmupParam_t$       &    adaptive warm-up parameter estimate at round $t$ \\[0.1cm] 
         $\eta$       &  $:=1$,  step-size parameter for $\wb_t$ \\[0.1cm] 
         $\WarmupStep$       &  $:= \frac{1}{2} + 3 \sqrt{2} B$,    step-size parameter for $\WarmupParam_t$ \\[0.1cm] 
         $\lambda$       &  $:=144 d$,  regularization parameter \\[0.1cm] 
         $\WarmupRegualizer$       &  $:= \max\{ 12\sqrt{2} \WarmupStep B, 144 \WarmupStep d, 2 \}$  regularization parameter for adaptive warm-up \\[0.1cm]
         $\nabla^2 \ell_t(\wb)$       &    $= \sum_{i \in S_t} p_t(i | S_t, \wb) x_{ti} x_{ti}^\top 
    -  \sum_{i \in S_t}  \sum_{j \in S_t} p_t(i | S_t, \wb) p_t(j | S_t, \wb) x_{ti} x_{tj}^\top$ \\[0.1cm] 
         $H_t$       &    $:= \lambda \Ib_d + \sum_{s \notin [t-1] \setminus \WarmupRounds} \nabla^2 \ell_s(\wb_{s+1})$ \\[0.1cm] 
         $\tilde{H}_{t}$       &  $:= H_t + \eta \nabla^2 \ell_t(\wb_t') \mathbbm{1}(t \notin \WarmupRounds) $ \\[0.1cm] 
         $H_t(\wb^\star)$      &  $:=  \frac{\lambda}{e} \Ib_d + \sum_{s \in [t-1] \setminus \WarmupRounds} \nabla^2 \ell_s(\wb^\star)$ \\[0.1cm] 
         $\WarmupHessian_t$     & $:= \WarmupRegualizer \Ib_d + \sum_{s \in \WarmupRounds \setminus [t,\dots,T]} \nabla^2 \ell_s (\wb_{s+1}) $  \\[0.1cm] 
         $\beta_t(\delta)$       &    $:= \BigO \left( \sqrt{d \log (t / \delta)} + B \sqrt{d} \right)$,  confidence radius for $\wb_t$ at round $t$ \\[0.1cm] 
         $\zeta_t(\delta)$       &    $:= \BigO \left( B \sqrt{d \log (t / \delta)} + B^{3/2} \sqrt{d} + B^2 \right)$,  confidence radius for $\WarmupParam_t$ at round $t$ \\[0.1cm] 
         $\tau_t$       &    $:= 6 \sqrt{2} \zeta_t(\delta)$,  threshold for determining whether to implement adaptive warm-up \\[0.1cm] 
         $\UCB_{ti}$       &    $:= x_{ti}^\top \wb_t + \beta_t(\delta) \| x_{ti} \|_{H_t^{-1}}$,  optimistic utility of item $i$ at round $t$ \\[0.1cm] 
         $\tilde{R}_{t}(S)$       &  $:= \frac{\sum_{i \in S} \exp( \UCB_{ti} ) r_{ti} }{1 + \sum_{j \in S} \exp(\UCB_{tj})}$,    optimistic expected revenue of assortment $S$ at round $t$ \\[0.1cm] 
         $\sigma_t^2$   &  $:= 
                            \EE_{i \sim p_t(\cdot | S_t, \wb^\star)} 
                                \!\left[
                                    \left(
                                        r_{ti} - 
                                        \EE_{j \sim p_t(\cdot | S_t, \wb^\star)} [r_{tj}]  
                                    \right)^2
                                \right]$, variance of rewards given $S_t$ at round $t$   \\[0.1cm]
         \bottomrule
    \end{tabular}
\end{table}

For notational simplicity, we express the loss function in two different forms throughout the proof, using them interchangeably as needed:
\begin{align*}
    \ell_t(\wb) &= - \sum_{i \in S_t} y_{ti} \log p_t(i | S_t, \wb)
    =- \sum_{i \in S_t} y_{ti} \log \left( \frac{\exp(x_{ti}^\top  \wb )}{ 1 \!+\!\sum_{j \in S_t }\exp( x_{tj}^\top \wb )}\right),
    \\
    \bar{\ell}_t(\zb_t) &= - \sum_{i \in S_t} y_{ti} \log \left( \frac{\exp( z_{ti} )}{ 1 \!+\!\sum_{j \in S_t }\exp( z_{tj} )}\right),
    \\
    \nabla_{\wb} \ell_t(\wb) &=
    \sum_{i \in S_t} \left(
        p_t(i | S_t, \wb) - y_{ti}
    \right)x_{ti},
    \\
    \nabla_{\zb} \bar{\ell}_t(\zb_t) &= 
    \sigmab_t(\zb_t^\star) - \yb_t,
    \\
    \nabla_{\wb}^2 \ell_t(\wb) &=
    \sum_{i \in S_t} p_t(i | S_t, \wb) x_{ti} x_{ti}^\top 
    -  \sum_{i \in S_t}  \sum_{j \in S_t} p_t(i | S_t, \wb) p_t(j | S_t, \wb) x_{ti} x_{tj}^\top,
    \\
    \nabla_{\zb}^2 \bar{\ell}_t(\zb_t) &= 
    \operatorname{diag}(\sigmab_t(\zb_t^\star)) - \sigmab_t(\zb_t^\star)\sigmab_t(\zb_t^\star)^\top
    ,
    \numberthis \label{eq:equi_loss}
\end{align*}
where  $z_{ti} = x_{ti}^\top \wb$, $\zb_t = (z_{ti})_{i \in S_t} \in \RR^{|S_t|}$, and $\yb_t = (y_{ti})_{i \in S_t} \in \RR^{|S_t|}$.
Hence, it is clear that $\ell_t(\wb) = \bar{\ell}_t(\zb_t)$.

\section{Self-Concordant Properties of MNL Function}
\label{app_subsec:self_concordant}
In this section, we present several key properties of self-concordant-like functions that are essential for proving the main theorems in this paper. 

For simplicity, we will work with the MNL loss in the form of $\bar{\ell}$ rather than $\ell$ throughout this section.
However, it is important to note that the properties introduced in this section also apply to $\ell$. 
Whenever these properties are used in the proofs of other lemmas or theorems, we will explicitly demonstrate their applicability to $\ell$. 

We begin by revisiting the definition of self-concordant-like functions.
\begin{definition}[Self-concordant-like function,~\citealt{tran2015composite}]
\label{def:self-concordant-like}
A convex function $f \in \mathcal{C}^3: \RR^K \rightarrow \RR$ is $M$-self-concordant-like  function with constant $M$ if:
    \begin{align*}
        |\phi^{\prime\prime\prime} (s) | \leq M \|\bb \|_2 \phi^{\prime\prime}(s) .
    \end{align*}
    for $s \in \RR$ and $M > 0$, where $\phi(s) := f(\ab + s\bb)$ for any  $\ab, \bb \in \RR^K$.
\end{definition}
To derive a tighter confidence bound in Theorem~\ref{thm:online_confidence_set} and a tighter regret bound in Theorem~\ref{thm:regret_main}, we redefine the concept of self-concordant-like functions specifically for the \textit{MNL loss function} $\bar{\ell}$.
\begin{definition}[$\ell_\infty$-norm self-concordant-like MNL loss]
\label{def:self-concordant-like_MNL}
The MNL loss function $\bar{\ell}(\zb): \RR^K \rightarrow \RR$ is $M$-self-concordant-like  function with constant $M$ if:
    \begin{align*}
        |\phi^{\prime\prime\prime} (s) | \leq M \|\bb \|_\infty \phi^{\prime\prime}(s) .
    \end{align*}
    for $s \in \RR$ and $M > 0$, where $\phi(s) := \bar{\ell}(\ab + s\bb)$ for any  $\ab, \bb \in \RR^K$.
\end{definition}
Note that because $\| \xb \|_\infty \leq \| \xb \|_2$ for any vector $\wb \in \RR^K$, the new definition of a self-concordant-like function (Definition~\ref{def:self-concordant-like_MNL}), which is specifically designed for the MNL loss function, is tighter than the original definition (Definition~\ref{def:self-concordant-like}).

Using this new definition, we show that the MNL loss defined in~\eqref{eq:loss} is a $3\sqrt{2}$-self-concordant-like function.
\begin{proposition}[Constant self-concordant-like MNL loss] \label{prop:self_concordant}
    For any $t \in [T]$, the multinomial logistic loss $\bar{\ell}_{t}$, defined in Equation~\eqref{eq:equi_loss}, is  $3\sqrt{2}$-self-concordant-like function under Definition~\ref{def:self-concordant-like_MNL}.
\end{proposition}
\begin{proof} [Proof of Proposition~\ref{prop:self_concordant}]
    Recall that the loss  $\bar{\ell}_{t}$ is defined as:
    \begin{align*}
        \bar{\ell}_{t} (\zb) 
        =  \underbrace{-\sum_{i \in S_t} y_{ti} z_{ti} }_{\text{linear}}
        + \underbrace{\log \left( 1 +\sum_{i \in S_t } e^{z_{ti}} \right)}_{=: f(\zb)}
    \end{align*}
    Since $\bar{\ell}_{t}$ consists of the linear term and $f(\zb): \RR^
    {|S_t|} \rightarrow \RR$, and the third derivatives of the linear term are zero, it suffices to show that  $f(\zb)$ is a $3\sqrt{2}$-self-concordant-like function.

    Fix any $t \in [T]$.
    For simplicity, let $K = |S_t|$.
    We define:
    \begin{align*}
        \phi(s) := f(\ab + s \bb)
    = \log \left( 1 + \sum_{i =1}^K e^{a_i+ s b_i} \right)
    = \log \left( \sum_{i =0}^K e^{a_i+ s b_i } \right),
    \end{align*}
    where  $\ab = [a_1, \dots, a_{K}]^\top \in \RR^{K}$ and  $\bb = [b_1, \dots, b_{K}]^\top \in \RR^{K}$, and $a_0 = b_0 =0$.
    Then, by simple calculus, we have
    \begin{align*}
        \phi^{\prime\prime} (s)
        = \frac{\sum_{i<j} (b_i - b_j)^2 e^{a_i + sb_i} e^{a_j + s b_j} }{\left(\sum_{i=0}^K e^{a_i + s b_i} \right)^2}
        \geq 0,
    \end{align*}
    and
    \begin{align*}
        \phi^{\prime\prime\prime} (s)
        = \frac{\sum_{i<j} (b_i - b_j)^2 e^{a_i + s b_i} e^{a_j + s b_j} 
        \left[ \sum_{k=0}^K (b_i + b_j -2b_k)e^{a_k + s b_k} \right]}{\left(\sum_{i=0}^K e^{a_i + s b_i} \right)^3}
        \leq  \left| 
            \frac{\sum_{k=0}^K (b_i + b_j -2a_k)e^{a_k + s b_k}}
            {\sum_{i=0}^K e^{a_i + s b_i}}
        \right| 
        \phi^{\prime\prime} (s) 
        .
        \numberthis \label{eq:self_concordant_third_derivative}
    \end{align*}
    Note that for all $i,j,k = 0, \dots, K$, 
    \begin{align*}
        |b_i + b_j -2b_k| \leq \sqrt{6}\sqrt{b_i^2 + b_j^2 + b_k^2} \leq 3\sqrt{2} \max_{i=0, \dots, K} | b_i|. 
    \end{align*}
    Hence, we obtain
    \begin{align*}
        \left| \sum_{k=0}^K (b_i + b_j -2b_k)e^{a_k +s b_k} \right|
        \leq  \sum_{k=0}^K \left|b_i + b_j -2b_k\right|e^{a_k + s b_k} 
        \leq 3\sqrt{2} \max_{i=0, \dots, K} | b_i | \sum_{i=0}^K e^{a_i + s b_i}.
        \numberthis \label{eq:self_concordant_improved_norm2}
    \end{align*}
    Plugging in~\eqref{eq:self_concordant_improved_norm2} into~\eqref{eq:self_concordant_third_derivative}, we derive that
    \begin{align*}
        \phi^{\prime\prime\prime}(s)
        \leq 3\sqrt{2} \max_{i=0, \dots, K} |b_i | \phi^{\prime\prime}(s)
        = 3\sqrt{2} \| \bb \|_{\infty} \phi^{\prime\prime}(s)
        .
    \end{align*}
    By Definition~\ref{def:self-concordant-like_MNL}, we conclude that the MNL loss is 
    $3\sqrt{2}$-self-concordant-like.
\end{proof}
Building on our new definition (Definition~\ref{def:self-concordant-like_MNL}), we establish several fundamental properties of the self-concordant-like MNL loss function.
The following proposition is analogous to Theorem 3 of~\citet{tran2015composite}. However, Proposition~\ref{prop:equi_self_concordant_MNL} provides a tighter result specifically tailored to the MNL loss function (though it may be extendable to other functions).
\begin{proposition}
\label{prop:equi_self_concordant_MNL}
    For a convex function $f \in \Ccal^3 : \RR^K \rightarrow \RR$,
    we define $D^3 f(\xb)[\ub, \ub, \ub] := \langle D^3 f(\xb)[\ub]\ub, \ub \rangle$.
    Then, if $f$ is the MNL loss function, i.e., $f = \bar{\ell}$, then for any $\xb, \ub_1, \ub_2 \in \RR^K$, we have:
    \begin{align*}
     \left| D^3 f(\xb) [\ub_1, \ub_2, \ub_2]  \right|
     \leq 3\sqrt{2} \, \| \ub_1 \|_{\infty} \| \ub_2 \|^2_{\nabla^2 f (\xb)}.
    \end{align*}
\end{proposition}
\begin{proof} [Proof of Proposition~\ref{prop:equi_self_concordant_MNL}]
    Let $\phi(s) = f(\ab + s \bb)$.
    Then, we have
    \begin{align*}
        \phi^{\prime \prime} (s) = \nabla^2 f(\ab + s \bb) \bb^\top \bb,
        \quad
        \phi^{\prime \prime \prime} (s) = D^3 f(\ab + s \bb)[\bb, \bb, \bb].
    \end{align*}
    By Definition~\ref{def:self-concordant-like_MNL} and Proposition~\ref{prop:self_concordant}, we know that
    \begin{align*}
        \left|\phi^{\prime \prime \prime} (s) \right|
        \leq 3\sqrt{2} \| \bb\|_{\infty} \phi^{\prime \prime \prime} (s).
    \end{align*}
    By substituting $s=0$, $\ab= \xb$, and $\bb = \ub_1$, we get
    \begin{align*}
        \left| D^3 f(\xb)[\ub_1, \ub_1, \ub_1]  \right|
        = \left| \phi^{\prime \prime \prime} (0) \right|
        \leq 3\sqrt{2} \| \ub_1 \|_{\infty} \phi^{\prime \prime \prime} (0)
        = 3\sqrt{2} \| \ub_1 \|_{\infty} \nabla^2 f(\xb) \ub_1^\top \ub_1,
    \end{align*}
    which can be equivalently expressed as
    \begin{align*}
        -3\sqrt{2} \| \ub_1 \|_{\infty} \nabla^2 f(\xb)
        \preceq D^3 f(\xb)[\ub_1] 
        \preceq 
        3\sqrt{2} \| \ub_1 \|_{\infty} \nabla^2 f(\xb).
    \end{align*}
    Therefore, for any $\ub_2 \in \RR^K$, we have
    \begin{align*}
        \left| \ub_2^\top D^3 f(\xb)[\ub_1] \ub_2 \right|
        &\leq  3\sqrt{2} \| \ub_1 \|_{\infty} \ub_2^\top \nabla^2 f(\xb) \ub_2
        \\
        \Longleftrightarrow
        \left| D^3 f(\xb) [\ub_1, \ub_2, \ub_2]   \right|
        &\leq 3\sqrt{2} \, \| \ub_1 \|_{\infty} \| \ub_2 \|^2_{\nabla^2 f (\xb)}.
    \end{align*}
    This concludes the proof.
\end{proof}
Proposition~\ref{prop:self_hessian_norm}, a variant of Theorem 4 in~\citet{tran2015composite}, establishes a key inequality for the Hessian of the MNL loss, which plays a crucial role in eliminating $B$-dependency.
\begin{proposition}
\label{prop:self_hessian_norm}
For any $t \in [T]$, 
the Hessian of the multinomial logistic loss $\bar{\ell}_t: \RR^{|S_t|} \rightarrow \RR$ satisfies that, for any $\zb_1, \zb_2 \in \RR^{|S_t|}$, we have:
\begin{align*}
    e^{-3\sqrt{2} \| \zb_1 - \zb_2 \|_{\infty} }  \nabla^2 \bar{\ell}_t  (\zb_1)
    \preceq \nabla^2  \bar{\ell}_t (\zb_2)
    \preceq e^{3\sqrt{2} \| \zb_1 - \zb_2 \|_{\infty} } \nabla^2 \bar{\ell}_t (\zb_1).
\end{align*}
\end{proposition}
\begin{proof} [Proof of Proposition~\ref{prop:self_hessian_norm}]
    We denote $\zb_s = \zb_1 + s(\zb_2 - \zb_1)$ for notational convenience, where $s \in [0,1]$.
    We define the function $\psi(s) := \ub^\top \nabla^2 \bar{\ell}_t (\zb_s) \ub = \| \ub \|^2_{\nabla^2 \bar{\ell}_t (\zb_s)}$.
    Note that $\psi(0) =  \| \ub \|^2_{\nabla^2 \bar{\ell}_t (\zb_1)}$
    and $\psi(1) =  \| \ub \|^2_{\nabla^2 \bar{\ell}_t (\zb_2)}$.
    Then, by Proposition~\ref{prop:equi_self_concordant_MNL}, we have
    \begin{align*}
        \left| \psi^{\prime} (s) \right|    
        = \left| D^3 \bar{\ell}_t (\zb_s) [\zb_2 - \zb_1, \ub, \ub ] \right|
        \leq 3\sqrt{2} \| \zb_2 - \zb_1 \|_{\infty}   \psi (s)
        ,
    \end{align*}
    which can be  equivalently written as follows:
    \begin{align*}
        \left| \frac{d \ln \psi (s) }{d s} \right|
        \leq 3\sqrt{2} \| \zb_2 - \zb_1 \|_{\infty}.
    \end{align*}
    By integrating both sides over $s \in [0,1]$, we conclude the proof.
\end{proof}
Additionally, we introduce an improved version of Proposition 6 in~\citet{perivier2022dynamic}, which serves as a useful tool for the subsequent proofs.
\begin{proposition}
\label{prop:hessian_usedful}
    For any $t \in [T]$, 
    the Hessian of the multinomial logistic loss $\bar{\ell}_t: \RR^{|S_t|} \rightarrow \RR$ satisfies the following for any $\ub, \zb_1, \zb_2 \in \RR^{|S_t|}$:
    \begin{align*}
        \ub^\top 
        \left(\int_0^1 (1-s) \nabla^2 \bar{\ell}_t (\zb_1 + s (\zb_2 - \zb_1) ) \dd s  \right)
        \ub
        \geq \frac{1}{ 2 + 3\sqrt{2} \| \zb_2 - \zb_1 \|_\infty } \ub^\top \nabla^2 \bar{\ell}_t (\zb_1) \ub.
    \end{align*}
\end{proposition}
\begin{proof} [Proof of Proposition~\ref{prop:hessian_usedful}]
    From Proposition~\ref{prop:self_hessian_norm}, we have
    \begin{align*}
        \ub^\top 
        \left(\int_0^1 (1-s) \nabla^2 \bar{\ell}_t (\zb_1 + s (\zb_2 - \zb_1) ) \dd s  \right) 
        \ub
        &\geq \ub^\top 
        \nabla^2 \bar{\ell}_t (\zb_1)
        \ub
        \int_0^1 (1-s) e^{-3 \sqrt{2} \| s (\zb_2 - \zb_1) \|_\infty } \dd s
        \\
        &\geq \ub^\top 
        \nabla^2 \bar{\ell}_t (\zb_1)
        \ub
        \left(
            \frac{1}{3\sqrt{2}  \| (\zb_2 - \zb_1) \|_\infty}
            + \frac{ e^{-3\sqrt{2}  \| (\zb_2 - \zb_1) \|_\infty}  - 1}{\left(3\sqrt{2}  \| (\zb_2 - \zb_1) \|_\infty \right)^2 }
        \right)
        \\
        &\geq \ub^\top 
        \nabla^2 \bar{\ell}_t (\zb_1)
        \ub 
        \left(
            \frac{1  }{2 + 3\sqrt{2}  \| (\zb_2 - \zb_1) \|_\infty}
        \right)
        ,
    \end{align*}
    where in the third inequality, we use the fact that $\frac{1}{x} \left( 1 + \frac{e^{-x}-1}{x} \right) \geq \frac{1}{2+x}$ for all $x \geq 0$.
\end{proof}

\section{Proof of Theorem~\ref{thm:online_confidence_set}} 
\label{app_sec:proof_thm_confidence}
In this section, we provide the proof of Theorem~\ref{thm:online_confidence_set}.
We begin with the main proof of the theorem, followed by the proof of the technical lemma that is used within the main argument.
\subsection{Main Proof of Theorem~\ref{thm:online_confidence_set}} \label{app_subsec:main_proof_thm_confidence}
\begin{proof} [Proof of Theorem~\ref{thm:online_confidence_set}]
The overall proof structure is similar to the analysis presented in~\citet{zhang2024online, lee2024nearly}.
However, as explained in the main paper, several novel analytical techniques are introduced to derive a $B$-improved, $K$-free confidence bound, including:
\begin{enumerate}
    \item $B,K$-independent step size $\eta$ by leveraging improved self-concordant properties,
    \item $\BigO(\sqrt{\log t} \log K)$ via a sharper analysis using Ville’s inequality~\citep{ville1939etude}.
\end{enumerate}
Throughout the proof of Theorem~\ref{thm:online_confidence_set},  we denote $\Tcal \subseteq [T]$ as the set of total update rounds.
For any round $t \in [T]$, we denote $\Tcal_t \subseteq \Tcal$ as the set of update rounds that occur before $t$, i.e., $\Tcal_t = \{ s \in \Tcal : s < t  \}= \Tcal \setminus \{t, t+1, \dots, T \} $.
We assume the following conditions hold:
\begin{condition} [Update condition]
\label{condition:update}
    For all $t \in \Tcal$, we assume that
    \begin{align*}
        \sup_{ \wb \in \Wcal_t}  
            |x_{ti}^\top (\wb  - \wb^\star) |
             \leq \alpha,
        \quad \forall i  \in S_{t},
    \end{align*}
    where  $\Wcal_t$  is a compact convex set,
    and $\alpha > 0$.
\end{condition}
We also denote the size of the assortment at round $t$ as $K_t$, i.e.,  $K_t= |S_t| \leq K$.

\begin{lemma} \label{lemma:online_parameter_gap_bound}
    Suppose Condition~\ref{condition:update} holds.
    The update rule for the parameter at round $t \in \Tcal$ is defined as:
    \begin{align*}
        \wb_t' &= \argmin_{\wb \in \Wcal_t} \| \wb - \wb_t \|_{H_t},
        \\
        \wb_{t+1} &= \argmin_{\wb \in \Wcal_t} \widetilde{\ell}_{t} (\wb)  + \frac{1}{2 \eta} \| \wb - \wb_t' \|_{H_{t}}^2,
    \end{align*}
    where $  \widetilde{\ell}_{t} (\wb) 
    = \ell_{t}(\wb_t') 
    + \langle \wb - \wb_t', \nabla \ell_{t}(\wb_t') \rangle 
    + \frac{1}{2} \|\wb - \wb_t' \|^2_{\nabla^2 \ell_{t}(\wb_t') }$.
    Let $\eta =  1 + \frac{3\sqrt{2}}{2} \alpha$ and $\lambda \geq 12 \sqrt{2} \eta \alpha$.
    Then,  under Assumption~\ref{assum:bounded_assumption}, 
    for any update round $t \in \Tcal$, we have
    \begin{align*}
        \| \wb_{t+1} - \wb^\star \|_{H_{t+1}}^2 
        \leq 2 \eta 
        \left(
            \sum_{s \in \Tcal_{t+ 1}} \ell_{s}(\wb^\star)
            - \sum_{s \in \Tcal_{t+1}}
        \ell_{s}(\wb_{s+1}) 
        \right)
        + 4 B^2 \lambda 
        -   \frac{1}{2} \sum_{s \in \Tcal_{t+1}}  \| \wb_{s+1} - \wb_s' \|_{H_s}^2.
        \numberthis \label{eq:online_parameter_gap_bound}
    \end{align*}
\end{lemma}
The proof is deferred to Appendix~\ref{app_subsubsec:proof_lemma:online_parameter_gap_bound}.
Following the approach of~\citet{faury2022jointly, zhang2024online, lee2024nearly}, to bound the first term in Equation~\eqref{eq:online_parameter_gap_bound}, we introduce an intermediary parameter that is $\Fcal_s$-measurable.
Note that $\wb_{s+1}$ is $\Fcal_s$-measurable.

To do so, we first define the softmax function at round $t$, denoted as $\sigmab_t(\zb): \RR^{K_t} \rightarrow \RR^{K_t}$, as follows:
\begin{align}
    [\sigmab_t(\zb)]_{i} = \frac{\exp([\zb]_{i})}{ 1+ \sum_{k =1}^{K_t}   \exp([\zb]_{k})}, \quad \forall i \in [K_t]
    , 
    \label{eq:softmax}
\end{align}
where $[\cdot]_{i}$ denotes $i$'th element of the input vector.
The probability of choosing the outside option is denoted as:
\begin{align*}
    [\sigmab_t(\zb)]_{0} = \frac{1}{ 1+ \sum_{k =1}^{K_t}   \exp([\zb]_{k})}
\end{align*}
Although $[\sigmab_t(\zb)]_{0}$ is not part of the output vector of the softmax function $\sigmab_t(\zb)$, it is expressed in a similar form to~\eqref{eq:softmax} for simplicity.
Then, the MNL user choice model can be equivalently expressed as $p_t(i | S_t, \wb) = \left[\sigmab_t\left( (x_{tj}^\top \wb)_{j \in S_t}  \right)\right]_{i}$ for all $i \in [K_t]$ and $p_t(0 | S_t, \wb) = \left[\sigmab_t\left( (x_{tj}^\top \wb)_{j \in S_t}  \right)\right]_{0}$.
Furthermore, the loss function in~\eqref{eq:loss} can also be expressed as $\ell(\zb_t, \yb_t) = \sum_{k=0}^{K_t} \mathbf{1}\left\{ y_{ti} = 1 \right\} \cdot \log\left(\frac{1}{[\sigmab_t(\zb_t)]_k}\right)$.

We also define a pseudo-inverse function of $\sigmab_t(\cdot)$ as $\sigmab_t^+ : \RR^{K_t} \rightarrow \RR^{K_t}$, where $[\sigmab_t^+(\qb)]_i = \log \left( q_i / (1- \| \qb \|_1)\right)$ for any $\qb \in \{ \pb \in [0,1]^{K_t} \mid \| \pb \|_1 < 1 \}$.
Then, we define the intermediary parameter as follows:
\begin{align*}
    \tilde{\zb}_s := \sigmab_s^+ \left( \EE_{\wb \sim P_s} \left[\sigmab_s\left( (x_{sj}^\top \wb)_{j \in S_s} \right) \right] \right).
    \numberthis \label{eq:def_intermediary_param}
\end{align*}
where $P_s := 
\mathcal{N} \left( \wb_s, c H_s^{-1} \right) $ is a multivariate normal distribution with mean $\wb_s$ and covariance $cH_s^{-1}$.
Here, $c >0$  is a positive constant to be specified later.
Note that $\tilde{\zb}_s$ is $\Fcal_s$-measurable unlike  $\wb_{s+1}$ ($\wb_{s+1}$ is $\Fcal_{s+1}$-measurable).
In general, $\tilde{\zb}_s$ cannot be expressed as a linear function of the features $\{ x_{sj} \}_{j \in S_s}$.
Then, the first term in Equation~\eqref{eq:online_parameter_gap_bound} can be decomposed into two terms as follows:
\begin{align*}
    \underbrace{\sum_{s \in \Tcal_{t+ 1}} \ell_{s}(\wb^\star)
    - \sum_{s \in \Tcal_{t+ 1}} \bar{\ell}_s(\tilde{\zb}_s)}_{(a)}
    + \underbrace{\sum_{s \in \Tcal_{t+ 1}} \bar{\ell}_s(\tilde{\zb}_s)
    - \sum_{s \in \Tcal_{t+1}} \ell_{s}(\wb_{s+1}) }_{(b)}
\end{align*}
First, we demonstrate that the term $(a)$ is bounded by  $\BigO\left( \log \frac{1}{\delta} \right)$ with high probability.
\begin{lemma} \label{lemma:online_regret_intermediate_(a)}
    Let $\delta \in (0,1]$.
    Assume that Condition~\ref{condition:update} holds.
    Define the intermediary parameter as Equation~\eqref{eq:def_intermediary_param} with 
    and $c > 0$.
    Then,
    for any $t \in \Tcal$, with probability at least $1-\delta$, we have
    \begin{align*}
        \sum_{s \in \Tcal_{t+ 1}} \ell_{s}(\wb^\star)
        - \sum_{s \in \Tcal_{t+ 1}} \bar{\ell}_s(\tilde{\zb}_s) 
        \leq \log \frac{1}{\delta}.
    \end{align*}
\end{lemma}
The proof is deferred to Appendix~\ref{app_subsubsec:proof_lemma:online_regret_intermediate_(a)}.
By setting $\frac{1}{\delta} = \BigO (t)$, we obtain the bound $\sum_{s \in \Tcal_{t+ 1}} \ell_{s}(\wb^\star)
        - \sum_{s \in \Tcal_{t+ 1}} \bar{\ell}_s(\tilde{\zb}_s)  = \BigO(\log t)$.
Compared to Lemma F.2 of~\citet{lee2024nearly}, which bound the similar term by $\BigO\left( (\log t)^2 \log K  \right)$, Lemma~\ref{lemma:online_regret_intermediate_(a)}  improves the bound by a factor of $\log t \log K$.
This improvement is primarily due to the use of a more refined analysis based on Ville’s inequality~\citep{ville1939etude}, 
rather than the Bernstein-type inequality with a \textit{smoothed intermediate} term adopted in~\citet{zhang2024online, lee2024nearly}. 
The latter approach incurs an additional $\log(Kt)$ factor, which ultimately leads to the looser bound of $\BigO\left( (\log t)^2 \log K \right)$ for the term $(a)$.

Now, we bound the term $(b)$ by the following lemma:
\begin{lemma} \label{lemma:bound_(b)}
    Let  
    $c>0$ and 
    $\lambda \geq \max\{ 2,  72 c d \}$.
    Then, for all $t \in  \Tcal$, we have
    \begin{align*}
        \sum_{s \in \Tcal_{t+1}}\bar{\ell}_s(\tilde{\zb}_{s}) 
    - \sum_{s \in \Tcal_{t+1}}\ell_{s}(\wb_{s+1}) 
    \leq \frac{1}{2c} \sum_{s \in \Tcal_{t+1}} \|   \wb_{s+1} - \wb_s' \|_{H_s}^2
    +  \sqrt{6} c d \log \left(t + 2\right).
    \end{align*}
\end{lemma}
The proof is deferred to Appendix~\ref{app_subsubsec:proof_lemma:bound_(b)}.

Finally, by combining Lemma~\ref{lemma:online_parameter_gap_bound}, Lemma~\ref{lemma:online_regret_intermediate_(a)}, and Lemma~\ref{lemma:bound_(b)}, we derive that
\begin{align*}
    &\| \wb_{t+1} - \wb^\star \|_{H_{t+1}}^2 
    \\
    &\leq 
    2   \eta
    \log \frac{1}{\delta}
    + \frac{\eta}{c} \sum_{s \in \Tcal_{t+1}} \|  \wb_{s+1}  -  \wb_s' \|_{H_s}^2
    +  2\sqrt{6} \eta c d  \log \left( t + 2 \right)
    + 4 B^2 \lambda 
    -  \frac{1}{2} \sum_{s \in \Tcal_{t+1}}  \| \wb_{s+1} - \wb_s' \|_{H_s}^2  
    \\
    &\leq 
     2   \eta
    \log \frac{1}{\delta}
    +  4\sqrt{6} \eta^2  d \log \left( t + 2\right)
    + 4 B^2 \lambda 
    \tag{Set $c = 2 \eta$}
    \\
    &= \BigO \left(
        \alpha^2
        \cdot
        d \log (t/\delta)
        +  B^2 \lambda 
    \right)
    \tag{$\eta = \frac{1 + 3 \sqrt{2} \alpha}{2}$}
    .
\end{align*}
This implies that for all $t \in \Tcal \setminus \{t_1\}$, where $t_1$ denote the first update round, we have
\begin{align*}
     \| \wb_t - \wb^\star \|_{H_{t}}
     \leq  \BigO \left(
        \alpha
        \sqrt{
            d \log (t/\delta)
            }
            + B \sqrt{\lambda}
        \right).
        \numberthis \label{eq:thm1_final}
\end{align*}
For $t_1 \in \Tcal$, we know that $\| \wb_{t_1} - \wb^\star \|_{H_t} \leq B \sqrt{\lambda} $.
Thus, Equation~\eqref{eq:thm1_final} holds for all  $t \in \Tcal$.
This concludes the proof of Theorem~\ref{thm:online_confidence_set}.
\end{proof}

\subsection{Proofs of Lemmas for Theorem~\ref{thm:online_confidence_set}} 
\label{app_subsec:proof_lemma:online_confidence_set}
\subsubsection{Proof of Lemma~\ref{lemma:online_parameter_gap_bound}  }
\label{app_subsubsec:proof_lemma:online_parameter_gap_bound}
\begin{proof}[Proof of Lemma~\ref{lemma:online_parameter_gap_bound}]
For any update round $s \in \Tcal_t$ (an update round occurring before $t \in \Tcal$), 
let $\tilde{\ell}_{s}(\wb) = \ell_{s}(\wb_s') + \langle  \nabla \ell_{s}(\wb_s'), \wb - \wb_s' \rangle + \frac{1}{2} \|\wb - \wb_s' \|^2_{\nabla^2 \ell_{s}(\wb_s')}$ be a second-order approximation of the original function $\ell_{s}(\wb)$ at the point $\wb_s'$, where $\wb_s' = \argmin_{\wb \in \Wcal_s} \| \wb - \wb_s \|_{H_s}$ is the projection of 
$\wb_s$
onto $\Wcal_s$.
Then, the update rule in~\eqref{eq:online_update} can be equivalently rewritten as follows:
\begin{align*}
    \wb_{s+1} 
    &= \argmin_{\wb \in \Wcal_s}  \, \langle \nabla \ell_t (\wb_{ s } ), \wb \rangle
    + \frac{1}{2 \eta} \| \wb - \wb_s' \|_{\tilde{H}_{s}}^2
    \\
    &= \argmin_{\wb \in \Wcal_s  } \, \langle \nabla \ell_t (\wb_{ s } ), \wb \rangle
    + \frac{1}{2} \|  \wb - \wb_s' \|_{\nabla^2 \ell_s (\wb_s)}
    + \frac{1}{2 \eta} \| \wb - \wb_s' \|_{H_s}^2
    \\
    &= \argmin_{\wb \in \Wcal_s } \, \tilde{\ell}_s(\wb)
    + \frac{1}{2 \eta} \| \wb - \wb_s' \|_{H_s}^2
    .
\end{align*}
Then, by applying Lemma~\ref{lemma:loss_firstorder_decomposition}, we get
    \begin{align*}
        \langle \nabla \tilde{\ell}_{s}(\wb_{s+1}), \wb_{s+1} - \wb^\star  \rangle
        &\leq \frac{1}{2\eta} \left( \| \wb_s' - \wb^\star  \|_{H_s}^2 - \| \wb_{s+1} - \wb^\star \|_{H_s}^2 - \| \wb_{s+1} - \wb_s' \|_{H_s}^2  \right)
        \\
        &\leq \frac{1}{2\eta} \left( \| \wb_s - \wb^\star  \|_{H_s}^2 - \| \wb_{s+1} - \wb^\star \|_{H_s}^2 - \| \wb_{s+1} - \wb_s' \|_{H_s}^2  \right),
        \numberthis \label{eq:concentration_firstorder_bound}
    \end{align*}
    where the last inequality holds due to the nonexpansive property of the projection mapping $P_{\Wcal_s}$, i.e., $\| \wb_s' - \wb^\star \|_{H_s}^2 
    = \| P_{\Wcal_s}(\wb_s) - P_{\Wcal_s}(\wb^\star) \|_{H_s}^2 
    \leq \| \wb_s - \wb^\star \|_{H_s}^2
    $.
    On the other hand, by applying Lemma~\ref{lemma:second_order_loss}, which is based on our improved self-concordant-like property (Proposition~\ref{prop:equi_self_concordant_MNL}), we obtain:
    \begin{align}
        \ell_{s}(\wb_{s+1}) - \ell_{s}(\wb^\star)
        \leq \langle \nabla \ell_{s}(\wb_{s+1}), \wb_{s+1} - \wb^\star  \rangle
        -\frac{1}{ 2 + 3\sqrt{2} \alpha} \| \wb_{s+1} - \wb^\star \|^2_{\nabla^2 \ell_{s}(\wb_{s+1})}.
        \label{eq:concentration_lossgap_bound}
    \end{align}
    Let $\eta = 1 + \frac{3\sqrt{2}}{2} \alpha$.
    Then, by combining~\eqref{eq:concentration_firstorder_bound} and~\eqref{eq:concentration_lossgap_bound}, we obtain that
    \begin{align*}
        \ell_{s}(\wb_{s+1}) - \ell_{s}(\wb^\star)
        &\leq \langle \nabla  \ell_{s}(\wb_{s+1})- \nabla\tilde{\ell}_{s}(\wb_{s+1}) , \wb_{s+1} - \wb^\star  \rangle
        \\
        &+ \frac{1}{2 \eta} \left( \| \wb_s - \wb^\star \|_{H_s}^2 
        - \| \wb_{s+1} - \wb^\star \|_{H_{s+1}}^2 
        - \| \wb_{s+1} - \wb_s' \|_{H_s}^2  \right).
        \numberthis \label{eq:lemma:online_parameter_gap_bound_mid}
    \end{align*}
    In the inequality above, the first term on the right-hand side can be further bounded as:
    \begin{align*}
        \langle \nabla  \ell_{s}(\wb_{s+1}) - &\nabla\tilde{\ell}_{s}(\wb_{s+1}) , \wb_{s+1} - \wb^\star  \rangle
        \\
        &= \langle \nabla  \ell_{s}(\wb_{s+1}) 
        - \nabla \ell_{s} (\wb_s') 
        - \nabla^2 \ell_{s} (\wb_s') (\wb_{s+1} - \wb_s')
        , \wb_{s+1} - \wb^\star
        \rangle
        \\
        &= \langle D^3 \ell_{s}(\bar{\wb}_{s}) [\wb_{s+1} - \wb_s'] (\wb_{s+1} - \wb_s')
        , \wb_{s+1} - \wb^\star \rangle
        & \tag*{(Taylor expansion)}
        \\
        &= D^3 \ell_{s}(\bar{\wb}_{s}) 
        [\wb_{s+1} - \wb^\star, \wb_{s+1} - \wb_s', \wb_{s+1} - \wb_s'],
        \numberthis \label{eq:lemma:online_parameter_gap_bound_1}
    \end{align*}
    where in the second equality, we use the Taylor expansion by introducing  $\bar{\wb}_{s}$, which is a convex combination of $\wb_{s+1}$ and $\wb_s'$.
    Recall that by the definition of loss (see Equation~\eqref{eq:equi_loss}), the loss $\ell_{s}(\bar{\wb}_{s})$ can be expressed as 
    $\bar{\ell}_s (\bar{\zb}_{s})$, where
    $\bar{\zb}_{s} = (x_{si}^\top \bar{\wb}_{s} )_{i \in S_s} \in \RR^{|S_s|}$.
    Moreover, let $\zb_{s+1} = (x_{si}^\top \wb_{s+1} )_{i \in S_s}$, $\zb_s' = (x_{si}^\top \wb_s' )_{i \in S_s}$, and  $\zb^{\star} = (x_{si}^\top \wb^{\star} )_{i \in S_s}$.
    Then,
    by simple calculus, we get 
     \begin{align*}
         D^3 \ell_{s}(\bar{\wb}_{s}) [\wb_{s+1} - \wb^\star, \wb_{s+1} - \wb_s', \wb_{s+1} - \wb_s']
        &= D^3 \bar{\ell}_s (\bar{\zb}_{s})
        [\zb_{s+1} - \zb^{\star}, \zb_{s+1} - \zb_s', \zb_{s+1} - \zb_s']
        \\
        &\leq 3 \sqrt{2} \| \zb_{s+1} - \zb^{\star} \|_{\infty} 
        \| \zb_{s+1} - \zb_s'\|^2_{\nabla^2 \bar{\ell}_s (\bar{\zb}_{s}) }
        \tag{Proposition~\ref{prop:equi_self_concordant_MNL}}
        \\
        &\leq  3 \sqrt{2} \max_{x \in \Xcal_s} |x^\top (\wb_{s+1} - \wb^\star )| 
        \| \zb_{s+1} - \zb_s'\|^2_{\nabla^2 \bar{\ell}_s (\bar{\zb}_{s}) }
        \\
        &\leq 3 \sqrt{2} \alpha 
        \| \zb_{s+1} - \zb_s'\|^2_{\nabla^2 \bar{\ell}_s (\bar{\zb}_{s}) }
        \\
        &=  3 \sqrt{2} \alpha 
        \| \wb_{s+1} - \wb_s' \|^2_{\nabla^2 \ell_s (\bar{\wb}_{s}) }
        \tag{Definition of $\bar{\ell}$}
        \\
        &\leq  3 \sqrt{2} \alpha 
        \| \wb_{s+1} - \wb_s' \|^2_2,
        \numberthis \label{eq:lemma:online_parameter_gap_bound_eq_loss_2}
    \end{align*}
    where the last inequality holds because 
    \begin{align*}
        \nabla^2 \ell_{s}(\bar{\wb}_{s} )
        &= \sum_{i \in S_{s}} p_{s}(i | S_{s}, \bar{\wb}_{s}) x_{si} x_{si}^\top 
        -  \sum_{i \in S_{s}}  \sum_{j \in S_{s}} p_{s}(i | S_{s}, \bar{\wb}_{s}) p_{s}(j | S_{s}, \bar{\wb}_{s}) x_{si} x_{sj}^\top
        \\
        &= \sum_{i \in S_{s}} p_{s}(i | S_{s}, \bar{\wb}_{s}) x_{si} x_{si}^\top 
        - \left[ \sum_{i \in S_{s}}  p_{s}(i | S_{s}, \bar{\wb}_{s}) x_{si}  \right]
        \left[ \sum_{i \in S_{s}}  p_{s}(i | S_{s}, \bar{\wb}_{s}) x_{si}  \right]^\top 
        \\
        & \preceq \sum_{i \in S_{s}} p_{s}(i | S_{s}, \bar{\wb}_{s}) x_{si} x_{si}^\top
        \preceq \Ib_d
        \tag{$\|x_{si}\|_2 \leq 1$, Assumption~\ref{assum:bounded_assumption}}
        .
    \end{align*}
    Hence, by plugging~\eqref{eq:lemma:online_parameter_gap_bound_1} and ~\eqref{eq:lemma:online_parameter_gap_bound_eq_loss_2} into~\eqref{eq:lemma:online_parameter_gap_bound_mid}, and summing over $s \in \Tcal_{t + 1}$, we obtain
    \begin{align*}
        \sum_{s \in \Tcal_{t+1}}
        &\ell_{s}(\wb_{s+1}) 
        - \sum_{s \in \Tcal_{t+ 1}} \ell_{s}(\wb^\star)
        \\
        &\leq 
         3 \sqrt{2} \alpha 
           \sum_{s \in \Tcal_{t+1}}  \| \wb_{s+1} - \wb_s'\|^2_2
        + \frac{1}{2 \eta} \sum_{s \in \Tcal_{t+1}} \left( 
        \| \wb_s - \wb^\star \|_{H_s}^2 
        - \| \wb_{s+1} - \wb^\star \|_{H_{s+1}}^2 
        - \| \wb_{s+1} - \wb_s' \|_{H_s}^2  \right)
        \\
        &= 3 \sqrt{2} \alpha 
           \sum_{s \in \Tcal_{t+1}}  \| \wb_{s+1} - \wb_s'\|^2_2
        + \frac{1}{2 \eta} \left( 
        \| \wb_{t_1} - \wb^\star \|_{H_{t_1}}^2 
        - \| \wb_{t+1} - \wb^\star \|_{H_{t + 1}}^2 
        -  \sum_{s \in \Tcal_{t+1}}  \| \wb_{s+1} - \wb_s' \|_{H_s}^2  \right)
        ,
    \end{align*}
    where in the equality, $t_1 \in \Tcal$ represents the first update round. 
    Additionally, we use the fact that the parameter  $\wb_t$, and the matrices $\tilde{H}_{t}$ and $H_{t} $ remain unchanged during non-update rounds.
    By rearranging the terms and using the fact that $\| \wb_{t_1} - \wb^\star \|_{H_{t_1}}^2 \leq \lambda  \| \wb_{t_1} - \wb^\star \|_2^2 \leq 4 B^2 \lambda$,  we get
    \begin{align*}
        &\| \wb_{t+1} - \wb^\star \|_{H_{t + 1}}^2 
        \\
        &\leq 2 \eta 
        \left(
            \sum_{s \in \Tcal_{t+ 1}} \ell_{s}(\wb^\star)
            - \sum_{s \in \Tcal_{t+1}}
        \ell_{s}(\wb_{s+1}) 
        \right)
        + 4 B^2 \lambda 
        -  \sum_{s \in \Tcal_{t+1}}  \| \wb_{s+1} - \wb_s' \|_{H_s}^2  
        + 6 \sqrt{2} \eta \alpha 
           \sum_{s \in \Tcal_{t+1}}  \| \wb_{s+1} - \wb_s'\|^2_2
        \\
        &\leq  2 \eta 
        \left(
            \sum_{s \in \Tcal_{t+ 1}} \ell_{s}(\wb^\star)
            - \sum_{s \in \Tcal_{t+1}}
        \ell_{s}(\wb_{s+1}) 
        \right)
        + 4 B^2 \lambda 
        -   \frac{1}{2} \sum_{s \in \Tcal_{t+1}}  \| \wb_{s+1} - \wb_s' \|_{H_s}^2  
        ,
    \end{align*}
    where the last inequality holds because, by  setting $\lambda \geq 12 \sqrt{2} \eta \alpha$, we have $ 6 \sqrt{2} \eta \alpha  \| \wb_{s+1} - \wb_s' \|_2^2 \leq \frac{1}{2} \| \wb_{s+1} - \wb_s' \|_{H_s}^2  $.
\end{proof}
\subsubsection{Proof of Lemma~\ref{lemma:online_regret_intermediate_(a)}  }
\label{app_subsubsec:proof_lemma:online_regret_intermediate_(a)}
\begin{proof}[Proof of Lemma~\ref{lemma:online_regret_intermediate_(a)}]
Recall the definition of $\tilde{\zb}_t$ in Equation~\eqref{eq:def_intermediary_param}.
Then, by definition of the pseudo-inverse function $\sigmab^{+}_t$, we have $\sigmab_t\left( \tilde{\zb}_t\right) = \EE_{\wb \sim P_t} \left[\sigmab_t\left( (x_{tj}^\top \wb)_{j \in S_t} \right) \right]$.
Let $i_t \in S_t \cup \{0\}$ denote the (random) index such that 
  $y_{t i_t} = 1$;
in other words, $i_t$ is the item selected in round  $t$.
Then, we can express $\exp \left(\bar{\ell}_t(\tilde{\zb}_t) \right)$ as follows:
\begin{align*}
    \exp\left( -  \bar{\ell}_t(\tilde{\zb}_t) \right)
    = \exp\left( \log 
    \left(
        \left[ \sigmab_t\left( \tilde{\zb}_t\right) \right]_{i_t}
    \right)
    \right)
    = \left[ \sigmab_t\left( \tilde{\zb}_s\right) \right]_{i_t}
    = 
    \Big[\EE_{\wb \sim P_t} \left[\sigmab_t\left( (x_{tj}^\top \wb)_{j \in S_t} \right) \right]
    \Big]_{i_t}
    \numberthis
    \label{eq:inter_param_exp_loss}
    .
\end{align*}
Similarly, denoting $\zb_{t}^\star = \left(x_{tj}^\top \wb^\star \right)_{j \in S_t} \in \RR^{K_t}$, we can also express 
$\exp \left( 
    -   \ell_{t}(\wb^\star)
\right)$ as follows:
\begin{align*}
    \exp\left(  -   \ell_{t}(\wb^\star) \right)
    = \exp\left(  -   \bar{\ell}_{t}( \zb^\star_t ) \right)
    = \exp\left( \log 
    \left(
        \left[ \sigmab_t\left( \zb^\star_t\right) \right]_{i_t}
    \right)
    \right)
    = \left[ \sigmab_t\left( \zb^\star_t\right) \right]_{i_t}.
    \numberthis
    \label{eq:true_param_exp_loss}
\end{align*}
Here, note that $ \left[ \sigmab_t\left( \zb^\star_t\right) \right]_{i_t} = p_t(i_t | S_t, \wb^\star)$.

Now, we define 
\begin{align*}
   A_t :=
   \exp \left( 
        \sum_{s \in \Tcal_{t+ 1}} \ell_{s}(\wb^\star)
        - \sum_{s \in \Tcal_{t+ 1}} \bar{\ell}_s(\tilde{\zb}_s)
        \right).
\end{align*}
For completeness, we define  $A_0 := 1$.
First, we show that $(A_t)_{t \geq 0}$ is a nonnegative supermartingale with respect to the filtration $\Fcal_t = \sigma \left( S_1, \yb_1, \dots, S_t \right)$.

If $t$ is an update round, 
 we have:
\begin{align*}
    A_t 
    &= 
    \frac{ 
        \prod_{s \in \Tcal_{t+ 1}} \exp \left( 
            -  \bar{\ell}_s(\tilde{\zb}_s) 
        \right) 
    }
    {
         \prod_{s \in \Tcal_{t+ 1}} \exp \left( 
            -   \ell_{s}(\wb^\star)
        \right) 
    }
    =
    A_{t-1}
    \frac{ 
        \exp \left( 
            -  \bar{\ell}_t(\tilde{\zb}_t) 
        \right) 
    }
    {
         \exp \left( 
            -   \ell_{t}(\wb^\star)
        \right) 
    } 
    = 
    A_{t-1}
    \frac{ 
        \Big[\EE_{\wb \sim P_t} \left[\sigmab_t\left( (x_{tj}^\top \wb)_{j \in S_t} \right) \right]
    \Big]_{i_t}
    }
    {
        \left[ \sigmab_t\left( \zb^\star_t \right) \right]_{i_t}
    }.
    \tag{Eqn.~\eqref{eq:inter_param_exp_loss} and~\eqref{eq:true_param_exp_loss}}
\end{align*}
Note that the term $
        \Big[\EE_{\wb \sim P_t} \left[\sigmab_t\left( (x_{tj}^\top \wb)_{j \in S_t} \right) \right]
    \Big]_{i_t}
    /    \left[ \sigmab_t\left( \zb^\star_t \right) \right]_{i_t}
    $ is $\Fcal_{t+1}$-measurable.
Further, we get
\begin{align*}
    \EE \left[ 
        A_t \mid \Fcal_t
    \right]
    &=  A_{t-1}
    \cdot 
    \EE \left[ 
         \frac{ 
        \Big[\EE_{\wb \sim P_t} \left[\sigmab_t\left( (x_{tj}^\top \wb)_{j \in S_t} \right) \right]
        \Big]_{i_t}
        }
        {
            \left[ \sigmab_t\left( \zb^\star_t \right) \right]_{i_t}
        }  \,\,\middle|\,\, \Fcal_t
    \right]
    \\
    &= A_{t-1}
    \cdot 
    \sum_{i \in S_t \cup \{0\}}
     p_t(i | S_t, \wb^\star)
     \frac{ 
        \Big[\EE_{\wb \sim P_t} \left[\sigmab_t\left( (x_{tj}^\top \wb)_{j \in S_t} \right) \right]
        \Big]_{i}
        }
        {
            \left[ \sigmab_t\left( \zb^\star_t \right) \right]_{i}
        } 
    \\
    &= A_{t-1}
    \cdot 
    \sum_{i \in S_t \cup \{0\}}
     \left[ \sigmab_t\left( \zb^\star_t \right) \right]_{i}
     \frac{ 
        \Big[\EE_{\wb \sim P_t} \left[\sigmab_t\left( (x_{tj}^\top \wb)_{j \in S_t} \right) \right]
        \Big]_{i}
        }
        {
            \left[ \sigmab_t\left( \zb^\star_t \right) \right]_{i}
        } 
    \tag{$\left[ \sigmab_t\left( \zb^\star_t\right) \right]_{i} = p_t(i | S_t, \wb^\star)$}
    \\
    &= A_{t-1}
    \cdot 
    \sum_{i \in S_t \cup \{0\}}
        \Big[\EE_{\wb \sim P_t} \left[\sigmab_t\left( (x_{tj}^\top \wb)_{j \in S_t} \right) \right]
        \Big]_{i}
    = 
    A_{t-1}.
\end{align*}
Moreover, if $t$ is not an update round, we simply set $A_{t} = A_{t-1}$.
It follows that $(A_t)_{t \geq 0}$ is indeed a martingale, and hence also a supermartingale.
Since $(A_t)_{t \geq 0}$ is a nonnegative supermartingale, we can apply Ville’s inequality~\citep{ville1939etude} to obtain:
\begin{align*}
    \PP
     \left(
         \sum_{s \in \Tcal_{t+ 1}} \ell_{s}(\wb^\star)
        - \sum_{s \in \Tcal_{t+ 1}} \bar{\ell}_s(\tilde{\zb}_s)
          \geq \log \frac{1}{\delta}
    \right)
    &= 
    \PP
     \left(
         A_t \geq \frac{1}{\delta}
    \right)
    \\
    &\leq 
    \PP
    \left(
        \sup_{t \geq 0} A_t \geq \frac{1}{\delta}
    \right)
    \\
    &\leq 
    \EE [A_0] \delta
    \tag{Ville’s inequality}
    \\
    &= \delta,
    \tag{$A_0 =1$}
\end{align*}
which concludes the proof.
\end{proof}
\subsubsection{Proof of Lemma~\ref{lemma:bound_(b)}  }
\label{app_subsubsec:proof_lemma:bound_(b)}
\begin{proof}[Proof of Lemma~\ref{lemma:bound_(b)}]
The proof closely follows Lemma F.3 of~\citet{lee2024nearly} (or Lemma 14 of~\citet{zhang2024online}), with the only difference being that the summation is taken over the subset of rounds $\Tcal_{t+1} \subseteq [t]$, rather than the full set of rounds $[t]$.
For completeness, we include the full proof below.

To begin with, the proof builds on an observation from Proposition 2 of~\citet{foster2018logistic}, which states that $\tilde{\zb}_s$ serves as an aggregation forecaster for the logistic function.
Accordingly, for any $s \in \Tcal_{t+1}$, the following holds:
\begin{align}
    \bar{\ell}_s(\tilde{\zb}_{s})
    \leq -\log \left( \EE_{\wb \sim  P_s}\left[ e^{-\ell_{s}(\wb)} \right] \right)
    = -\log \left( \frac{1}{Z_{s}} \int_{ \RR^d } e^{-L_{s}(\wb)} \dd \wb \right),
    \label{eq:ztilde_loss_upper_ln}
\end{align}
where $L_{s}(\wb) := \ell_{s}(\wb) + \frac{1}{2c} \| \wb - \wb_s' \|_{H_s}^2$ and 
$Z_{s} := \int_{\RR^d } e^{ -\frac{1}{2c} \| \wb - \wb_s' \|_{H_s}^2  } \dd \wb $.

We define the the quadratic approximation of $L_{s}(\wb)$ as follows:
\begin{align*}
    \tilde{L}_{s}(\wb) := L_{s}(\wb_{s+1}) 
    + \langle \nabla L_{s}(\wb_{s+1}) , \wb - \wb_{s+1} \rangle
    + \frac{1}{2c}\| \wb - \wb_{s+1}\|_{H_s}^2.
\end{align*}
Then, by Lemma~\ref{lemma:zhang_lemma18} and considering the fact that
$\ell_{s}$ is $3\sqrt{2}$-self-concordant-like function by Proposition~\ref{prop:self_concordant},  we get
\begin{align}
      L_{s}(\wb) \leq \tilde{L}_{s}(\wb) + e^{18\| \wb - \wb_{s+1}\|_2^2 } \| \wb - \wb_{s+1} \|_{\nabla^2 \ell_{s}(\wb_{s+1})}^2.
      \label{eq:L_s_upperbound}
\end{align}
Furthermore, we define the function $\tilde{f}_{s+1}: \mathcal{W} \rightarrow \RR$ as
\begin{align*}
    \tilde{f}_{s+1} (\wb)
    = \exp\left( -\frac{1}{2c} \| \wb - \wb_{s+1} \|_{H_s}^2 -  e^{18\| \wb - \wb_{s+1}\|_2^2 } \| \wb - \wb_{s+1} \|_{\nabla^2 \ell_{s}(\wb_{s+1})}^2   \right).
\end{align*}
Then, we can then derive a lower bound for the expectation in Equation~\eqref{eq:ztilde_loss_upper_ln} as follows:
\begin{align*}
    \EE_{\wb \sim  P_s}\left[ e^{-\ell_{s}(\wb)} \right]
    &= \frac{1}{Z_{s}} \int_{ \RR^d } \exp(-L_{s}(\wb)) \dd \wb
    \\
    &\geq \frac{1}{Z_{s}}  \int_{ \RR^d } \exp(-\tilde{L}_{s}(\wb) - e^{18\| \wb - \wb_{s+1}\|_2^2 } \| \wb - \wb_{s+1} \|_{\nabla \ell_{s}(\wb_{s+1})}^2 ) \dd \wb
    \tag{Eqn.~\eqref{eq:L_s_upperbound}}
    \\
    &= \frac{\exp(-L_{s}(\wb_{s+1}) )}{Z_{s}}   \int_{ \RR^d } 
    \tilde{f}_{s+1}(\wb) 
    \cdot \exp(-\langle \nabla L_{s} (\wb_{s+1}), \wb - \wb_{s+1} \rangle) \dd \wb
    \tag{Definition of $\tilde{f}_{s+1}(\wb)$}
    .
\end{align*}
Moreover, we define $\tilde{Z}_{s+1} = \int_{\RR^d } \tilde{f}_{s+1} (\wb)  \dd \wb < + \infty$, and  denote the distribution whose density function is $\tilde{f}_{s+1} (\wb)/\tilde{Z}_{s+1} $ as $\tilde{P}_{s+1}$.
Then, we have
\begin{align*}
    \EE_{\wb \sim  P_s}\left[ e^{-\ell_{s}(\wb)} \right]
    &\geq \frac{\exp(-L_{s}(\wb_{s+1}) ) \tilde{Z}_{s+1}}{Z_{s}} \EE_{\wb \sim \tilde{P}_{s+1}} \left[ \exp(-\langle \nabla L_{s} (\wb_{s+1}), \wb - \wb_{s+1} \rangle) \right]
    \\
    &\geq \frac{\exp(-L_{s}(\wb_{s+1}) ) \tilde{Z}_{s+1}}{Z_{s}}
    \exp \bigg( - 
    \underbrace{\EE_{\wb \sim \tilde{P}_{s+1}} \left[\langle \nabla L_{s} (\wb_{s+1}), \wb - \wb_{s+1} \rangle \right]}_{= 0}   \bigg) 
    \tag{Jensen's inequality}
    \\
    &= \frac{\exp(-L_{s}(\wb_{s+1}) ) \tilde{Z}_{s+1}}{Z_{s}},
    \numberthis \label{eq:exp_loss_lowerbound2}
\end{align*}
where the equality holds because $\tilde{P}_{s+1}$ is symmetric around $\wb_{s+1}$.
Plugging~\eqref{eq:exp_loss_lowerbound2} into~\eqref{eq:ztilde_loss_upper_ln}, we get
\begin{align*}
    \bar{\ell}_s(\tilde{\zb}_{s})
    \leq \ell_{s}(\wb_{s+1}) + \frac{1}{2c} \|  \wb_s' - \wb_{s+1} \|_{H_s}^2
    + \log Z_{s}
    - \log \tilde{Z}_{s+1}
    .
\end{align*}
We can further bound the term $- \log \tilde{Z}_{s+1}$ as follows:
\begin{align*}
    - \log \tilde{Z}_{s+1}
    &= - \log 
    \left(
        \int_{\RR^d }
        \exp\left( -\frac{1}{2c} \| \wb - \wb_{s+1} \|_{H_s}^2 -  e^{18\| \wb - \wb_{s+1}\|_2^2 } \| \wb - \wb_{s+1} \|_{\nabla^2 \ell_{s}(\wb_{s+1})}^2   \right) \dd \wb 
    \right)
    \\
    &= -\log \left( \widehat{Z}_{s+1} \cdot \EE_{\wb \sim \widehat{P}_{s+1} } \left[ \exp \left(  -  e^{18\| \wb - \wb_{s+1}\|_2^2 } \| \wb - \wb_{s+1} \|_{\nabla^2 \ell_{s}(\wb_{s+1})}^2  \right)  \right]  \right)
    \\
    &\leq - \log \widehat{Z}_{s+1}
    + \EE_{\wb \sim \widehat{P}_{s+1} }  \left[  e^{18\| \wb - \wb_{s+1}\|_2^2 } \| \wb - \wb_{s+1} \|_{\nabla^2 \ell_{s}(\wb_{s+1})}^2  \right],
    \tag{Jensen’s inequality}
\end{align*}
where in the second equality, we define
$\widehat{P}_{s+1}  = \mathcal{N} (\wb_{s+1}, c H_s^{-1} )$ and
$\widehat{Z}_{s+1} := \int_{ \RR^d}e^{ -\frac{1}{2c} \| \wb - \wb_{s+1} \|_{H_s}^2  } \dd \wb$.
Hence, we get
\begin{align*}
    \bar{\ell}_s(\tilde{\zb}_{s})
    \leq \ell_{s}(\wb_{s+1}) + \frac{1}{2c} \|  \wb_s' - \wb_{s+1} \|_{H_s}^2
    + \log \frac{Z_{s}}{\widehat{Z}_{s+1}}
    + \EE_{\wb \sim \widehat{P}_{s+1} }  \left[  e^{18\| \wb - \wb_{s+1}\|_2^2 } \| \wb - \wb_{s+1} \|_{\nabla^2 \ell_{s}(\wb_{s+1})}^2  \right]
    \numberthis
    \label{eq:ztilde_loss_upper_ln_intermediate}
\end{align*}
To bound $\frac{Z_{s}}{\widehat{Z}_{s+1}}$ in Equation~\eqref{eq:ztilde_loss_upper_ln_intermediate}, we have
\begin{align*}
    \frac{Z_{s}}{\widehat{Z}_{s+1}}
    &= \frac{ 
        \int_{ \RR^d }
         e^{ -\frac{1}{2c} \| \wb - \wb_s' \|_{H_s}^2  } \dd \wb
    }{
        \int_{ \RR^d }
         e^{ -\frac{1}{2c} \| \wb - \wb_{s+1} \|_{H_s}^2  } \dd \wb
    }
    =
    \frac{ 
        \int_{ \RR^d }
         e^{ -\frac{1}{2c} \| \wb \|_{H_s}^2  } \dd \wb
    }{
        \int_{ \RR^d }
         e^{ -\frac{1}{2c} \| \wb \|_{H_s}^2  } \dd \wb
    }
    = 1,
\end{align*}
which indicates that
\begin{align*}
    \log \frac{Z_{s}}{\widehat{Z}_{s+1}}
    = 0.
    \numberthis
    \label{eq:ztilde_loss_upper_ln_intermediate_Z_ratio}
\end{align*}
Now, we bound the last term in Equation~\eqref{eq:ztilde_loss_upper_ln_intermediate}.
Using the Cauchy-Schwarz inequality, we have
\begin{align*}
    \EE_{\wb \sim \widehat{P}_{s+1} }  &\left[  e^{18\| \wb - \wb_{s+1}\|_2^2 } \| \wb - \wb_{s+1} \|_{\nabla^2 \ell_{s}(\wb_{s+1})}^2  \right]
    \\
    &\leq \underbrace{\sqrt{ \EE_{\wb \sim \widehat{P}_{s+1} }  \left[  e^{36\| \wb - \wb_{s+1}\|_2^2 }\right] } }_{\texttt{(b)-1}}
    \underbrace{\sqrt{  \EE_{\wb \sim \widehat{P}_{s+1} }  \left[  \| \wb - \wb_{s+1} \|_{\nabla^2 \ell_{s}(\wb_{s+1})}^4  \right] }}_{\texttt{(b)-2}}
    \numberthis \label{eq:expectation_decomposition}
    .
\end{align*}
Then, there exist orthogonal bases $\eb_1, \dots, \eb_d \in \RR^d$ such that $\wb - \wb_{s+1}$ follows the same distribution as $\widehat{P}_{s+1}$, and can be expressed as:
\begin{align}
    \sum_{j=1}^d \sqrt{c \lambda_j \left( H_s^{-1} \right)} X_j \eb_j,
    \quad
    \text{where }
    X_j \stackrel{i.i.d.}{\sim} \mathcal{N}(0,1), 
    \forall j \in [d],
    \label{eq:basis_expression}
\end{align}
and $\lambda_j \left(  H_s^{-1} \right)$  denotes the $j$-th largest eigenvalue of $H_s^{-1}$.
Now, we bound the term \texttt{(b)-1} in~\eqref{eq:expectation_decomposition} as follows:
\begin{align*}
    \sqrt{ \EE_{\wb \sim \widehat{P}_{s+1} }  \left[  e^{36\| \wb - \wb_{s+1}\|_2^2 }\right] }
    &= \sqrt{\EE_{X_j} \left[  \prod_{j=1}^d e^{36c \lambda_j \left( H_s^{-1} \right) X_j^2}  \right] }
    \\
    &\leq \sqrt{ \prod_{j=1}^d\EE_{X_j} \left[   e^{ 36 c/ \lambda X_j^2}  \right] }
    \tag{$c \lambda_j \left( H_s^{-1} \right) \leq c /\lambda$}
    \\
    &= \left( \EE_{X \sim \chi^2} \left[ e^{36c/ \lambda X} \right] \right)^{\frac{d}{2}}
    \leq  \EE_{X \sim \chi^2} \left[ e^{ 18cd/ \lambda X} \right]
    \tag{Jensen’s inequality}
    ,
\end{align*}
where $ \chi^2$ represents the chi-square distribution.
By setting $\lambda \geq 72 c d$, we get
\begin{align}
    \sqrt{ \EE_{\wb \sim \widehat{P}_{s+1} }  \left[  e^{36\| \wb - \wb_{s+1}\|_2^2 }\right] }
    \leq \EE_{X \sim \chi^2} \left[ e^{\frac{X}{4}} \right] 
    \leq \sqrt{2},
    \label{eq:term (a)-1}
\end{align}  
where the last inequality holds because the moment-generating function of the  $\chi^2$-distribution satisfies $\EE_{X \sim \chi^2} [e^{tX}] \leq 1/\sqrt{1-2t} $ for all $t \leq 1/2$.

To bound the term \texttt{(b)-2} in~\eqref{eq:expectation_decomposition}, let  $M_{s} = ( \nabla^2 \ell_{s}(\wb_{s+1}) )^{-1/2} H_s (\nabla^2 \ell_{s}(\wb_{s+1}))^{-1/2}$
and
$\lambda'_j = \lambda_j \left( c  M_{s}^{-1} \right)$ be the $j$-th largest eigenvalue of the matrix $c M_{s}^{-1}$.
Then, we have
\begin{align*}
    \sqrt{  \EE_{\wb \sim \widehat{P}_{s+1} }  \left[  \| \wb - \wb_{s+1} \|_{\nabla^2 \ell_{s}(\wb_{s+1})}^4  \right] }
    &= \sqrt{  \EE_{\wb \sim  \mathcal{N}(0, c H_s^{-1}) }  \left[  \| \wb  \|_{\nabla^2 \ell_{s}(\wb_{s+1})}^4  \right] }
    = \sqrt{  \EE_{\wb \sim  \mathcal{N}(0, c  M_{s}^{-1}) }  \left[  \| \wb  \|_2^4  \right] }.
\end{align*}
Furthermore, by performing an analysis similar to that in Equation~\eqref{eq:basis_expression}, we obtain
\begin{align*}
    \sqrt{  \EE_{\wb \sim  \mathcal{N}(0, c  M_{s}^{-1}) }  \left[  \| \wb  \|_2^4  \right] }
    &= \sqrt{\EE_{X_j \sim \mathcal{N}(0,1)} \left[ \left\| \sum_{j=1}^d \sqrt{\lambda'_j} X_j \eb_j  \right\|_2^4 \right] }
    = \sqrt{\EE_{X_j \sim \mathcal{N}(0,1)} \left[ \left( \sum_{j=1}^d \lambda'_j X_j^2\right)^2 \right] }
    \\
    &= \sqrt{ \sum_{j=1}^d \sum_{j'=1}^d \lambda'_j \lambda'_{j'} \EE_{X_j, X_{j'} \sim \mathcal{N}(0,1)} \left[X_j^2 X_{j'}^2\right] }
    \\
    &\leq \sqrt{3 \sum_{j=1}^d \sum_{j'=1}^d \lambda'_j \lambda'_{j'}}
    \tag{$\EE_{X_j, X_{j'} \sim \mathcal{N}(0,1)} [X_j^2 X_{j'}^2] \leq 3$,  $\forall j, j' \in [d]$}
    \\
    &= \sqrt{3}c \operatorname{Tr} \left(M_{s}^{-1}\right)
    \tag{$\sum_{j=1}^d \lambda'_j 
    = \operatorname{Tr}\left(c M_{s}^{-1} \right)$}
    .
\end{align*}
Here, $\operatorname{Tr}(A)$ denotes the trace of the matrix $A$.

Define the matrix $Q_{s+1} := \frac{\lambda}{2} \Ib_d 
+ \sum_{s' \in \Tcal_{s + 1} } \nabla^2 \ell_{s'} (\wb_{s'+1})$.
By setting $\lambda \geq 2$, we can ensure that $\nabla^2 \ell_{s} (\wb_{s+1}) \preceq \Ib_d \leq \frac{\lambda}{2} \Ib_d $.
As a result, we have $ Q_{s+1} \preceq \lambda \Ib_d + \sum_{s' \in \Tcal_s} \nabla^2 \ell_{s'} (\wb_{s'+1}) 
= H_s$.
Using this relationship, we can bound the trace as follows:
\begin{align*}
    \operatorname{Tr} \left(M_{s}^{-1}\right)
     &= \operatorname{Tr}\left( 
        H_s^{-1} \nabla^2 \ell_s(\wb_{s+1}) 
     \right)
     \leq \operatorname{Tr}\left( 
        Q_{s+1}^{-1} \nabla^2 \ell_s(\wb_{s+1}) 
     \right)
     \\
     &= \operatorname{Tr}\left( 
        Q_{s+1}^{-1} 
        \left(
            Q_{s+1} - Q_s
        \right)
     \right)
     \leq \log \frac{\operatorname{det}(Q_{s+1})}{\operatorname{det}(Q_{s})},
\end{align*}
where in the last inequality, we apply Lemma 4.5 of~\citet{hazan2016introduction}.
Hence, we get
\begin{align}
    \sqrt{  \EE_{\wb \sim \widehat{P}_{s+1} }  \left[  \| \wb - \wb_{s+1} \|_{\nabla^2 \ell_{s}(\wb_{s+1})}^4  \right] } 
    \leq \sqrt{3} c \log \frac{\operatorname{det}(Q_{s+1})}{\operatorname{det}(Q_{s})}.
    \label{eq:term (a)-2}
\end{align}
By substituting \eqref{eq:term (a)-1} and \eqref{eq:term (a)-2} into \eqref{eq:expectation_decomposition}, combining the result with \eqref{eq:ztilde_loss_upper_ln_intermediate} and \eqref{eq:ztilde_loss_upper_ln_intermediate_Z_ratio}, and summing over $s \in \Tcal_{t+1}$, we obtain
\begin{align*}
    \sum_{s \in \Tcal_{t+1}}\bar{\ell}_s(\tilde{\zb}_{s}) 
    - \sum_{s \in \Tcal_{t+1}}\ell_{s}(\wb_{s+1}) 
    &\leq \frac{1}{2c} \sum_{s \in \Tcal_{t+1}} \|  \wb_s' - \wb_{s+1} \|_{H_s}^2
    +  \sqrt{6} c \sum_{s \in \Tcal_{t+1}} \log \frac{\operatorname{det}(Q_{s+1})}{\operatorname{det}(Q_{s})}
    \\
    &= \frac{1}{2c} \sum_{s \in \Tcal_{t+1}} \|  \wb_s' - \wb_{s+1} \|_{H_s}^2
    +  \sqrt{6} c \log \frac{\operatorname{det}(Q_{t+1})}{\operatorname{det}\left(\frac{\lambda}{2} \Ib_d\right)}
    \\
    &\leq \frac{1}{2c} \sum_{s \in \Tcal_{t+1}} \|  \wb_s' - \wb_{s+1} \|_{H_s}^2
    +  \sqrt{6} c d \log \left( t + 2\right)
    .
\end{align*}
This concludes the proof.
\end{proof}

\subsection{Technical Lemmas for Theorem~\ref{thm:online_confidence_set}} 
\label{app_sec:technical_lemmas_for_cs}
\begin{lemma}[Proposition 4.1 of~\citealt{campolongo2020temporal}]
\label{lemma:loss_firstorder_decomposition}
    Let the $\wb_{t+1}$ be the solution of the update rule
    \[
    \wb_{t+1} = \arg\min_{\wb \in \mathcal{V}} \eta_t \ell_t(\wb) + D_{\psi}(\wb, \wb_t),
    \]
    where $\mathcal{V} \subseteq \mathcal{W} \subseteq \mathbb{R}^d$ is a non-empty convex set and $D_{\psi}(\wb_1, \wb_2) = \psi(\wb_1) - \psi(\wb_2) - \langle \nabla\psi(\wb_2), \wb_1 - \wb_2 \rangle$ is the Bregman Divergence w.r.t. a strictly convex and continuously differentiable function $\psi : \mathcal{W} \rightarrow \mathbb{R}$. Further supposing $\psi(\wb)$ is $1$-strongly convex w.r.t. a certain norm $\|\cdot\|$ in $\mathcal{W}$, then there exists a $\gb'_t \in \partial\ell_t(\wb_{t+1})$ such that
    \begin{align*}
        \langle \eta_t \gb'_t, \wb_{t+1} - \ub \rangle 
        \leq \langle \nabla \psi(\wb_t) - \nabla \psi(\wb_{t+1}), \wb_{t+1} - \ub \rangle     
    \end{align*}
    for any $\ub \in \mathcal{W}$.
\end{lemma}
\begin{lemma} \label{lemma:second_order_loss}
For any $t \in [T]$ and $\wb, \wb' \in \RR^d$ such that $\max_{i \in S_t} |x_{ti}^\top (\wb - \wb') | \leq \alpha$, 
the multinomial logistic loss $\ell_t: \RR^{d} \rightarrow \RR$, defined in~\eqref{eq:loss}, satisfies the following property:
    \begin{align*}
        \ell_t(\wb)
         &\geq \ell_t(\wb')
         + \nabla \ell_t(\wb')^\top (\wb - \wb')
         + 
         \frac{1}{2 + 3\sqrt{2} \alpha }
         (\wb - \wb')^\top
         \nabla^2 \ell_t(\wb') 
         (\wb - \wb')
         .
    \end{align*}
\end{lemma}
\begin{proof} [Proof of Lemma~\ref{lemma:second_order_loss}]
    Recall that by definition (see Equation~\eqref{eq:equi_loss}), the loss $\ell_t(\wb)$ can be rewritten as $\bar{\ell}_t (\zb_t)$, where $\zb_t = (x_{ti}^\top \wb )_{i \in S_t} \in \RR^{|S_t|}$.
    Similarly, $\ell_t(\wb') = \bar{\ell}_t (\zb_t')$.
    Then, by a second order Taylor expansion, we have
     \begin{align*}
         \bar{\ell}_t(\zb_t) 
         &= \bar{\ell}_t(\zb_t')
         + \nabla \bar{\ell}_t(\zb_t')^\top \left(\zb_t - \zb_t' \right)
         +  \left(\zb_t - \zb_t' \right)^\top
         \left( \int_0^1 (1-s) \nabla^2 \bar{\ell}_t (\zb_t' + s (\zb_t - \zb_t') ) \dd s\right)
         \left(\zb_t - \zb_t' \right)
         \\
         &\geq 
         \bar{\ell}_t(\zb_t')
         + \nabla \bar{\ell}_t(\zb_t')^\top \left(\zb_t - \zb_t' \right)
         + 
         \frac{1}{2 + 3\sqrt{2} \| \zb_t - \zb_t' \|_\infty}
         \left(\zb_t - \zb_t' \right)^\top
         \nabla^2 \bar{\ell}_t (\zb_t')
         \left(\zb_t - \zb_t' \right),
         \numberthis \label{eq:lemma:second_order_loss_1}
     \end{align*}
     where the inequality holds by Proposition~\ref{prop:hessian_usedful}.
     Moreover, by definition, we know that
     \begin{align*}
         \nabla \bar{\ell}_t(\zb_t')^\top \left(\zb_t - \zb_t' \right)
         &= \nabla \ell_t(\wb')^\top (\wb - \wb'),
         \\
         \text{and\,\,}
         \left(\zb_t - \zb_t' \right)^\top
         \nabla^2 \bar{\ell}_t (\zb_t')
         \left(\zb_t - \zb_t' \right)
         &=(\wb - \wb')^\top
         \nabla^2 \ell_t(\wb') 
         (\wb - \wb').
     \end{align*}
     Hence, we can rewrite Equation \eqref{eq:lemma:second_order_loss_1} equivalently as follows:
     \begin{align*}
         \ell_t(\wb)
         &\geq \ell_t(\wb')
         + \nabla \ell_t(\wb')^\top (\wb - \wb')
         + 
         \frac{1}{2 + 3\sqrt{2} \max_{i \in S_t} |x_{ti}^\top (\wb - \wb') | }
         (\wb - \wb')^\top
         \nabla^2 \ell_t(\wb') 
         (\wb - \wb')
         \\
         &\geq \ell_t(\wb')
         + \nabla \ell_t(\wb')^\top (\wb - \wb')
         + 
         \frac{1}{2 + 3\sqrt{2} \alpha }
         (\wb - \wb')^\top
         \nabla^2 \ell_t(\wb') 
         (\wb - \wb')
         ,
     \end{align*}
     which concludes the proof.
\end{proof}
\begin{lemma}[Lemma 18 of \citealt{zhang2024online}] \label{lemma:zhang_lemma18}
    For any $H_t \succeq 0$,
    let $L_t(\wb) = \ell_t(\wb) + \frac{1}{2c} \| \wb - \wb_t \|_{H_t}^2$. 
    Assume that $\ell_t$ is a $M$-self-concordant-like function.
    Then, for any $\wb,  \wb_t \in \mathcal{W}$, the quadratic approximation $\tilde{L}_t(\wb) = L_t(\wb_{t+1}) 
            + \langle \nabla L_t(\wb_{t+1}) , \wb - \wb_{t+1} \rangle
            + \frac{1}{2c}\| \wb - \wb_{t+1}\|_{H_t}^2$ satisfies
    \begin{align*}
         L_t(\wb) \leq \tilde{L}_t(\wb) + e^{M^2\| \wb - \wb_{t+1}\|_2^2 } \| \wb - \wb_{t+1} \|_{\nabla \ell_t(\wb_{t+1})}^2.
    \end{align*}
\end{lemma}
%
%

\section{Proof of Theorem~\ref{thm:regret_main}} 
\label{app_sec:proof_thm_regret_main}
In this section, we present the proof of Theorem~\ref{thm:regret_main}.
To begin, we define a set of adaptive warm-up rounds as follows:
\begin{align*}
    \WarmupRounds 
    &:= \left\{
        t \in [T]:
        \max_{x \in \Xcal_t} \| x \|_{(\WarmupHessian_t)^{-1}}^2 \geq 1 / \Threshold_t^2
    \right\}, 
    \numberthis \label{eq:def_sets}
\end{align*} 
where we define the threshold $\Threshold_t$ as:
\begin{align*}
    \Threshold_t := 6\sqrt{2} \zeta_t(\delta) 
    = \BigO \left( B \sqrt{d \log (t/\delta)} + B^{3/2} \sqrt{d} + B^2 \right).
\end{align*}
Moreover, we define the following two confidence sets for all $t \in [T]$:
\begin{align*}
    \WarmupConfidenceSet_t (\delta)
    &:= \left\{
        \wb \in \RR^d 
        \mid 
        \|   \wb - \WarmupParam_t \|_{\WarmupHessian_t}
        \leq \zeta_t(\delta)
    \right\},
    \quad \text{and}
    \\
    \Ccal_t (\delta)
    &:= \left\{
        \wb \in \RR^d 
        \mid 
        \|  \wb - \wb_t \|_{H_t}
        \leq \beta_t(\delta)
    \right\}
    ,
    \numberthis \label{eq:CS_def}
\end{align*}
where
\begin{align*}
    \zeta_t (\delta)
    &:= 
    \sqrt{
         2   \WarmupStep
        \log \frac{1}{\delta}
        +  4\sqrt{6} (\WarmupStep)^2  d \log \left( t + 2\right)
        + 4 B^2 \WarmupRegualizer 
        }
        \\
    &= \BigO \left( B \sqrt{d \log (t/\delta)} +  B^{3/2} \sqrt{d} + B^2 \right)
    \tag{set $\alpha = 2 B$, $\WarmupStep = \frac{1}{2}(1 + 3\sqrt{2} \alpha)$, $\WarmupRegualizer = \max \{ 12 \sqrt{2}\WarmupStep\alpha, 144 \WarmupStep d, 2  \}$}
    , 
\end{align*}
and
\begin{align*}
    \beta_t(\delta) 
    &:= \sqrt{
    2   \eta
    \log \frac{1}{\delta}
    +  4\sqrt{6} \eta^2  d \log \left( t + 2\right)
    + 4 B^2 \lambda 
    }
    \\
    &= \BigO \left( 
    \sqrt{
        d \log ( t/\delta)
        }
        + B\sqrt{d}
    \right)
    \tag{set $\alpha = \frac{1}{3\sqrt{2}}$, $\eta = 1$, $\lambda = \max \{ 12 \sqrt{2}\eta\alpha, 144 \eta d, 2  \} = 144 d$}
    .
\end{align*}
Then, the true parameter $\wb^\star$ lies within both confidence sets with high probability.
\begin{corollary} [Confidence set for adaptive warm-up]
\label{cor:CS_warm-up}
    Let $\delta \in (0,1]$.
    We set $\WarmupStep = \frac{1}{2} (1+ 3\sqrt{2} B)$ and $\WarmupRegualizer =  \max \{ 12 \sqrt{2}\WarmupStep\alpha, 144 \WarmupStep d, 2  \}$.
    Then, we have 
    \begin{align*}
        \textup{Pr}[\forall t \geq 1, \wb^\star \in \WarmupConfidenceSet_t(\delta)] \geq 1- \delta.
    \end{align*}
\end{corollary}
The proof can be found in Appendix~\ref{app_subsubsec:proof_of_cor_CS-warm-up}.
\begin{corollary} [Restatement of Corollary~\ref{cor:CS_main}, Confidence set for planning \& learning]
\label{cor:CS}
    Let $\delta \in (0,1]$.
    We set $\eta =  1$,
    $\lambda =  144 d$, and 
    $\Threshold_t = 6 \sqrt{2} \zeta_t(\delta) = \BigO \left( B \sqrt{d \log (t/\delta)} + B^{3/2}\sqrt{d} + B^2 \right)$.
    Then, if $\wb^\star \in \WarmupConfidenceSet_t(\delta)$ for all $t \geq 1$, we have
    \begin{align*}
        \textup{Pr}
        \left[ \forall t \geq 1, \wb^\star \in \Ccal_t(\delta)
        \right] \geq 1- \delta.
    \end{align*}
\end{corollary}
The proof is provided in Appendix~\ref{app_subsubsec:proof_of_cor_CS}.

Furthermore, we introduce several useful lemmas.
Lemma~\ref{lemma:utility} shows that $\UCB_{ti}$ provides an optimistic estimate of the true utility. 
\begin{lemma} [Lemma E.1 of~\citealt{lee2024nearly}] \label{lemma:utility}
    Let $\UCB_{ti} = x_{ti}^\top \wb_t + \beta_t (\delta) \|  x_{ti}\|_{H_t^{-1}}$.
    Assume that $\wb^\star \in \mathcal{C}_t(\delta)$, where
    $\Ccal_t(\delta) 
        := \{ \wb \in \mathcal{W} \mid \| \wb_t - \wb \|_{H_t} \leq \beta_{t}(\delta) \}$.
    Then, we have
    \begin{align*}
        0 \leq \UCB_{ti} - x_{ti}^\top \wb^\star \leq 2 \beta_t(\delta) \| x_{ti} \|_{H_t^{-1}}.
    \end{align*}
\end{lemma}
Lemma~\ref{lemma:optimism}  shows that $\tilde{R}_{t}(S_t)$, defined in~\eqref{eq:opt_revenue}, is an upper bound of the true expected revenue of
the optimal assortment, $R_{t}(S_t^\star, \wb^\star) $.
\begin{lemma} [Optimism, Lemma 4 of~\citealt{oh2021multinomial}] \label{lemma:optimism}
Let 
$\tilde{R}_{t}(S) = \frac{\sum_{i \in S} \exp( \UCB_{ti} ) r_{ti} }{1 + \sum_{j \in S} \exp(\UCB_{tj}) }$.
And suppose $S_t = \argmax_{S \in \mathcal{S}} \tilde{R}_{t}(S)$.
If for every item $i \in S_t^\star$, $\UCB_{ti} \geq x_{ti}^\top \wb^\star$, then for all $t \geq 1$, the following inequalities hold:
    \begin{align*}
        R_{t}(S_{t}^\star, \wb^\star) \leq \tilde{R}_t(S_{t}^\star)
        \leq \tilde{R}_t(S_{t}).
    \end{align*}
\end{lemma}
It is important to note that Lemma~\ref{lemma:optimism} does not assert that the expected revenue is a monotonic function in general. 
Rather, it specifically states that the expected revenue associated with the ``optimal'' assortment increases as the MNL parameters increase~\citep{agrawal2019mnl, oh2021multinomial, lee2024nearly}.

Lemma~\ref{lemma:increasing_R} shows that $\tilde{R}_{t}(S_t)$ increases as the utility values of the items in $S_t$ further grow.
\begin{lemma} [Overly optimism, Lemma H.2 of~\citealt{lee2024nearly}]  \label{lemma:increasing_R}
    We define 
    $\tilde{R}_{t}(S) := \frac{\sum_{i \in S} \exp( \UCB_{ti} ) r_{ti} }{1 + \sum_{j \in S} \exp(\UCB_{tj}) }$ and $S_t = \argmax_{S \in \mathcal{S}} \tilde{R}_{t}(S)$.
    Assume $\widebar{\UCB}_{ti} \geq \UCB_{ti} \geq 0$ for all $i \in [N]$.
    Then, we have
    \begin{align*}
        \tilde{R}_{t}(S_t) 
        \leq \frac{\sum_{i \in S_t} \exp( \widebar{\UCB}_{ti} ) r_{ti} }{1 + \sum_{j \in S_t} \exp(\widebar{\UCB}_{tj}) }.
    \end{align*}
\end{lemma}
Moreover, we demonstrate that the rewards for the chosen assortment, $r_{ti}$ for all $i \in S_t$ satisfy the condition $R_t(S_t, \wb^\star)$.
\begin{lemma}
\label{lemma:r_geq_R}
For all round $t \in [T]$, we have
    \begin{align*}
         r_{ti}
         \geq R_{t}(S_{t}, \wb^\star),
         \quad \forall i \in S_t.
    \end{align*}
\end{lemma}
The proof is provided in Appendix~\ref{app_subsubsec:proof_of_lemma:r_geq_R}.

We introduce an elliptical potential lemma that will be used in our proof.
\begin{lemma} [Elliptical potential lemma]
\label{lemma:epl_H}
    Define $H_t(\wb) := \lambda \Ib_d + \sum_{s \in [t-1] \setminus \WarmupRounds } \nabla^2 \ell_s(\wb)$.
    If $\| x_{si} \|_{H_s(\wb)^{-1}}^2 \leq \frac{1}{2}$ for all $i \in S_s$ and $s \in [t] \setminus \WarmupRounds$, then we have
    \begin{align*}
        \sum_{s \in [t] \setminus \WarmupRounds } 
         \sum_{i \in S_s \cup \{0\}} 
        p_s(i | S_s, \wb) \left\| 
            x_{si} - \EE_{ j \sim p_s(\cdot | S_s, \wb) }[x_{sj}]
        \right\|_{( H_s(\wb) )^{-1}}^2
        \leq 2d \log \left( 1+ \frac{t}{d \lambda} \right).
    \end{align*}
\end{lemma}
The proof is deferred to Appendix~\ref{app_subsubsec:proof_of_lemma:epl_H}.

Lemma~\ref{lemma:Hessian_bound} shows that $H_t$ and $H_t(\wb^\star)$ remain similar when updated only for $t \notin \WarmupRounds$.
\begin{lemma} 
\label{lemma:Hessian_bound}
    Let
    $H_t= \lambda \Ib_d + \sum_{s \in [t-1] \setminus \WarmupRounds } \nabla^2 \ell_s(\wb_{s+1})$ and
    $H_t(\wb^\star) = \frac{\lambda}{e} \Ib_d + \sum_{s \in [t-1] \setminus \WarmupRounds } \nabla^2 \ell_s(\wb^\star)$.
    Then, we have
    \begin{align*}
        \frac{1}{e}  H_t(\wb^\star) 
        \preceq H_t 
        \preceq e  H_t(\wb^\star).
    \end{align*}
\end{lemma}
The proof is provided in Appendix~\ref{app_subsubsec:proof_of_lemma:Hessian_bound}.

Additionally, we present a useful lemma that will be employed to bound the second-order term of the regret.
\begin{lemma} [Lemma E.3 of~\citealt{lee2024nearly}]
\label{lemma:revenue_second_pd}
Define $Q:\RR^K \rightarrow \RR$, such that for any $\ub = (u_1, \dots, u_K) \in \RR^K$, $Q(\ub) = \sum_{i=1}^K \frac{\exp(u_i)}{1 + \sum_{k=1}^K \exp(u_k)}$.
Let $p_i(\ub) = \frac{\exp(u_i)}{1 + \sum_{k=1}^K \exp(u_k)}$.
Then, for all $i \in [K]$, we have
\begin{align*}
    \left| \frac{\partial^2 Q}{\partial i \partial j} \right|
    \leq
    \begin{cases}
        3 p_i(\ub) & \text{if} \,\,\, i=j,
        \\
        2p_i(\ub) p_j(\ub) & \text{if} \,\,\, i \neq j.
    \end{cases}
\end{align*}
\end{lemma}
The size of the set $\WarmupRounds $ is bounded as described in the following lemma:
\begin{lemma} 
\label{lemma:bound_T_w-T_0} 
    The size of the set $\WarmupRounds$, defined in Equation~\eqref{eq:def_sets}, is bounded as follows:
    \begin{align*}
         \left| \WarmupRounds  \right|
         \leq \frac{2}{\kappa} \Threshold_T^2 d \log \left( 1+ \frac{T}{d \lambda} \right).
    \end{align*}
\end{lemma}
The proof is deferred to Appendix~\ref{app_subsubsec:proof_of_lemma:bound_T_w-T_0}.

We are now ready to provide the proof of Theorem~\ref{thm:regret_main}.

\subsection{Main Proof of Theorem~\ref{thm:regret_main}} \label{app_subsec:main_proof_thm_regret_main}
\begin{proof} [Proof of Theorem~\ref{thm:regret_main}]
    Throughout the proof of the theorem, assume the following event holds:
    \begin{align*}
        \left\{ \forall t \geq 1, \wb^\star \in \WarmupConfidenceSet_t(\delta) \right\} \,\, 
        \mathsmaller{\bigcup} \,\,
    \left\{ \forall t \geq 1, \wb^\star \in \Ccal_t(\delta) \right\}, 
    \numberthis \label{eq:good_event}
    \end{align*}
    which occurs with a probability of at least $1- 2\delta$ by Corollary~\ref{cor:CS_warm-up} and~\ref{cor:CS}.

    From the definition of $\WarmupRounds$ (see Equation~\eqref{eq:def_sets}), we decompose the regret as follows:
    \begin{align*}
        \Regret (\wb^\star )
        &= 
        \sum_{t=1}^T  R_{t}(S_{t}^\star, \wb^\star) -  R_{t}(S_{t}, \wb^\star) 
        \\
        &= \sum_{t \in \WarmupRounds} 
        R_{t}(S_{t}^\star, \wb^\star) -  R_{t}(S_{t}, \wb^\star) 
        + \sum_{t \notin  \WarmupRounds }  R_{t}(S_{t}^\star, \wb^\star) -  R_{t}(S_{t}, \wb^\star) 
        \\
        &\leq \left| \WarmupRounds \right|
        + \sum_{t \notin   \WarmupRounds }  R_{t}(S_{t}^\star, \wb^\star) -  R_{t}(S_{t}, \wb^\star)
        \tag{$ R_{t}(S_{t}^\star, \wb^\star) -  R_{t}(S_{t}, \wb^\star) \leq 1$}
        \\
        &\leq 
        \frac{2}{\kappa} \Threshold_T^2 d \log \left( 1+ \frac{T}{d \lambda} \right)
        + \sum_{t \notin  \WarmupRounds }  R_{t}(S_{t}^\star, \wb^\star) -  R_{t}(S_{t}, \wb^\star),
        \numberthis \label{eq:proof_regret}
    \end{align*}
    where the last inequality holds by Lemma~\ref{lemma:bound_T_w-T_0}.
    Next, we concentrate on deriving a bound for the last term.
    We define $\widebar{\UCB}_{ti}$ as $\widebar{\UCB}_{ti} := x_{ti}^\top \wb^\star + 2 \beta_t(\delta) \| x_{ti} \|_{H_t^{-1}}$.
    Under the event in Equation~\eqref{eq:good_event},
    by Lemma~\ref{lemma:utility}, we have
    \begin{align*}
        \UCB_{ti} 
        \leq x_{ti}^\top \wb^\star + 2 \beta_t(\delta) \| x_{ti} \|_{H_t^{-1}}
        =: \widebar{\UCB}_{ti}.
    \end{align*}
    Then, we define the \textit{overly optimistic} expected revenue, $\dbtilde{R}_t(S_{t})$, as  
    \begin{align*}
        \dbtilde{R}_t(S_{t}) := \frac{\sum_{i \in S_t} \exp( \widebar{\UCB}_{ti} ) r_{ti} }{1 + \sum_{j \in S_t} \exp(\widebar{\UCB}_{tj}) }.
    \end{align*}
    Using this definition and applying the optimism lemmas, we can derive an upper bound for the regret as follows:
    \begin{align*}
        \sum_{t \notin   \WarmupRounds }   R_{t}(S_{t}^\star, \wb^\star) -  R_{t}(S_{t}, \wb^\star) 
        &\leq \sum_{t \notin   \WarmupRounds }   \tilde{R}_t(S_{t}) -  R_{t}(S_{t}, \wb^\star)  
        \tag{Lemma~\ref{lemma:optimism}}
        \\
        &\leq \sum_{t \notin  \WarmupRounds }   \dbtilde{R}_t(S_{t}) -  R_{t}(S_{t}, \wb^\star)
        .
        \tag{Lemma~\ref{lemma:increasing_R}}
    \end{align*}
    Now, we define a function  $\tilde{Q}:\RR^{|S_t|} \rightarrow \RR$, such that for all $\ub = (u_1, \dots, u_{|S_t|})^\top \in \RR^{|S_t|}$, $\tilde{Q}(\ub) = \sum_{k=1}^{|S_t|} \frac{\exp(u_k)r_{ti_k} }{1 + \sum_{j=1}^{|S_t|} \exp(u_j)}$.
    Here, we denote $S_t = \{i_1, \dots, i_{|S_t|} \}$ for simplicity.
    Additionally, let $\ub_t = (u_{ti_1}, \dots u_{ti_{|S_t|}} )^\top = (\widebar{\UCB}_{ti_1} , \dots, \widebar{\UCB}_{ti_{{|S_t|}}} )^\top$ 
    and $\ub^\star_t = (u_{ti_1}^\star, \dots u_{ti_{|S_t|}}^\star )^\top = (x_{ti_1}^\top \wb^\star, \dots, x_{ti_{{|S_t|}}}^\top \wb^\star)^\top$.

    Then, by a second order Taylor expansion, we derive
    \begin{align*}
        \sum_{t \notin   \WarmupRounds }  \dbtilde{R}_t(S_{t}) -  R_{t}(S_{t}, \wb^\star)  
        &= \sum_{t \notin   \WarmupRounds }  \tilde{Q}(\ub_t) - \tilde{Q}(\ub^\star_t) 
        \\
        &= \underbrace{  \sum_{t \notin  \WarmupRounds }  \nabla \tilde{Q}(\ub_t^\star)^\top (\ub_t - \ub^\star_t) }_{I_1}
            + \underbrace{\frac{1}{2} \sum_{t \notin   \WarmupRounds }  (\ub_t - \ub^\star_t)^\top \nabla^2 \tilde{Q}(\bar{\ub}_t) (\ub_t - \ub^\star_t)}_{I_2},
        \numberthis
        \label{eq:proof_regret_decompose_first}
    \end{align*}
    where $\bar{\ub}_t = (\bar{u}_{ti_1}, \dots, \bar{u}_{ti_{|S_t|}})^\top \in \RR^{|S_t|}$ is the convex combination of $\ub_t$ and $\ub^\star_t$.
    
    First, we bound the term $I_1$.
    \begin{align*}
        & \sum_{t \notin   \WarmupRounds } \nabla \tilde{Q}(\ub_t^\star)^\top (\ub_t - \ub^\star_t) 
        \\
        &=  \sum_{t \notin   \WarmupRounds } \sum_{i \in S_t} \frac{\exp(x_{ti}^\top \wb^\star )r_{ti}}{1 + \sum_{k\in S_t} \exp(x_{tk}^\top \wb^\star ) } (u_{ti} - u_{ti}^\star ) 
        -  \sum_{j \in S_t}   \frac{\exp(x_{tj}^\top \wb^\star ) r_{tj} \sum_{i \in S_t} \exp(x_{ti}^\top \wb^\star )}{(1 + \sum_{k\in S_t} \exp(x_{tk}^\top \wb^\star ))^2 } (u_{ti} - u_{ti}^\star )   
        \\
        &=  \sum_{t \notin   \WarmupRounds } \sum_{i \in S_t} p_t(i | S_t, \wb^\star) r_{ti} (u_{ti} - u_{ti}^\star )
        - \sum_{i \in S_t}  \sum_{j \in S_t}  p_t(i | S_t, \wb^\star) r_{ti} p_t(j | S_t, \wb^\star)  (u_{tj} - u_{tj}^\star )
        \\
        &=  \sum_{t \notin   \WarmupRounds } \sum_{i \in S_t} p_t(i | S_t, \wb^\star)  r_{ti} \left( (u_{ti} - u_{ti}^\star ) -  \sum_{j \in S_t} p_t(j | S_t, \wb^\star)  (u_{tj} - u_{tj}^\star )  \right)
        \\
        &=  \sum_{t \notin   \WarmupRounds }
        2 \beta_t(\delta)
        \underbrace{\sum_{i \in S_t} p_t(i | S_t, \wb^\star)  r_{ti} \left(  \| x_{ti} \|_{H_t^{-1}} -  \sum_{j \in S_t} p_t(j | S_t, \wb^\star)   \| x_{tj} \|_{H_t^{-1}} \right)}_{\geq 0}
        \\
        &\leq 2 \beta_T(\delta)  \sum_{t \notin   \WarmupRounds }  \sum_{i \in S_t} p_t(i | S_t, \wb^\star)  r_{ti} \left(  \| x_{ti} \|_{H_t^{-1}} -  \sum_{j \in S_t} p_t(j | S_t, \wb^\star) \| x_{tj} \|_{H_t^{-1}} \right)
        \tag{$\beta_t(\delta)$ is non-decreasing}
        ,
    \end{align*}
    where in the last inequality, we use the fact that $\beta_t(\delta)$ is non-decreasing and that the following holds:
    \begin{align*}
        \sum_{i \in S_t} &p_t(i | S_t, \wb^\star)  r_{ti} \bigg(  \| x_{ti} \|_{H_t^{-1}} -  \sum_{j \in S_t} p_t(j | S_t, \wb^\star)   \| x_{tj} \|_{H_t^{-1}} \bigg)
        \\
        &= \sum_{i \in S_t} p_t(i | S_t, \wb^\star) 
         \| x_{ti} \|_{H_t^{-1}}
         \bigg(
            r_{ti} - \underbrace{\sum_{j \in S_t} p_t(j | S_t, \wb^\star) r_{tj}}_{= R_t(S_t, \wb^\star)}
         \bigg)
        \geq 0.
        \tag{Lemma~\ref{lemma:r_geq_R}}
    \end{align*}
    Let $x_{t0} = \mathbf{0}$ and $r_{t0} = 0$. 
    For simplicity, we denote
    $\EE^{\wb}_{t}[x_{ti}] = \EE_{j \sim p_t(\cdot | S_t, \wb)}[x_{ti}]$,
    and
    $\EE^{\wb}_t[r_{ti}] = \EE_{j \sim p_t(\cdot | S_t, \wb)}[r_{ti}]$.
    Here, $\EE^{\wb}_t$ represents the expectation taken with respect to the distribution $p_t(\cdot | S_t, \wb)$.
    Note that $\EE^{\wb}_t[r_{ti}] = R_t(S_t, \wb)$.
    Then, we can rewrite the above inequality in the following form:
    \begin{align*}
        \sum_{t \notin   \WarmupRounds } 
        &\nabla \tilde{Q}(\ub_t^\star)^\top (\ub_t - \ub^\star_t) 
        \\
        &\leq
        2 \beta_T(\delta)  \sum_{t \notin   \WarmupRounds } 
        \left(
            \EE^{\wb^\star}_t \Big[
                r_{ti} \| x_{ti} \|_{H_t^{-1}}
            \Big]
            - 
            \EE^{\wb^\star}_t \Big[
                r_{ti} 
            \Big]
                \EE^{\wb^\star}_t \Big[ \| x_{tj} \|_{H_t^{-1}}  \Big]
        \right)
        \tag{$x_{t0} = \mathbf{0}$, $r_{t0} = 0$}
        \\
        &= 
        2 \beta_T(\delta)  \sum_{t \notin   \WarmupRounds } 
            \underbrace{\EE^{\wb^\star}_t 
            \left[
                \left(
                    r_{ti} - 
                    \EE^{\wb^\star}_t [r_{tj}]  
                \right) 
                \bigg(
                     \| x_{ti} \|_{H_t^{-1}} - 
                     \EE^{\wb^\star}_t \left[ \| x_{tj} \|_{H_t^{-1}}  \right] 
                \bigg)
                \right]
                }_{\text{Covariance between }  r_{ti} \text{ and } \| x_{ti} \|_{H_t^{-1}} \text{ given } S_t}.
            \numberthis 
            \label{eq:proof_regret_cov}
    \end{align*}
    By Lemma~\ref{lemma:r_geq_R}, we know that $r_{ti} \geq R_{t}(S_{t}, \wb^\star)
    = \EE^{\wb^\star}_t[r_{ti}]$ for all $i \in S_t$.
    Therefore, we can bound the term inside the expectation in~\eqref{eq:proof_regret_cov} as follows:
    \begin{align*}
         \Big(
            \underbrace{r_{ti} - 
            \EE^{\wb^\star}_t [r_{tj}]  }_{\geq 0 \text{ by Lemma~\ref{lemma:r_geq_R}}}
         \Big) 
         \left(
             \| x_{ti} \|_{H_t^{-1}} - 
             \EE^{\wb^\star}_t \left[ \| x_{tj} \|_{H_t^{-1}}  \right] 
         \right)
        &\leq 
        \left(
            r_{ti} - 
            \EE^{\wb^\star}_t [r_{tj}]  
        \right) 
        \left(
             \| x_{ti} \|_{H_t^{-1}} - 
              \| \EE^{\wb^\star}_t \left[ x_{tj} \right]  \|_{H_t^{-1}}  
        \right)
        \tag{Jensen's inequality}
        \\
        &\leq 
        \left(
            r_{ti} - 
            \EE^{\wb^\star}_t [r_{tj}]  
        \right) 
             \left\| x_{ti}  - 
            \EE^{\wb^\star}_t \left[ x_{tj} \right]  \right\|_{H_t^{-1}},
        \numberthis 
        \label{eq:eq:proof_regret_cov_bound}
    \end{align*}
    where the last inequality holds due to the fact that $\| \ab \|  = \| \ab - \bb + \bb\| \leq \| \ab-\bb \| + \|\bb\|$ for any vectors $\ab, \bb \in \RR^d$.
    Plugging~\eqref{eq:eq:proof_regret_cov_bound} into~\eqref{eq:proof_regret_cov}, we obtain
    \begin{align*}
        \sum_{t \notin   \WarmupRounds } 
        &\nabla \tilde{Q}(\ub_t^\star)^\top (\ub_t - \ub^\star_t) 
        \\
        &\leq
        2 \beta_T(\delta)  \sum_{t \notin   \WarmupRounds } 
        \EE^{\wb^\star}_t 
            \left[
                (
                    r_{ti} - 
                    \EE^{\wb^\star}_t [r_{tj}]  
                ) 
                     \left\| x_{ti}  - 
                    \EE^{\wb^\star}_t \left[ x_{tj} \right]  \right\|_{H_t^{-1}}
            \right] 
        \\
        &\leq 
        2 \beta_T(\delta)
        \usqrt{
            \sum_{t \notin   \WarmupRounds } 
            \ubrace{
            \EE^{\wb^\star}_t 
                \bigg[
                    \left(
                        r_{ti} - 
                        \EE^{\wb^\star}_t [r_{tj}]  
                    \right)^2
                \bigg]
            }{= \VV^{\wb^\star}_t[r_{ti}] =: \sigma_t^2 }
            }
        \sqrt{
            \sum_{t \notin   \WarmupRounds } 
            \EE^{\wb^\star}_t 
                \bigg[
                    \left\| x_{ti}  - 
                    \EE^{\wb^\star}_t \left[ x_{tj} \right]  \right\|_{H_t^{-1}}^2
                \bigg]
            }
        \tag{Cauchy-Schwartz inequality}
        \\
        &= 2 \beta_T(\delta)
        \sqrt{ \sum_{t \notin   \WarmupRounds } \sigma_t^2 
        }
        \sqrt{
            \sum_{t \notin   \WarmupRounds } 
            \sum_{i \in S_t \cup \{0 \} }
                p_t (i | S_t, \wb^\star)
                \left\| x_{ti}  - 
                    \EE^{\wb^\star}_t \left[ x_{tj} \right]  \right\|_{H_t^{-1}}^2
            }        
        \numberthis \label{eq:proof_regret_first_mid}
    \end{align*}
    where in the last equality, we define the variance of the rewards under 
    $\wb^\star$, given $S_t$, as $\sigma_t^2 := \VV^{\wb^\star}_t[r_{ti}] =  \EE^{\wb^\star}_t 
                \left[
                    \left(
                        r_{ti} - 
                        \EE^{\wb^\star}_t [r_{tj}]  
                    \right)^2
                \right]$.
    We define $H_t(\wb^\star) =  \frac{\lambda}{e}\Ib_d + \sum_{s \in [t-1] \setminus \WarmupRounds} \nabla^2 \ell_s(\wb^\star)$.
    Using this definition, we further bound the right-hand side of Equation~\eqref{eq:proof_regret_first_mid} as follows:
    \begin{align*}
        \sum_{t \notin   \WarmupRounds } 
            \sum_{i \in S_t \cup \{0 \} }
                p_t (i | S_t, \wb^\star)
                \left\| x_{ti}  - 
                    \EE^{\wb^\star}_t \left[ x_{tj} \right]  \right\|_{H_t^{-1}}^2
        &\leq  
        e \sum_{t \notin   \WarmupRounds } 
            \sum_{i \in S_t \cup \{0 \} }
                p_t (i | S_t, \wb^\star)
                \left\| x_{ti}  - 
                    \EE^{\wb^\star}_t \left[ x_{tj} \right]  \right\|_{H_t(\wb^\star)^{-1}}^2
        \tag{Lemma~\ref{lemma:Hessian_bound}}
        \\
        &\leq e \sum_{t \notin  \WarmupRounds } 
            \sum_{i \in S_t \cup \{0 \} }
                p_t (i | S_t, \wb^\star)
                \left\| x_{ti}  - 
                    \EE^{\wb^\star}_t \left[ x_{tj} \right]  \right\|_{H_t(\wb^\star)^{-1}}^2
        \\
        &\leq 2 e d \log \left( 1 + \frac{e T}{d \lambda} \right)
        \tag{Lemma~\ref{lemma:epl_H}}
        ,
    \end{align*}
    where when applying the elliptical potential lemma (Lemma~\ref{lemma:epl_H}), we verify the condition $\| x_{ti} \|_{H_t(\wb^\star)^{-1}}^2 \leq \frac{1}{2}$ for all $i \in S_t$ as follows:
    \begin{align*}
        \| x_{ti} \|_{H_t(\wb^\star)^{-1}}^2
        &\leq e \| x_{ti} \|_{H_t^{-1}}^2
        \tag{Lemma~\ref{lemma:Hessian_bound}, $\lambda = 144d$}
        \leq \frac{e}{\lambda}
        = \frac{e}{144 d}
        \leq \frac{1}{2}.
    \end{align*}
    Therefore, we can bound the term $I_1$ in Equation~\eqref{eq:proof_regret_decompose_first} as follows:
    \begin{align*}
        \sum_{t \notin   \WarmupRounds }  \nabla \tilde{Q}(\ub_t^\star)^\top (\ub_t - \ub^\star_t)
        \leq 2 \beta_T(\delta)
        \sqrt{ \sum_{t=1}^T \sigma_t^2 
        }
        \sqrt{
             2 e d \log \left( 1 + \frac{e T}{d \lambda} \right)
        }.
        \numberthis
        \label{eq:proof_regret_decompose_first_result}
    \end{align*}
    Now, we bound the term $I_2$ in~\eqref{eq:proof_regret_decompose_first}.
    We define a function $Q:\RR^{|S_t|} \rightarrow \RR$, such that for all $\ub = (u_1, \dots, u_{|S_t|}) \in \RR^{|S_t|}$, $Q(\ub) = \sum_{i=1}^{|S_t|} \frac{\exp(u_i) }{1 + \sum_{j=1}^{|S_t|} \exp(u_j)}$.
    Then, it is clear that $\left| \frac{\partial^2 \tilde{Q}}{\partial i \partial j} \right| 
    \leq  \left| \frac{\partial^2 Q}{\partial i \partial j} \right| $ since $r_{ti} \in [0,1]$.
    Hence, we get
    \begin{align*}
        \frac{1}{2} \sum_{t \notin   \WarmupRounds }  (\ub_t - \ub^\star_t)^\top \nabla^2 \tilde{Q}(\bar{\ub}_t) (\ub_t - \ub^\star_t)
        &\leq \frac{1}{2} \sum_{t \notin   \WarmupRounds }
        \sum_{i \in S_t} \sum_{j \in S_t}(u_{ti} - u_{ti}^\star ) \frac{\partial^2 \tilde{Q}}{\partial i \partial j} (u_{tj} - u_{tj}^\star )
        \\
        &\leq \frac{1}{2} \sum_{t=1}^T 
        \sum_{i \in S_t}
        \sum_{j \in S_t} |u_{ti} - u_{ti}^\star | \left| \frac{\partial^2 Q}{\partial i \partial j} \right| |u_{tj} - u_{tj}^\star |
        \tag{$r_{ti} \in [0,1]$}
        .
    \end{align*}
    Furthermore, we denote $p_i(\bar{\ub}_t) = \frac{\exp(\bar{u}_{ti})}{1 + \sum_{k=1}^{|S_t|} \exp(\bar{u}_{tk})}$.
    Then, we have
    \begin{align*}
        \frac{1}{2} \sum_{t=1}^T& \sum_{i \in S_t} \sum_{j \in S_t} |u_{ti} - u_{ti}^\star | \left| \frac{\partial^2 Q}{\partial i \partial j}\right| |u_{tj} - u_{tj}^\star |
        \\
        &= \frac{1}{2} \sum_{t=1}^T \sum_{i \in S_t} \sum_{j \in S_t, j \neq i} |u_{ti} - u_{ti}^\star | \left| \frac{\partial^2 Q}{\partial i \partial j} \right| |u_{tj} - u_{tj}^\star |
        + \frac{1}{2} \sum_{t=1}^T \sum_{i \in S_t } |u_{ti} - u_{ti}^\star | \left|\frac{\partial^2 Q}{\partial i \partial i} \right| |u_{ti} - u_{ti}^\star |
        \\
        &\leq \sum_{t=1}^T \sum_{i \in S_t} \sum_{j \in S_t, j \neq i} |u_{ti} - u_{ti}^\star| p_i(\bar{\ub}_t) p_j(\bar{\ub}_t) |u_{tj} - u_{tj}^\star|
        + \frac{3}{2} \sum_{t=1}^T \sum_{i \in S_t } (u_{ti} - u_{ti}^\star)^2 p_i(\bar{\ub}_t)
        \tag{Lemma~\ref{lemma:revenue_second_pd}}
        \\
        &\leq \sum_{t=1}^T\sum_{i \in S_t} \sum_{j \in S_t} |u_{ti} - u_{ti}^\star| p_i(\bar{\ub}_t) p_j(\bar{\ub}_t) |u_{tj} - u_{tj}^\star|
        + \frac{3}{2} \sum_{t=1}^T \sum_{i \in S_t } (u_{ti} - u_{ti}^\star)^2 p_i(\bar{\ub}_t)
        \\
        &\leq \frac{1}{2} \sum_{t=1}^T\sum_{i \in S_t} \sum_{j \in S_t} (u_{ti} - u_{ti}^\star)^2 p_i(\bar{\ub}_t) p_j(\bar{\ub}_t) 
        + \frac{1}{2} \sum_{i \in S_t} \sum_{j \in S_t}  (u_{tj} - u_{tj}^\star)^2 p_i(\bar{\ub}_t) p_j(\bar{\ub}_t)
        \tag{AM-GM inequality}
        \\
        &+ \frac{3}{2} \sum_{t=1}^T \sum_{i \in S_t } (u_{ti} - u_{ti}^\star)^2 p_i(\bar{\ub}_t)
        \\
        &\leq  \frac{5}{2} \sum_{t=1}^T \sum_{i \in S_t } (u_{ti} - u_{ti}^\star)^2 p_i(\bar{\ub}_t)
        .
    \end{align*}
    Therefore, we can bound the term $I_2$ in Equation~\eqref{eq:proof_regret_decompose_first} as follows:
    \begin{align*}
         \frac{1}{2} \sum_{t \notin   \WarmupRounds }  (\ub_t - \ub^\star_t)^\top \nabla^2 \tilde{Q}(\bar{\ub}_t) (\ub_t - \ub^\star_t)
        &\leq \frac{5}{2} \sum_{t=1}^T \sum_{i \in S_t } (u_{ti} - u_{ti}^\star)^2 p_i(\bar{\ub}_t)
        \\
        &= 10 \sum_{t=1}^T\sum_{i \in S_t }  p_i(\bar{\ub}_t) \beta_t(\delta)^2  \| x_{ti} \|_{H_t^{-1}}^2 
        \tag{definitions of $\ub_t$ and $\ub^\star_t$}
        \\
        &\leq  10 \sum_{t=1}^T \max_{i \in S_t } \beta_t(\delta)^2  \| x_{ti} \|_{H_t^{-1}}^2 
        \\
        &\leq 10 \beta_T(\delta)^2  \sum_{t=1}^T \max_{i \in S_t }  \| x_{ti} \|_{H_t^{-1}}^2
        \\
        &\leq \frac{20}{\kappa} \beta_T(\delta)^2  d \log \left(1 + \frac{T}{d \lambda} \right).
        \numberthis \label{eq:proof_regret_decompose_second_result}
    \end{align*}
    Finally, by substituting~\eqref{eq:proof_regret_decompose_first_result} and~\eqref{eq:proof_regret_decompose_second_result} into~\eqref{eq:proof_regret}, and  setting $\lambda = 144 d$, $\tau_T =\BigO \left( B \sqrt{d \log (T/\delta)} + B^{3/2}\sqrt{d} + B^2 \right)$, and $\beta_T(\delta) = \BigO \left( 
    \sqrt{
        d \log ( T/\delta)
        } + B \sqrt{d}
    \right)$, we obtain
    \begin{align*}
        \Regret (\wb^\star ) 
        &\leq
        \frac{2}{\kappa} \Threshold_T^2 d \log \left( 1+ \frac{T}{d \lambda} \right)
        + 2 \beta_T(\delta)
        \sqrt{ \sum_{t=1}^T \sigma_t^2 
        }
        \sqrt{
             2 e d \log \left( 1 + \frac{e T}{d \lambda} \right)
        }
        + \frac{20}{\kappa} \beta_T(\delta)^2  d \log \left(1 + \frac{T}{d \lambda} \right)
        \\
        &= \BigO \left(
            \left(d \log T 
            + B d \sqrt{ \log T} 
            \right)
            \sqrt{ \sum_{t=1}^T \sigma_t^2 
            }
            + \frac{1}{\kappa} B^3 d^2 
            \left(\log T\right)^2
            + \frac{1}{\kappa} B^4 d \log T
        \right)
        .
    \end{align*}
    By setting $\delta \leftarrow \delta/2$, we complete the proof of  Theorem~\ref{thm:regret_main}.
\end{proof}

\subsection{Proofs of Corollaries and Lemmas for Theorem~\ref{thm:regret_main}} 
\label{app_subsec:useful_lemmas_thm:regret_main}
\subsubsection{Proof of Corollary~\ref{cor:CS_warm-up}}
\label{app_subsubsec:proof_of_cor_CS-warm-up}
\begin{proof} [Proof of Corollary~\ref{cor:CS_warm-up}]
    In Theorem~\ref{thm:online_confidence_set}, we consider the case where $\Wcal_t = \Wcal = \{\wb \in \RR^d \mid \|\wb\|_2 \leq B \}$ for all $t \geq 1$ and the Hessian matrix is  $\WarmupHessian_t$.

    \underline{\textit{Condition: }  $\sup_{\wb \in  \Wcal_t  }
            |x_{ti}^\top (\wb  - \wb^\star) |
             \leq \alpha$ for all $i \in S_t$} $\quad\Longrightarrow \quad
             \alpha = 2B$.
    
    We set $\alpha = 2B$, as shown below:
    \begin{align*}
        \sup_{\wb \in \Wcal_t } 
        | x_{ti}^\top (\wb - \wb^\star) |
        &= \sup_{\wb \in \Wcal} 
        | x_{ti}^\top (\wb - \wb^\star) |
        \leq \sup_{\wb \in \Wcal} \| x_{ti}\|_2  \| \wb - \wb^\star\|_2 
        \leq 2B
        .
    \end{align*}
    Substituting $\alpha = 2B$ (which gives $\WarmupStep = \frac{1}{2} + 3\sqrt{2}B$) into Theorem~\ref{thm:online_confidence_set}, while setting $\WarmupRegualizer = \max \{ 12 \sqrt{2}\WarmupStep\alpha, 144 \WarmupStep d, 2  \}$, for $t \in \WarmupRounds$, we obtain
    \begin{align*}
        \|  \wb^\star - \WarmupParam_t\|_{\WarmupHessian_{t}} 
        \leq \zeta_t(\delta)  
        = \BigO \left( B \sqrt{d \log (t/\delta)} + B^{3/2} \sqrt{d} + B^2 \right).
    \end{align*}
    For $t \notin \WarmupRounds$, the confidence set $\WarmupConfidenceSet_t(\delta)$, along with $\WarmupParam_t$ and $\WarmupHessian_t$, remains unchanged.
    Thus, the proof is complete.
\end{proof}
%
\subsubsection{Proof of Corollary~\ref{cor:CS}}
\label{app_subsubsec:proof_of_cor_CS}
\begin{proof} [Proof of Corollary~\ref{cor:CS}]
    As with Corollary~\ref{cor:CS_warm-up}, we prove this using Theorem~\ref{thm:online_confidence_set}.
    Let $\Wcal_t = \WarmupConfidenceSet_t(\delta)$ and let the Hessian matrix be $H_t$.
    
    \underline{\textit{Condition: }  $\sup_{ \wb \in  \Wcal_t} 
            |x_{ti}^\top (\wb  - \wb^\star) |
             \leq \alpha$ for all $i \in S_t$}
             $\quad\Longrightarrow \quad
             \alpha = \frac{1}{3\sqrt{2}}$.
    
    We set $\alpha = \frac{1}{3\sqrt{2}}$, as shown below:
    \begin{align*}
        \sup_{\wb \in \Wcal_t} 
        | x_{ti}^\top (\wb - \wb^\star) |
        &= \sup_{\wb \in \WarmupConfidenceSet_t(\delta)} 
        | x_{ti}^\top (\wb - \wb^\star) |
        \\
        &\leq 
        \max_{x \in \Xcal_t} \| x\|_{(\WarmupHessian_t)^{-1}}
        \left(
            \max_{ \wb \in \WarmupConfidenceSet_t(\delta)} \| \wb - \WarmupParam_t \|_{\WarmupHessian_t}
            + \| \WarmupParam_t  - \wb^\star \|_{\WarmupHessian_t}
        \right)
        \tag{Hölder's inequality}
        \\
        &\leq \frac{1}{ \Threshold_t}
        \left(
            \max_{ \wb \in \WarmupConfidenceSet_t(\delta)} \| \wb - \WarmupParam_t \|_{\WarmupHessian_t}
            + \| \WarmupParam_t  - \wb^\star \|_{\WarmupHessian_t}
        \right)
        \tag{$t \notin \WarmupRounds$}
        \\
        &\leq \frac{1}{ 6 \sqrt{2} \zeta_t(\delta)}
        \left(
            \zeta_t (\delta)
            + \| \WarmupParam_t  - \wb^\star \|_{\WarmupHessian_t}
        \right)
        \tag{Definitions of $\Threshold_t$ and $\WarmupConfidenceSet_t(\delta)$}
        \\
        &\leq 
         \frac{ \zeta_t (\delta)}{ 3\sqrt{2} \zeta_t(\delta)}
         \tag{Corollary~\ref{cor:CS_warm-up}}
        \\
        &= \frac{1}{3\sqrt{2}}
        .
    \end{align*}
    Plugging $\alpha = \frac{1}{3\sqrt{2}}$ (which implies $\eta = \frac{1}{2}(1 + 3\sqrt{2}\alpha) =  1$) into Theorem~\ref{thm:online_confidence_set}, while setting $\lambda = \max \{ 12 \sqrt{2}\eta\alpha, 144 \eta d, 2  \}
    = 144 d
    $, for $t \notin \WarmupRounds$, we derive
    \begin{align*}
        \| \wb^\star - \wb \|_{H_t} 
        \leq \beta_t(\delta) 
        = \BigO \left( 
        \sqrt{d \log ( t/\delta)}
        + B \sqrt{d}
        \right).
    \end{align*}
    For $t \in \WarmupRounds$, the confidence set $\Ccal_t(\delta)$, along with $\wb_t$ and $H_t$, remains the same.
    This conclude the proof of Corollary~\ref{cor:CS}.
\end{proof}
%
\subsubsection{Proof of Lemma~\ref{lemma:r_geq_R}}
\label{app_subsubsec:proof_of_lemma:r_geq_R}
\begin{proof}[Proof of Lemma~\ref{lemma:r_geq_R}]
    By the definition of the optimal assortment and Lemma~\ref{lemma:optimism}, we have
    \begin{align*}
        R_t(S_t, \wb^\star)
        \leq R_{t}(S_{t}^\star, \wb^\star) 
        \leq \tilde{R}_t(S_{t}^\star)
        \leq \tilde{R}_t(S_{t}).
    \end{align*}
    Thus, it is sufficient to show that $r_{ti} \geq \tilde{R}_t(S_{t})$ for all $i \in S_t$.
    
    We prove this by contradiction. 
    Suppose there exists an item $i \in S_t$ such that $r_{ti} < \tilde{R}_t(S_{t})$.
    If we remove item $i$ from the assortment $ S_t$,
    it would result in higher expected revenue. 
    This contradicts the optimality of $S_t = \argmax_{S \in \Scal} \tilde{R}_t(S)$.
    Hence, we conclude
    \begin{align*}
        r_{ti} \geq \tilde{R}_t(S_{t}), \quad \forall i \in S_t,
    \end{align*}
    which completes the proof.
\end{proof}
%
\subsubsection{Proof of Lemma~\ref{lemma:epl_H}}
\label{app_subsubsec:proof_of_lemma:epl_H}
\begin{proof}[Proof of Lemma~\ref{lemma:epl_H}]
    For simplicity, let  $\EE^{\wb}_t[x_{tj}] = \EE_{j \sim p_t(\cdot | S_t, \wb)}[x_{tj}]$ and $x_{t0} = \mathbf{0}$.
    Then, we can express $\nabla^2 \ell_s (\wb)$ as follows:
    \begin{align*}
        \nabla^2 \ell_s (\wb)
        &= \sum_{i \in S_s} p_s(i | S_s, \wb) x_{si} x_{si}^\top -  \sum_{i \in S_s}  \sum_{j \in S_s} p_s(i | S_s, \wb) p_s(j | S_s, \wb) x_{si} x_{sj}^\top 
        \\
        &= \sum_{i \in S_s \cup \{0\}} p_s(i | S_s, \wb) x_{si} x_{si}^\top -  \sum_{i \in S_s \cup \{0\}}  \sum_{j \in S_s \cup \{0\}} p_s(i | S_s, \wb) p_s(j | S_s, \wb) x_{si} x_{sj}^\top
        \\
        &= \EE^{\wb}_s[x_{si}x_{si}^\top] 
        - \EE^{\wb}_s[x_{si}] \left(\EE^{\wb}_s[x_{si}] \right)^\top
        \\
        &= \EE^{\wb}_s\left[(x_{si} - \EE^{\wb}_s[x_{sj}]) (x_{si} - \EE^{\wb}_s[x_{sj}])^\top\right]
        \\
        &= \sum_{i \in S_s \cup \{ 0\} } p_s(i | S_s, \wb)
        (x_{si} - \EE^{\wb}_s[x_{sj}]) (x_{si} - \EE^{\wb}_s[x_{sj}])^\top
        . 
    \end{align*}
    Using the definition  $H_t(\wb) = \lambda \Ib_d + \sum_{s \in [t-1] \setminus \WarmupRounds} \nabla^2 \ell_s(\wb)$,  it follows that for any update round $s \in [t] \setminus \WarmupRounds$, we have
    \begin{align*}
        \det \left( H_{s+1} \right)
        = \det \left( H_s \right) \left( 1 +  \sum_{i \in S_s \cup \{ 0 \}}  p_s(i | S_s, \wb) 
        \left\| 
            x_{si} - \EE^{\wb}_s[x_{sj}]
        \right\|_{H_{s}(\wb)^{-1}}^2 
        \right).
    \end{align*}
    By the assumption that $\| x_{si} \|_{H_s(\wb)^{-1}}^2 \!\!\leq \frac{1}{2}$ for all $i \in S_s$, we know that $\sum_{i \in S_s \cup \{ 0 \}}  p_s(i | S_s, \wb) 
        \left\| 
            x_{si}  - \EE^{\wb}_s[x_{sj}]
        \right\|_{H_{s}(\wb)^{-1}}^2 \!\!\leq 1$.
    Then, using the fact that $z \leq 2 \log ( 1 + z)$ for any $z \in [0,1]$, we obtain
    \begin{align*}
        \sum_{s \in [t] \setminus \WarmupRounds} 
         \sum_{i \in S_s \cup \{ 0 \}}&
        p_s(i | S_s, \wb) \left\| 
            x_{si} - \EE^{\wb}_s[x_{sj}]
        \right\|_{H_{s}(\wb)^{-1}}^2 
        \\
        &\leq 2 \sum_{s \in [t] \setminus \WarmupRounds}
        \log \left(
            1 +  \sum_{i \in S_s \cup \{ 0 \}}
        p_s(i | S_s, \wb) \left\| 
            x_{si} - \EE^{\wb}_s[x_{sj}]
        \right\|_{H_{s}(\wb)^{-1}}^2 
        \right)
        \\
        &=   2 \sum_{s \in [t] \setminus \WarmupRounds}
        \log \left(
            \frac{\det(H_{s+1})}{\det(H_1)}
        \right)
        \\
        &= 2 \log \left( \frac{\det (H_{t+1})}{\det(H_1)} \right)
        \\
        &\leq 2 d \log  \left( \frac{\tr(H_{t+1}) }{d \lambda}  \right)
        \leq 2d \log \left(1 + \frac{t}{d \lambda} \right),
    \end{align*}
    which concludes the proof.
\end{proof}
%
\subsubsection{Proof of Lemma~\ref{lemma:Hessian_bound}}
\label{app_subsubsec:proof_of_lemma:Hessian_bound}
\begin{proof}[Proof of Lemma~\ref{lemma:Hessian_bound}]
    For any $s \in [t-1] \setminus \WarmupRounds$,
    let $X_s \in \RR^{|S_t| \times d}$ be the matrix whose $i$'th row is $x_{si}^\top$.
    Then, by the equivalent notation of the loss  (see Equation~\eqref{eq:equi_loss}), we have
    \begin{align*}
        \nabla^2 \ell_s (\wb_{s+1}) 
        &= X_s^\top \nabla_{\zb}^2 \bar{\ell}_s(\zb_{s+1}) X_s
        \tag{Eqn.~\eqref{eq:equi_loss}}
        \\
        &\preceq 
        e^{3\sqrt{2} \| \zb_{s+1} - \zb^\star_s \|_{\infty} } 
        X_s^\top  
        \nabla_{\zb}^2 \bar{\ell}_s(\zb^\star_s) 
        X_s
        \tag{Proposition~\ref{prop:self_hessian_norm}}
        \\
        &\preceq
        e X_s^\top  
        \nabla_{\zb}^2 \bar{\ell}_s(\zb^\star_s) 
        X_s
        \tag{$\| \zb_{s+1} - \zb^\star_s \|_{\infty} \leq \frac{1}{3\sqrt{2}}$}
        \\
        &= e \nabla^2 \ell_s (\wb^\star),
    \end{align*}
    where the last inequality holds because, for any $s \notin \WarmupRounds$, the following holds:
    \begin{align*}
        \| \zb_{s+1} - \zb^\star_s \|_{\infty} 
        &= \max_{i \in S_s} |x_{si}^\top (\wb_{s+1} - \wb^\star)|
        \\
        &=  \max_{i \in S_s} \| x_{si} \|_{(\WarmupHessian_s)^{-1}}
        \left(
            \|\wb_{s+1} - \WarmupParam_s \|_{\WarmupHessian_s}
            + \| \WarmupParam_s - \wb^\star \|_{\WarmupHessian_s}
        \right)
        \tag{Hölder's inequality}
        \\
        &\leq \frac{1}{\Threshold_s} \left(
            \|\wb_{s+1} - \WarmupParam_s \|_{\WarmupHessian_s}
            + \| \WarmupParam_s - \wb^\star \|_{\WarmupHessian_s}
        \right)
        \tag{$s \notin \WarmupRounds$}
        \\
        &\leq \frac{1}{6 \sqrt{2} \zeta_s(\delta)} \left(  
            \zeta_s(\delta) + \| \WarmupParam_s - \wb^\star \|_{\WarmupHessian_s}
        \right)
        \tag{Definitions of $\Threshold_s$ and $\wb_{s+1} \in \WarmupConfidenceSet_s(\delta)$}
        \\
        &\leq 
        \frac{ \zeta_s (\delta)}{ 3\sqrt{2} \zeta_s(\delta)}
        \tag{Corollary~\ref{cor:CS_warm-up}} 
        \\
        &= \frac{1}{3\sqrt{2}}.
    \end{align*}
    Thus, we get
    \begin{align*}
        H_t= \lambda \Ib_d + \sum_{s \in [t-1] \setminus \WarmupRounds } \nabla^2 \ell_s(\wb_{s+1})
        \preceq
        \lambda \Ib_d + 
        e \sum_{s \in [t-1] \setminus \WarmupRounds } \nabla^2 \ell_s(\wb^\star)
        = e H_t(\wb^\star).
    \end{align*}
    To prove the other inequality, we use a similar line of reasoning: 
    \begin{align*}
         \nabla^2 \ell_s (\wb^\star) 
         &= X_s^\top \nabla_{\zb}^2 \bar{\ell}_s(\zb^\star_s) X_s
        \preceq 
        e^{3\sqrt{2} \| \zb_{s+1} - \zb^\star_s \|_{\infty} } 
        X_s^\top  
        \nabla_{\zb}^2 \bar{\ell}_s(\zb_{s+1}) 
        X_s
        \preceq
        e X_s^\top  
        \nabla_{\zb}^2 \bar{\ell}_s(\zb_{s+1}) 
        X_s
        = e \nabla^2 \ell_s (\wb_{s+1}),
    \end{align*}
    which implies that
    \begin{align*}
        H_t= \lambda \Ib_d + \sum_{s \in [t-1] \setminus \WarmupRounds } \nabla^2 \ell_s(\wb_{s+1})
        \succeq
        \lambda \Ib_d + \frac{1}{e}  \sum_{s \in [t-1] \setminus \WarmupRounds } \nabla^2 \ell_s(\wb^\star)
        \succeq
        \frac{1}{e} H_t(\wb^\star).
    \end{align*}
    This concludes the proof.
\end{proof}
\subsubsection{Proof of Lemma~\ref{lemma:bound_T_w-T_0}}
\label{app_subsubsec:proof_of_lemma:bound_T_w-T_0}
\begin{proof}[Proof of Lemma~\ref{lemma:bound_T_w-T_0}]
    Recall that by the definition of $\WarmupHessian_t $, we have
    \begin{align*}
        \WarmupHessian_t 
        &= \lambda \Ib_d + \sum_{s \in \WarmupRounds \setminus \{t, \dots, T \} } \nabla^2 \ell_s (\wb_{s+1})
        \\
        &= \lambda \Ib_d 
        + \sum_{s \in \WarmupRounds \setminus \{t, \dots, T \} } p_s(i_s | \{i_s\}, \wb_{s+1}) x_{s i_s} x_{s i_s}^\top 
        -  p_s(i_s | \{i_s\}, \wb_{s+1}) p_s(i_s | \{i_s\}, \wb_{s+1}) x_{s i_s} x_{s i_s}^\top
        \\
        &= \lambda \Ib_d 
        + \sum_{s \in \WarmupRounds \setminus \{t, \dots, T \} } 
        p_s(i_s | \{i_s\}, \wb_{s+1})
         p_s(0 | \{i_s\}, \wb_{s+1}) 
        x_{s i_s} x_{s i_s}^\top,
    \end{align*}
    where $i_s$ is the index of the item such that $x_{s i_s} = \argmax_{x \in \Xcal_s} \|x \|_{(\WarmupHessian_s)^{-1}}^2 $.
    Then, we get
    \begin{align*}
        \sum_{t \in \WarmupRounds} 
        \max_{x \in \Xcal_t} \| x \|_{(\WarmupHessian_t)^{-1}}^2
        &= \sum_{t \in \WarmupRounds} 
         \| x_{t i_t} \|_{(\WarmupHessian_t)^{-1}}^2
         \tag{$x_{t i_t} = \argmax_{x \in \Xcal_t} \|x \|_{(\WarmupHessian_t)^{-1}}^2$ for  $t \in \WarmupRounds$}
        \\
        &\leq 
        \frac{1}{\kappa}
        \sum_{t \in \WarmupRounds} 
        p_t(i_t | \{i_t\}, \wb_{t+1})
         p_t(0 | \{i_t\}, \wb_{t+1}) 
         \| x_{t i_t} \|_{(\WarmupHessian_t)^{-1}}^2
         \tag{Definition of $\kappa$}
         \\
         &= 
         \frac{1}{\kappa}
         \sum_{t \in \WarmupRounds } 
         \min \left\{
            \frac{1}{2}, \,
             p_t(i_t | \{i_t\}, \wb_{t+1})
             p_t(0 | \{i_t\}, \wb_{t+1}) 
             \| x_{t i_t} \|_{(\WarmupHessian_t)^{-1}}^2 
         \right\}
         \tag{$\| x_{t i_t} \|_{(\WarmupHessian_t)^{-1}}^2 \leq
         \frac{1}{\WarmupRegualizer}\| x_{t i_t} \|_2 \leq 
         \frac{1}{2}$, $\WarmupRegualizer \geq 2$}, 
         \\
         &\leq \frac{2}{\kappa} d \log \left( 1+ \frac{T}{d \lambda} \right).
         \tag{Lemma~\ref{lemma:elliptical_x_tilde_lee}}
    \end{align*}
    On the other hand, by the definition of the update rule in Algorithm~\ref{alg:main_online}, we have
    \begin{align*}
        \sum_{t \in \WarmupRounds} 
        \max_{x \in \Xcal_t} \| x \|_{(\WarmupHessian_t)^{-1}}^2
        \geq
        \sum_{t \in \WarmupRounds} 
        \frac{1}{\Threshold_t^2} 
        \geq
        \frac{1}{\Threshold_T^2} \left|\WarmupRounds \right|
        \tag{$\Threshold_t$ is non-decreasing}
        .
    \end{align*}
    By combining the two results above, we obtain
    \begin{align*}
        \left|\WarmupRounds\right|
        \leq \frac{2}{\kappa} \Threshold_T^2 d \log \left( 1+ \frac{T}{d \lambda} \right),
    \end{align*}
    which concludes the proof.
\end{proof}

\subsection{Technical Lemmas} 
\label{app_sec:technical_leamms_regret_main}
\begin{lemma} [Lemma F.2 and H.3 of~\citealt{lee2024nearly}]  \label{lemma:elliptical_x_tilde_lee}
Let $H_t = \lambda \Ib_d + \sum_{s=1}^{t-1} \nabla^2 \ell_s(\wb_{s+1})$.
Define $\tilde{x}_{si} := x_{si} - \EE_{ j \sim p_s(\cdot | S_s, \wb_{s+1}) }[x_{sj}]$.
If $\| x_{si} \|_{H_s^{-1}}^2 \leq \frac{1}{2}$ for all $i \in S_t$ and $s \in [t]$, then the following statements hold true:
\begin{enumerate}[label={(\arabic*)}]
    \item $\sum_{s=1}^{t} 
    \sum_{i \in S_s} p_s(i | S_s, \wb_{s+1}) p_s(0 | S_s, \wb_{s+1}) \| x_{si}\|_{H_s^{-1}}^2
    \leq 2d \log \left( 1+ \frac{t}{d \lambda} \right)$,
    
    \item $\sum_{s=1}^{t} \sum_{i \in S_s} p_s(i | S_s, \wb_{s+1}) \| \tilde{x}_{si}\|_{H_s^{-1}}^2
    \leq 2d \log \left( 1+ \frac{t}{d \lambda} \right)$,
    
    \item $\sum_{s=1}^{t} \max_{i \in S_s} \| x_{si}\|_{H_s^{-1}}^2
    \leq \frac{2}{\kappa} d \log \left( 1+ \frac{t}{d \lambda} \right)$,
    \item $\sum_{s=1}^{t} \max_{i \in S_s} \| \tilde{x}_{si}\|_{H_s^{-1}}^2
    \leq \frac{2}{\kappa} d \log \left( 1+ \frac{t}{d \lambda} \right)$.
\end{enumerate}
\end{lemma}
%

\section{Instance-Dependent Regret}
\label{app_sec:discussion_instance_regret}
As a special case, if the rewards are uniform (i.e., $r_{ti}=1$), we can establish an instance-dependent regret bound.
\begin{proposition} [Restatement of Proposition~\ref{prop:instance_regret_uniform_r} Instance-dependent regret under uniform rewards]
\label{prop:instance_regret_uniform_r_app}
    Define $\kappa_t^\star \!:=\! \sum_{i \in S_t^\star} p_t(i | S_t^\star, \mathbf{w}^\star)p_t(0 | S_t^\star, \mathbf{w}^\star)$.
    Under the same conditions as Theorem~\ref{thm:regret_main} and assuming uniform rewards, the regret of \textup{\AlgName{}} is upper bounded by
    \begin{align*}
       \Regret(\wb^\star) 
       = \BigO \left( \left( d \log T + B d \sqrt{ \log T}  \right)\sqrt{ \sum_{t=1}^T  \kappa^\star_t }
       + \frac{1}{\kappa} B^3 d^2 
            \left(\log T \right)^2
            + \frac{1}{\kappa} B^4 d \log T
       \right).
    \end{align*}
\end{proposition}
\subsection{Proof of Proposition~\ref{prop:instance_regret_uniform_r} }
\label{app_sec:proof_prop:instance_regret_uniform_r}
In this section, we present the proof of Proposition~\ref{prop:instance_regret_uniform_r}.
In the case of uniform rewards, where $r_{ti=1}$ for all $i \in [N]$, the $\sigma_t^2$ term can be upper-bounded by  $\kappa^\star_t$ plus an additive term.
\begin{proof} [Proof of Proposition~\ref{prop:instance_regret_uniform_r}]
    From Theorem~\ref{thm:regret_main}, we have
    \begin{align*}
        \Regret (\wb^\star ) 
        = \BigO \left(
            \left(d \log T 
            + B d \sqrt{ \log T} 
            \right)
            \sqrt{ \sum_{t=1}^T \sigma_t^2 
            }
            + \frac{1}{\kappa} B^3 d^2 
            \left(\log T \right)^2
            + \frac{1}{\kappa} B^4 d \log T
        \right)
        .
    \end{align*}
    When rewards are uniform, we can rewrite the $\sum_{t=1}^T \sigma_t^2$ term as follows:
    \begin{align*}
        \sum_{t=1}^T\sigma_t^2
        &=\sum_{t=1}^T \EE_{i \sim p_t(\cdot | S_t, \wb^\star)} 
        \left[
            \left(
                r_{ti} - 
                \EE_{j \sim p_t(\cdot | S_t, \wb^\star)} [r_{tj}]  
            \right)^2
        \right]
        \\
        &= \sum_{t=1}^T\sum_{i \in S_t}
        p_t(i | S_t, \wb^\star) r_{ti}^2
        -  \left(\sum_{i \in S_t}  p_t(i | S_t, \wb^\star) r_{ti} \right)^2
        \\
        &=  \sum_{t=1}^T\sum_{i \in S_t}
        p_t(i | S_t, \wb^\star)
        p_t(0 | S_t, \wb^\star)
        \tag{$r_{ti}=1$}
        \\
        &\leq 
        \sum_{t=1}^T \kappa_t^\star + \Regret(\wb^\star).
    \end{align*}
    where the last inequality holds by the following lemma:
    \begin{lemma} [Lemma 11 of~\citealt{perivier2022dynamic}]
        \label{lemma:kappa_star_bound}
        Let $\kappa_t^\star := \sum_{i \in S_t^\star} p_t(i | S_t^\star, \mathbf{w}^\star)p_t(0 | S_t^\star, \mathbf{w}^\star)$.
        Then, we have
        \begin{align*}
            \sum_{t=1}^T\sum_{i \in S_t} p_t(i | S_t, \mathbf{w}^\star)p_t(0 | S_t, \mathbf{w}^\star)
            \leq \sum_{t=1}^T \kappa_t^\star + \Regret(\wb^\star).
        \end{align*}
    \end{lemma}
    Hence, we get
    \begin{align*}
        \Regret (\wb^\star ) 
        = \BigO \left(
            \left(d \log T 
            + B d \sqrt{ \log T} 
            \right)
            \sqrt{ \sum_{t=1}^T \kappa_t^\star + \Regret(\wb^\star) 
            }
            + \frac{1}{\kappa} B^3 d^2 
            \left(\log T\right)^2
            + \frac{1}{\kappa} B^4 d \log T
        \right)
        .
    \end{align*}
    Solving the above equation completes the proof of Proposition~\ref{prop:instance_regret_uniform_r}.
\end{proof}
\subsection{Discussion on Instance-Dependent Regret}
\label{app_subsec:instance_regret}
In this subsection, we further discuss about the instance-dependent parameter $\kappa^\star_t$ and the variance of rewards $\sigma_t$.

\textbf{True meaning of $\kappa^\star_t$. }
The instance-dependent parameter $\kappa^\star_t$, 
which appears in the regret bounds of many existing (multinomial) logistic and GLM bandits~\citep{abeille2021instance, faury2022jointly, perivier2022dynamic, lee2024nearly, sawarni2024generalized}, is indeed the \textit{variance of the uniform rewards given $S^\star_t$}.
In contrast, $\sigma_t^2$ denotes the variance of general rewards (including both uniform and non-uniform) for the offered assortment $S_t$.
Under uniform rewards, as shown in the analysis of Proposition~\ref{prop:instance_regret_uniform_r}, $\kappa^\star_t$ and $\sigma_t^2$ are closely related, as the assortment size remains the same and the rewards are identical.

\textbf{Possibility of Instance-Dependent Regret under Non-Uniform Rewards. }
Readers might expect an instance-dependent regret bound for general non-uniform rewards. 
However, we cautiously argue that establishing such a bound in the non-uniform case is non-trivial using existing analytical approaches.
Unlike prior works on binary logistic bandits~\citep{abeille2021instance, faury2022jointly}, uniform rewards MNL bandits~\citep{perivier2022dynamic, lee2024nearly}, and generalized linear bandits~\citep{sawarni2024generalized}, the size and rewards of the offered assortment  $S^\star$ and the optimal assortment $S^\star_t$ are different.
This fundamental discrepancy makes it impossible to bound quantities related to $S_t$ using those related to $S^\star_t$.

\section{Proof of Theorem~\ref{thm:MLE}} 
\label{app_sec:proof_thm:MLE}
In this section, we provide the proof of Theorem~\ref{thm:MLE}.
For ease of reference, Table~\ref{table_symbols_MLE} summarizes the notations used for \AlgNameMLE{}.
\begin{table}[h!]
\centering
    \caption{Symbols for \AlgNameMLE{}}
    \label{table_symbols_MLE}
    \begin{tabular}{ll}
         \toprule
         $ \mathcal{L}_t(\wb) $       &   $:= - \sum_{s=1}^{t-1} \sum_{i \in S_s} y_{ts} \log p_s(i | S_s, \wb)$. \\[0.1cm]
         $\hat{\wb}_t$       &    maximum likelihood estimate (MLE) estimate at round $t$ \\[0.1cm] 
         $\MLERegualizer$       &  $:=\frac{1}{8 B^2}$,  regularization parameter for MLE \\[0.1cm] 
         $\MLEHessian_t(\wb) $       &    $:= \sum_{s=1}^{t-1} \nabla^2 \ell_s (\wb) + \MLERegualizer \Ib_d$ \\[0.1cm] 
         $\nub_t^\star$       &   parameter that satisfies $\frac{1}{2} \left\|
            \wb^\star - \MLEParam_t
            \right\|_{\nabla^2 \mathcal{L}_t(\nub_t^\star)}^2
            =  \left\|
                    \wb^\star - \MLEParam_t
                \right\|_{\int_0^1 (1-v) \nabla^2 \mathcal{L}_t (\MLEParam_t + v (\wb^\star - \MLEParam_t)) \dd v }^2$  \\[0.1cm] 
         $\MLEConfRadius_t(\delta)$       &    $:= \BigO \left( \sqrt{d \log (B t / \delta)} \right)$ \\[0.1cm] 
         $\MLEConfRadiusEllip_t(\delta)$       &    $:= \sqrt{2 \MLEConfRadius_t(\delta)^2 + 1} $ \\[0.1cm] 
         $\MLEConfidenceSet_t (\delta) $       &    $:= 
                                                    \left\{
                                                        \wb \in \Wcal : 
                                                        \mathcal{L}_t (\wb) - \mathcal{L}_t (\MLEParam_t)
                                                        \leq 
                                                        \MLEConfRadius_t(\delta)^2
                                                    \right\} $ \\[0.1cm] 
         $\MLEUCBParam_{ti}$       &    $:=\argmax_{\wb \in \MLEConfidenceSet_t(\delta)} x_{ti}^\top \wb$,  optimistic utility of item $i$ at round $t$ \\[0.1cm] 
         $\MLERevenue_{t}(S)$       &  $:= \frac{\sum_{i \in S} \exp( x_{ti}^\top \MLEUCBParam_{ti} ) r_{ti} }{1 + \sum_{j \in S} \exp(x_{tj}^\top \MLEUCBParam_{tj} )}$,    optimistic expected revenue of assortment $S$ at round $t$ \\[0.1cm] 
         $\sigma_t^2$   &  $:= 
                            \EE_{i \sim p_t(\cdot | S_t, \wb^\star)} 
                                \!\left[
                                    \left(
                                        r_{ti} - 
                                        \EE_{j \sim p_t(\cdot | S_t, \wb^\star)} [r_{tj}]  
                                    \right)^2
                                \right]$, variance of rewards given $S_t$ at round $t$   \\[0.1cm]
         \bottomrule
    \end{tabular}
\end{table}

We define $\mathcal{L}_t(\wb)$ as
the negative log-likelihood of $\wb$ with respect to data collected up to $t-1$, 
and $\MLEParam_t$ as the corresponding maximum likelihood estimate (MLE) estimate:
\begin{align*}
    \mathcal{L}_t(\wb) := \sum_{s=1}^{t-1} \ell_s (\wb)
    = - \sum_{s=1}^{t-1} \sum_{i \in S_s} y_{ts} \log p_s(i | S_s, \wb), \quad
    \MLEParam_t := \argmin_{\wb \in \Wcal} \mathcal{L}_t(\wb).
\end{align*}
And the confidence set is defined as follows:
\begin{align*}
    \MLEConfidenceSet_t (\delta) := 
        \left\{
            \wb \in \Wcal : 
            \mathcal{L}_t (\wb) - \mathcal{L}_t (\MLEParam_t)
            \leq 
            \MLEConfRadius_t(\delta)^2
        \right\},
\end{align*}    
where
\begin{align*}
    \MLEConfRadius_t(\delta) 
    :=\sqrt{ \log \frac{1}{\delta}
    + d \log \left( \max \left\{e,  \frac{4 e B (t-1)}{d}   \right\}  \right)
    }.
\end{align*}
The confidence radius $\MLEConfRadius_t(\delta)$ follows directly from Theorem 3.1 in~\citet{lee2024unified}. 
This result is derived by incorporating the Lipschitz constant for the MNL loss, i.e., $L_t = \max_{\wb \in \Wcal} \left\| \nabla \mathcal{L}_t(\wb) \right\|_2 
\leq (t-1) \left\| \nabla \ell_t(\wb) \right\|_2
\leq 2(t-1)$ (under Assumption~\ref{assum:bounded_assumption}).
\begin{lemma} [Unified CS for generalized linear models (GLMs), Theorem 3.1 of~\citealt{lee2024unified}]
\label{lemma:thm3.2_lee2024unified}
    Let $L_t := \max_{\wb \in \Wcal} \left\| \nabla \mathcal{L}_t(\wb) \right\|_2$ be the Lipschitz constant of $\mathcal{L}_t( \cdot )$, which  may depend on $\{(x_s, r_s )\}_{s=1}^{t-1}$. 
    Then,
    we have 
    $\textup{Pr}[\forall t \geq 1, \wb^\star \in \MLEConfidenceSet_t (\delta)] \geq 1- \delta$, where
    \begin{align*}
        \MLEConfidenceSet_t (\delta) := 
        \left\{
            \wb \in \Wcal : 
            \mathcal{L}_t (\wb) - \mathcal{L}_t (\MLEParam_t)
            \leq 
            \MLEConfRadius_t(\delta)^2
            =  \log \frac{1}{\delta}
            + d \log \left( \max \left\{e,  \frac{2 e B L_t}{d}   \right\}  \right)
            \right\}.
    \end{align*}
\end{lemma}
Then, we offer an assortment $S_t$ that maximizes the optimistic expected revenue $\MLERevenue_t(S)$ as follows:
\begin{align*}
    S_t &= \argmax_{S \in \Scal} \MLERevenue_t(S)
    = \argmax_{S \in \Scal} 
    \frac{\sum_{i \in S} \exp( x_{ti}^\top \MLEUCBParam_{ti} ) r_{ti} }{1 + \sum_{j \in S} \exp(x_{tj}^\top \MLEUCBParam_{tj} )},
    \quad 
    \text{where }\,\,
    \MLEUCBParam_{ti} = \argmax_{\wb \in \MLEConfidenceSet_t(\delta)} x_{ti}^\top \wb.
    \numberthis \label{eq:MLE_optimization}
\end{align*}
Additionally, we define the Hessian of the regularized loss at $\wb$ as:
\begin{align*}
    \MLEHessian_t(\wb)
    := \sum_{s=1}^{t-1} \nabla^2 \ell_s (\wb)
    + \MLERegualizer \Ib_d,
    \quad 
    \text{where }\,\,
    \MLERegualizer = \frac{1}{8B^2}.
\end{align*}
Now, we present useful lemmas that will be used in the proof of Theorem~\ref{thm:MLE}.
\begin{lemma} [Restatement of Lemma~\ref{lemma:MLE_CS_main},  Improved MLE confidence bound]
\label{lemma:MLE_CS}
    For any $t \in [T]$,
    we define $\nub_t^\star$ such that 
    $\frac{1}{2} \left\|
            \wb^\star - \MLEParam_t
        \right\|_{\nabla^2 \mathcal{L}_t(\nub_t^\star)}^2
    =  \left\|
            \wb^\star - \MLEParam_t
        \right\|_{\int_0^1 (1-v) \nabla^2 \mathcal{L}_t (\MLEParam_t + v (\wb^\star - \MLEParam_t)) \dd v }^2$ and
    $\MLEHessian_t(\nub_t^\star)
    := \nabla^2 \mathcal{L}_t (\nub_t^\star) + \MLERegualizer \Ib_d
    = \sum_{s=1}^{t-1} \nabla^2 \ell_s(\nub_t^\star) + \MLERegualizer \Ib_d$.
    Let $\MLERegualizer = \frac{1}{8B^2}$.
    Then, for any $t \geq 1$, 
    if $\wb^\star \in \MLEConfidenceSet_t(\delta)$ and Assumption~\ref{assum:bounded_assumption} holds, then 
    we have
    \begin{align*}
        \left\|
            \wb^\star - \MLEParam_t
        \right\|_{\MLEHessian_t(\nub_t^\star)}^2
        \leq  \underbrace{2 \MLEConfRadius_t(\delta)^2  + 1}_{=: \MLEConfRadiusEllip_t(\delta)^2}
        = \BigO \left(d \log (B t) \right)
        .
    \end{align*}
\end{lemma}
The proof is provided in Appendix~\ref{app_subsec:useful_lemmas_thm:MLE}.
By Lemma~\ref{lemma:MLE_CS}, we define the ellipsoidal version of the confidence radius as follows:
\begin{align*}
    \MLEConfRadiusEllip_t(\delta)
    := 
    \sqrt{2 \log \frac{1}{\delta}
    + 2 d \log \left( \max \left\{e,  \frac{4 e B (t-1)}{d}   \right\}  \right)
    + 1}
    = \BigO \left(\sqrt{d \log (B t)} \right).
\end{align*}
Additionally, we present useful technical lemmas.
\begin{lemma}
    \label{lemma:gap_p_wb1_wb2}
    For any $t \in [T]$, $\wb_1, \wb_2 \in \MLEConfidenceSet_t(\delta)$, and $\omega_{ti} \geq 0$, we have
    \begin{align*}
        \sum_{i \in S_t} 
        \left|
            p_t(j | S_t, \wb_1) 
            - p_t(j | S_t, \wb_2) 
        \right|
        \omega_{ti}
        \leq 
         4\MLEConfRadiusEllip_t(\delta) 
        \max_{i \in S_t} \omega_{ti}
        \max_{i \in S_t} 
        \| x_{ti} \|_{\MLEHessian_t(\nub^\star_t)^{-1}}.
    \end{align*}
\end{lemma}
The proof is deferred to Appendix~\ref{app_subsubsec:proof_of_lemma:gap_p_wb1_wb2}.
\begin{lemma} [Elliptical potential count lemma, Lemma 4 of~\citealt{kim2022improved}]
\label{lemma:epcl_kim}
    For $X, L >0$, let $x_1, \dots, x_T \in \RR^d$  be a sequence of vectors with $\|x_t\|_2 \leq X$ for all $t \in [T]$.
    Let
     $V_t := \lambda \Ib_d + \sum_{s=1}^{t-1} x_s x_s^\top$ for some $\lambda > 0$.
    Let $\Jcal \subseteq [T]$ be the set of indices where $\| x_t \|_{V_t^{-1}}^2 \geq L$.
    Then, 
    \begin{align*}
        |\Jcal|
        \leq 
        \frac{2}{ \log (1 + L) }
        d
        \log \left(
            1 + \frac{X^2}{ \log (1 + L) \lambda }
        \right).
    \end{align*}
\end{lemma}
%
%
\begin{algorithm*}[t!]
   \caption{\AlgNameMLE{}, \textbf{OFU}-\textbf{M}aximum Likelihood Estimation \textbf{MNL}}
   \label{alg:MLE}
    \begin{algorithmic}[1]
       \State {\bfseries Input:} failure level $\delta$, 
       confidence radius $\MLEConfRadius_t(\delta)$.
       %
       \For{ round $t = 1, \dots, T$}
            \State Observe feature set $\Xcal_t$.
            \State Calculate the norm-constrained MLE: $\MLEParam_t \leftarrow \argmin_{\wb \in \Wcal} \mathcal{L}_t(\wb) $.
            \State Update $\MLEConfidenceSet_t(\delta) \leftarrow 
            \left\{
                \wb \in \Wcal : 
                \mathcal{L}_t (\wb) - \mathcal{L}_t (\MLEParam_t)
                \leq 
                \MLEConfRadius_t(\delta)^2
            \right\}$.
            \State Set $\MLEUCBParam_{ti} \leftarrow \argmax_{\wb \in \MLEConfidenceSet_t(\delta)} x_{ti}^\top \wb$ \, for all $i \in [N]$.
            \label{algo_eq:opt_mle_app}
            \State Offer $S_t = \argmax_{S \in \mathcal{S}} \MLERevenue_t (S)$ and observe $\yb_t$.
       \EndFor
    \end{algorithmic}
\end{algorithm*}
\subsection{Main Proof of Theorem~\ref{thm:MLE}} \label{app_subsec:main_proof_thm_MLE}
\begin{proof} [Proof of Theorem~\ref{thm:MLE}]
    We follow a reasoning process similar to that used in the proof of Theorem~\ref{thm:regret_main}.

    First, we define the set of large elliptical potential rounds as follows:
    \begin{align*}
        \MLEWarmupRounds_0 
        &:= \left\{
            t \in [T]:
            \| x_{ti} \|_{\MLEHessian_t(\nub^\star_t)^{-1}}^2 
            \geq \frac{1}{2},
            \quad
            \forall i \in S_t
        \right\}.
    \end{align*}    

    Let $\UCB_{ti} = x_{ti}^\top \MLEParam_t + \MLEConfRadiusEllip_t (\delta) \|  x_{ti}\|_{\MLEHessian_t(\nub_t^\star)^{-1}}$ and 
    $\widebar{\UCB}_{ti}$ as $\widebar{\UCB}_{ti} := x_{ti}^\top \wb^\star + 2 \MLEConfRadiusEllip_t(\delta) \| x_{ti} \|_{\MLEHessian_t(\nub_t^\star)^{-1}}$.
    Then, for all $i \in [N]$ and $t \geq 1$, we have
    \begin{align*}
        x_{ti}^\top \MLEUCBParam_{ti} - x_{ti}^\top \wb^\star
        &\leq \UCB_{ti} - x_{ti}^\top \wb^\star
        \tag{Definition of $\MLEUCBParam_{ti}$, Lemma~\ref{lemma:MLE_CS}}
        \leq 2 \MLEConfRadiusEllip_t(\delta) \| x_{ti} \|_{\MLEHessian_t(\nub_t^\star)^{-1}}
        ,
    \end{align*}
    which implies $x_{ti}^\top \MLEUCBParam_{ti} \leq \widebar{\UCB}_{ti}$.
    Thus, by Lemma~\ref{lemma:increasing_R}, we get
    \begin{align*}
        \MLERevenue_{t}(S_t) 
        \leq \dbtilde{R}^{\text{MLE}}_t(S_{t}),
        \numberthis \label{eq:proof_MLE_overly_R}
    \end{align*}
    where $\dbtilde{R}^{\text{MLE}}_t(S_{t}) := \frac{\sum_{i \in S_t} \exp( \widebar{\UCB}_{ti} ) r_{ti} }{1 + \sum_{j \in S_t} \exp(\widebar{\UCB}_{tj}) }$.
    
    Define a function  $\tilde{Q}:\RR^{|S_t|} \rightarrow \RR$, such that for all $\ub = (u_1, \dots, u_{|S_t|})^\top \in \RR^{|S_t|}$, $\tilde{Q}(\ub) = \sum_{k=1}^{|S_t|} \frac{\exp(u_k)r_{ti_k} }{1 + \sum_{j=1}^{|S_t|} \exp(u_j)}$.
    For simplicity, we write $S_t = \{i_1, \dots, i_{|S_t|} \}$.
    Furthermore, we denote $\ub_t = (u_{ti_1}, \dots u_{ti_{|S_t|}} )^\top = (\widebar{\UCB}_{ti_1} , \dots, \widebar{\UCB}_{ti_{{|S_t|}}} )^\top$ 
    and $\ub^\star_t = (u_{ti_1}^\star, \dots u_{ti_{|S_t|}}^\star )^\top = (x_{ti_1}^\top \wb^\star, \dots, x_{ti_{{|S_t|}}}^\top \wb^\star)^\top$.

    Then, by the elliptical potential count lemma (Lemma~\ref{lemma:epcl_kim}), we get
    \begin{align*}
        \sum_{t = 1}^T   R_{t}(S_{t}^\star, \wb^\star) -  R_{t}(S_{t}, \wb^\star)
        &= 
        \left| \MLEWarmupRounds_t \right|
        + \sum_{t \notin \MLEWarmupRounds_t}   R_{t}(S_{t}^\star, \wb^\star) -  R_{t}(S_{t}, \wb^\star) 
        \\
        &\leq 
        \frac{2}{\log ( 3/2)} d
        \log \left(
            1 + 
            \frac{1}{\log(3/2) \MLERegualizer }
        \right)
        + \sum_{t \notin \MLEWarmupRounds_t}   R_{t}(S_{t}^\star, \wb^\star) -  R_{t}(S_{t}, \wb^\star). 
        \numberthis \label{eq:proof_MLE_regret}
    \end{align*}
    Moreover, we have
    \begin{align*}
    \sum_{t \notin \MLEWarmupRounds_t}   R_{t}(S_{t}^\star, \wb^\star) -  R_{t}(S_{t}, \wb^\star)
        &\leq \sum_{t \notin \MLEWarmupRounds_t}   \MLERevenue_t(S_{t}) -  R_{t}(S_{t}, \wb^\star)  
        \tag{Lemma~\ref{lemma:optimism}}
        \\
        &\leq \sum_{t \notin \MLEWarmupRounds_t}   \dbtilde{R}^{\text{MLE}}_t(S_{t}) -  R_{t}(S_{t}, \wb^\star)  
        \tag{Eqn.~\eqref{eq:proof_MLE_overly_R}}
        \\
        &= \sum_{t \notin \MLEWarmupRounds_t} \tilde{Q}(\ub_t) - \tilde{Q}(\ub^\star_t) 
        \\
        &=  \underbrace{  \sum_{t \notin \MLEWarmupRounds_t}  \nabla \tilde{Q}(\ub_t^\star)^\top (\ub_t - \ub^\star_t) }_{I_3}
            + \underbrace{\frac{1}{2} \sum_{t \notin \MLEWarmupRounds_t}  (\ub_t - \ub^\star_t)^\top \nabla^2 \tilde{Q}(\bar{\ub}_t) (\ub_t - \ub^\star_t)}_{I_4},
        \numberthis \label{eq:proof_MLE_I3_I4}
    \end{align*}
    where $\bar{\ub}_t = (\bar{u}_{ti_1}, \dots, \bar{u}_{ti_{|S_t|}})^\top \in \RR^{|S_t|}$ is the convex combination of $\ub_t$ and $\ub^\star_t$.

    We first bound the term $I_3$.
    For simplicity, let
    $\EE^{\wb}_{t}[x_{ti}] = \EE_{j \sim p_t(\cdot | S_t, \wb)}[x_{ti}]$,
    and
    $\EE^{\wb}_t[r_{ti}] = \EE_{j \sim p_t(\cdot | S_t, \wb)}[r_{ti}]$.
    Then, we get
    \begin{align*}
        &\sum_{t \notin \MLEWarmupRounds_t}  \nabla \tilde{Q}(\ub_t^\star)^\top (\ub_t - \ub^\star_t)
        \\
        &=  \sum_{t \notin \MLEWarmupRounds_t} \sum_{i \in S_t} \frac{\exp(x_{ti}^\top \wb^\star )r_{ti}}{1 + \sum_{k\in S_t} \exp(x_{tk}^\top \wb^\star ) } (u_{ti} - u_{ti}^\star ) 
        -  \sum_{j \in S_t}   \frac{\exp(x_{tj}^\top \wb^\star ) r_{tj} \sum_{i \in S_t} \exp(x_{ti}^\top \wb^\star )}{(1 + \sum_{k\in S_t} \exp(x_{tk}^\top \wb^\star ))^2 } (u_{ti} - u_{ti}^\star )   
        \\
        &=  \sum_{t \notin \MLEWarmupRounds_t} \sum_{i \in S_t} p_t(i | S_t, \wb^\star)  r_{ti} \left( 2 \MLEConfRadiusEllip_t(\delta) \| x_{ti} \|_{\MLEHessian_t(\nub_t^\star)^{-1}} -  \sum_{j \in S_t} p_t(j | S_t, \wb^\star)  2 \MLEConfRadiusEllip_t(\delta) \| x_{tj} \|_{\MLEHessian_t(\nub_t^\star)^{-1}} \right) 
        \\
        &\leq 2 \MLEConfRadiusEllip_T(\delta)  \sum_{t \notin \MLEWarmupRounds_t}  \sum_{i \in S_t} p_t(i | S_t, \wb^\star)  r_{ti} \left(  \| x_{ti} \|_{\MLEHessian_t(\nub_t^\star)^{-1}} -  \sum_{j \in S_t} p_t(j | S_t, \wb^\star) \| x_{tj} \|_{\MLEHessian_t(\nub_t^\star)^{-1}} \right)
        \\
        &=2 \MLEConfRadiusEllip_T(\delta)  \sum_{t \notin \MLEWarmupRounds_t}
            \EE^{\wb^\star}_t 
            \left[
                \left(
                    r_{ti} - 
                    \EE^{\wb^\star}_t [r_{tj}]  
                \right) 
                \bigg(
                     \| x_{ti} \|_{\MLEHessian_t(\nub_t^\star)^{-1}} - 
                     \EE^{\wb^\star}_t \left[ \| x_{tj} \|_{\MLEHessian_t(\nub_t^\star)^{-1}}  \right] 
                \bigg)
                \right]
        \tag{$x_{t0} = \mathbf{0}$, $r_{t0} = 0$}
        \\
        &\leq 2 \MLEConfRadiusEllip_T(\delta)  \sum_{t \notin \MLEWarmupRounds_t}
            \EE^{\wb^\star}_t 
            \left[
                \left(
                    r_{ti} - 
                    \EE^{\wb^\star}_t [r_{tj}]  
                \right) 
                     \left\| x_{ti} 
                        - \EE^{\wb^\star}_t[x_{tj}]
                     \right\|_{\MLEHessian_t(\nub_t^\star)^{-1}} 
                \right]
        \tag{Similar to Eqn.\eqref{eq:eq:proof_regret_cov_bound}}
    \end{align*}
    We further decompose the last term as follows:
    \begin{align*}
        &  \sum_{t \notin \MLEWarmupRounds_t}
            \EE^{\wb^\star}_t 
            \left[
                \left(
                    r_{ti} - 
                    \EE^{\wb^\star}_t [r_{tj}]  
                \right) 
                     \left\| x_{ti} 
                        - \EE^{\wb^\star}_t[x_{tj}]
                     \right\|_{\MLEHessian_t(\nub_t^\star)^{-1}} 
                \right]
        \\
        &=   \sum_{t \notin \MLEWarmupRounds_t}
        \sum_{i \in S_t} 
        \sqrt{
            p_t (i | S_t, \wb^\star) p_t (i | S_t, \nub^\star_t) 
        }
            \left(
                r_{ti} - 
                \EE^{\wb^\star}_t [r_{tj}]  
            \right) 
                 \left\| x_{ti} 
                    - \EE^{\nub_t^\star}_t[x_{tj}]
                 \right\|_{\MLEHessian_t(\nub_t^\star)^{-1}} 
        \\
        &+  \sum_{t \notin \MLEWarmupRounds_t}
        \sum_{i \in S_t} 
        \left(
        \sqrt{ p_t (i | S_t, \wb^\star)}  - 
             \sqrt{p_t (i | S_t, \nub^\star_t)}
        \right)
        \sqrt{
            p_t (i | S_t, \wb^\star) 
        }
            \left(
                 r_{ti} - 
                    \EE^{\wb^\star}_t [r_{tj}]
            \right) 
                 \left\| x_{ti} 
                    - \EE^{\nub_t^\star}_t[x_{tj}]
                 \right\|_{\MLEHessian_t(\nub_t^\star)^{-1}}
        \\
        &+  \sum_{t \notin \MLEWarmupRounds_t}
        \sum_{i \in S_t} 
            p_t (i | S_t, \wb^\star) 
            \left(
                r_{ti} - 
                \EE^{\wb^\star}_t [r_{tj}]  
            \right) 
            \left(
            \left\| x_{ti} 
                    - \EE^{\wb^\star}_t[x_{tj}]
                 \right\|_{\MLEHessian_t(\nub_t^\star)^{-1}} 
                 - 
                 \left\| x_{ti} 
                    - \EE^{\nub^\star_t}_t[x_{tj}]
                 \right\|_{\MLEHessian_t(\nub_t^\star)^{-1}} 
            \right)
        .
        \numberthis 
        \label{eq:proof_MLE_first_1}
    \end{align*}
    Then, the first term in~\eqref{eq:proof_MLE_first_1} can be bounded as follows:
    \begin{align*}
        \sum_{t \notin \MLEWarmupRounds_t}
        &\sum_{i \in S_t} 
        \sqrt{
            p_t (i | S_t, \wb^\star) p_t (i | S_t, \nub^\star_t) 
        }
            \left(
                r_{ti} - 
                \EE^{\wb^\star}_t [r_{tj}]  
            \right) 
                 \left\| x_{ti} 
                    - \EE^{\nub_t^\star}_t[x_{tj}]
                 \right\|_{\MLEHessian_t(\nub_t^\star)^{-1}} 
        \\
        &\leq \usqrt{
            \sum_{t \notin \MLEWarmupRounds_t}
            \ubrace{
            \sum_{i \in S_t} 
            p_t (i | S_t, \wb^\star)
            \left(
                r_{ti} - 
                \EE^{\wb^\star}_t [r_{tj}]  
            \right)^2
            }{=: \sigma_t^2}
        }
        \sqrt{
            \sum_{t \notin \MLEWarmupRounds_t}
            \sum_{i \in S_t} 
             p_t (i | S_t, \nub^\star_t)
            \left\| x_{ti} 
                    - \EE^{\nub_t^\star}_t[x_{tj}]
                 \right\|_{\MLEHessian_t(\nub_t^\star)^{-1}}^2 
        }
        \tag{Cauchy-Schwarz inequality}
        \\
        &\leq \sqrt{
            \sum_{t=1}^T
            \sigma_t^2
        }
        \sqrt{
            2d \log \left( 1+ \frac{T}{d \MLERegualizer} \right)
        }.
        \tag{Lemma~\ref{lemma:epl_H}}
    \end{align*}
    Note that when applying the elliptical potential lemma (Lemma~\ref{lemma:epl_H}), we ensure that the condition $\| x_{ti} \|_{H_t(\nub^\star_t)^{-1}}^2 \leq \frac{1}{2}$ holds for all $ t \notin \MLEWarmupRounds_0$.
    Additionally, the second term in~\eqref{eq:proof_MLE_first_1} can be bounded as follows:
    \begin{align*}
        &\sum_{t \notin \MLEWarmupRounds_t}
        \sum_{i \in S_t} 
        \left(
        \sqrt{ p_t (i | S_t, \wb^\star)}  - 
             \sqrt{p_t (i | S_t, \nub^\star_t)}
        \right)
        \sqrt{
            p_t (i | S_t, \wb^\star) 
        }
            \left(
                 r_{ti} - 
                    \EE^{\wb^\star}_t [r_{tj}]
            \right) 
                 \left\| x_{ti} 
                    - \EE^{\nub^\star_t}_t[x_{tj}]
                 \right\|_{\MLEHessian_t(\nub_t^\star)^{-1}}
        \\
        &\leq 
        \sum_{t \notin \MLEWarmupRounds_t}
        \sum_{i \in S_t}
        \frac{
            |p_t (i | S_t, \wb^\star) - p_t (i | S_t, \nub^\star_t) |
        }{
            \sqrt{ p_t (i | S_t, \wb^\star)}  + 
             \sqrt{p_t (i | S_t, \nub^\star_t)}
             }
        \sqrt{
            p_t (i | S_t, \wb^\star) 
        }
         \left\| x_{ti} 
                    - \EE^{\nub^\star_t}_t[x_{tj}]
                 \right\|_{\MLEHessian_t(\nub_t^\star)^{-1}}
        \\
        &\leq 
        \sum_{t \notin \MLEWarmupRounds_t}
        \sum_{i \in S_t}
            |p_t (i | S_t, \wb^\star) - p_t (i | S_t, \nub^\star_t) |
         \left\| x_{ti} 
                    - \EE^{\nub^\star_t}_t[x_{tj}]
                 \right\|_{\MLEHessian_t(\nub_t^\star)^{-1}}
        \\
        &\leq 
        4 \MLEConfRadiusEllip_T(\delta) 
        \sum_{t \notin \MLEWarmupRounds_t}
        \max_{i \in S_t} 
         \left\| x_{ti} 
                    - \EE^{\nub^\star_t}_t[x_{tj}]
                 \right\|_{\MLEHessian_t(\nub_t^\star)^{-1}}
        \max_{i \in S_t}  \| x_{ti} \|_{\MLEHessian_t(\nub^\star_t)^{-1}}
        \tag{Lemma~\ref{lemma:gap_p_wb1_wb2}}
        \\
        &\leq 
        4 \MLEConfRadiusEllip_T(\delta)
        \sqrt{
            \sum_{t \notin \MLEWarmupRounds_t}
            \max_{i \in S_t} 
            \left\| x_{ti} 
                    - \EE^{\nub^\star_t}_t[x_{tj}]
                 \right\|_{\MLEHessian_t(\nub_t^\star)^{-1}}^2
        }
        \sqrt{
            \sum_{t \notin \MLEWarmupRounds_t}
            \max_{i \in S_t}  \| x_{ti} \|_{\MLEHessian_t(\nub^\star_t)^{-1}}^2
        }
        \tag{Cauchy-Schwarz inequality}
        \\
        &\leq 
        \frac{4}{\sqrt{\kappa}}
        \MLEConfRadiusEllip_T(\delta)
        \sqrt{
            \sum_{t \notin \MLEWarmupRounds_t}
            \sum_{i \in S_t}
            p_t (i | S_t, \nub^\star_t) 
            \left\| x_{ti} 
                    - \EE^{\nub^\star_t}_t[x_{tj}]
                 \right\|_{\MLEHessian_t(\nub_t^\star)^{-1}}^2
        }
        \sqrt{
            \sum_{t \notin \MLEWarmupRounds_t}
            \max_{i \in S_t}  \| x_{ti} \|_{\MLEHessian_t(\nub^\star_t)^{-1}}^2
        }
        \tag{Definition of $\kappa$}
        \\
        &\leq \frac{8}{\kappa}
        \MLEConfRadiusEllip_T(\delta) 
        d \log \left(1 + \frac{T}{d \MLERegualizer} \right)
        \tag{Lemma~\ref{lemma:epl_H} and~\ref{lemma:elliptical_x_tilde_lee}}
        .
    \end{align*}
    Finally, we bound the last term in~\eqref{eq:proof_MLE_first_1}.
    Using the inequality $\| \ab \| -  \|\bb\| \leq \| \ab-\bb \| $ for any vectors $\ab, \bb \in \RR^d$, we get
    \begin{align*}
        \sum_{t \notin \MLEWarmupRounds_t}
        &\sum_{i \in S_t} 
            p_t (i | S_t, \wb^\star) 
            \left(
                r_{ti} - 
                \EE^{\wb^\star}_t [r_{tj}]  
            \right) 
            \left(
            \left\| x_{ti} 
                    - \EE^{\wb^\star}_t[x_{tj}]
                 \right\|_{\MLEHessian_t(\nub_t^\star)^{-1}} 
                 - 
                 \left\| x_{ti} 
                    - \EE^{\nub^\star_t}_t[x_{tj}]
                 \right\|_{\MLEHessian_t(\nub_t^\star)^{-1}} 
            \right)
        \\
        &\leq \sum_{t \notin \MLEWarmupRounds_t}
        \sum_{i \in S_t} 
            p_t (i | S_t, \wb^\star) 
            \left\| 
                \sum_{j \in S_t}
                (
                    p_t(j | S_t, \nub^\star_t)
                    - p_t(j | S_t, \wb^\star)
                )
                x_{tj}
             \right\|_{\MLEHessian_t(\nub_t^\star)^{-1}} 
        \\
        &\leq 
        4\MLEConfRadiusEllip_T(\delta)
        \sum_{t \notin \MLEWarmupRounds_t}
        \sum_{i \in S_t}
        |
            p_t(i | S_t, \nub^\star_t)
                    - p_t(i | S_t, \wb^\star)
        |
        \|x_{ti}\|_{\MLEHessian_t(\nub_t^\star)^{-1}} 
        \\
        &\leq 4\MLEConfRadiusEllip_T(\delta)
        \sum_{t \notin \MLEWarmupRounds_t} 
        \max_{i \in S_t} \|x_{ti}\|_{\MLEHessian_t(\nub_t^\star)^{-1}}^2
        \tag{Lemma~\ref{lemma:gap_p_wb1_wb2}}
        \\
        &\leq 
        \frac{8}{\kappa}
        \MLEConfRadiusEllip_T(\delta) 
        d \log \left(1 + \frac{T}{d \MLERegualizer} \right)
        \tag{Lemma~\ref{lemma:elliptical_x_tilde_lee}}
        .
    \end{align*}
    By combining the three results above, we can establish a bound for $I_3$.
    \begin{align*}
        \sum_{t \notin \MLEWarmupRounds_t}  \nabla \tilde{Q}(\ub_t^\star)^\top (\ub_t - \ub^\star_t)
        \leq 
        2 \MLEConfRadiusEllip_T(\delta) 
        \sqrt{
            \sum_{t=1}^T
            \sigma_t^2
        }
        \sqrt{
            2d \log \left( 1+ \frac{T}{d \MLERegualizer} \right)
        }
        + \frac{36}{\kappa}
        \MLEConfRadiusEllip_T(\delta)^2 
        d \log \left(1 + \frac{T}{d \MLERegualizer} \right).
        \numberthis \label{eq:proof_MLE_I3_result}
    \end{align*}
    On the other hand, the term $I_4$ in Equation~\eqref{eq:proof_MLE_I3_I4} can be bounded by following the same process as in Equation~\eqref{eq:proof_regret_decompose_second_result} in Appendix~\ref{app_sec:proof_thm_regret_main}.
    \begin{align*}
        \frac{1}{2} \sum_{t \notin \MLEWarmupRounds_t}  (\ub_t - \ub^\star_t)^\top \nabla^2 \tilde{Q}(\bar{\ub}_t) (\ub_t - \ub^\star_t)
        \leq \frac{20}{\kappa} \MLEConfRadiusEllip_T(\delta)^2  d \log \left(1 + \frac{T}{d \MLERegualizer} \right).
        \numberthis \label{eq:proof_MLE_I4_result}
    \end{align*}
    Plugging~\eqref{eq:proof_MLE_I3_result} and~\eqref{eq:proof_MLE_I4_result} into~\eqref{eq:proof_MLE_regret}, and setting $\MLERegualizer = \frac{1}{8B^2}$ and 
    $\MLEConfRadiusEllip_T(\delta) = \BigO (\sqrt{d \log (BT)})$,
    we obtain
    \begin{align*}
        \sum_{t = 1}^T   R_{t}(S_{t}^\star, \wb^\star) -  R_{t}(S_{t}, \wb^\star)
        &\leq 
        \frac{2}{\log ( 3/2)} d
        \log \left(
            1 + 
            \frac{1}{\log(3/2) \MLERegualizer }
        \right)
        \\
        &+ 2 \MLEConfRadiusEllip_T(\delta) 
        \sqrt{
            \sum_{t=1}^T
            \sigma_t^2
        }
        \sqrt{
            2d \log \left( 1+ \frac{T}{d \MLERegualizer} \right)
        }
        + \frac{56}{\kappa}
        \MLEConfRadiusEllip_T(\delta)^2 
        d \log \left(1 + \frac{T}{d \MLERegualizer} \right)
        \\
        &= \BigO \left(
            d \log (B T) 
            \sqrt{ \sum_{t=1}^T \sigma_t^2 
            }
            + \frac{1}{\kappa}  d^2 
            \left(\log (B T)\right)^2
        \right)
        .
    \end{align*}

\end{proof}

\subsection{Proof of Lemmas for Theorem~\ref{thm:MLE}} 
\label{app_subsec:useful_lemmas_thm:MLE}
\subsubsection{Proof of Lemma~\ref{lemma:MLE_CS}}
\label{app_subsubsec:proof_of_lemma:MLE_CS}
\begin{proof} [Proof of Lemma~\ref{lemma:MLE_CS}]
    By using a Taylor expansion and applying the first-order optimality condition for a convex function, we obtain
    \begin{align*}
        \mathcal{L}_t (\wb^\star) - \mathcal{L}_t (\MLEParam_t)
        &= 
        \underbrace{\langle \nabla \mathcal{L}_t  (\MLEParam_t),
            \wb^\star - \MLEParam_t
        \rangle}_{\geq 0, \text{ first order optimality condition}}
        + \left(
            \wb^\star - \MLEParam_t\right)^\top
        \left(
            \int_0^1 (1-v) \nabla^2 \mathcal{L}_t (\MLEParam_t + v (\wb^\star - \MLEParam_t)) \dd v
        \right)
        \left(
            \wb^\star - \MLEParam_t\right)  
        \\
        &\geq \left\|
            \wb^\star - \MLEParam_t
        \right\|_{\int_0^1 (1-v) \nabla^2 \mathcal{L}_t (\MLEParam_t + v (\wb^\star - \MLEParam_t)) \dd v }^2
        \\
        &=  \frac{1}{2}
        \left\|
            \wb^\star - \MLEParam_t
        \right\|_{\nabla^2 \mathcal{L}_t (\nub_t^\star)}^2
        \tag{Definition of $\nub^\star_t$}
        \\
        &\geq \frac{1}{2}
        \left\|
            \wb^\star - \MLEParam_t
        \right\|_{\MLEHessian_t(\nub_t^\star)}^2
        - 4 B^2 \MLERegualizer.
        \tag{Definition of $\MLEHessian_t(\nub_t^\star)$}
    \end{align*}
    By setting $L_t = 2(t-1)$ and $\MLERegualizer = \frac{1}{8B^2}$, and applying Lemma~\ref{lemma:thm3.2_lee2024unified}, we derive
    \begin{align*}
         \left\|
            \wb^\star - \MLEParam_t
        \right\|_{\MLEHessian_t(\nub_t^\star)}^2
        \leq 2 \log \frac{1}{\delta}
                + 2 d \log \left( \max \left\{e,  \frac{4 e B (t-1)}{d}   \right\}  \right)
            + 1
        =  2 \MLEConfRadius_t(\delta)^2  + 1
        ,
    \end{align*}
    which concludes the proof.
\end{proof}

\subsubsection{Proof of Lemma~\ref{lemma:gap_p_wb1_wb2}}
\label{app_subsubsec:proof_of_lemma:gap_p_wb1_wb2}
\begin{proof} [Proof of Lemma~\ref{lemma:gap_p_wb1_wb2}]
    By the mean value theorem, there exists $\xib = (1-c) \wb_1 + c \wb_2$ for some $c \in (0,1)$ such that
    \begin{align*}
        \sum_{i \in S_t} &
        \left|
            p_t(i | S_t, \wb_1) 
            - p_t(i | S_t, \wb_2) 
        \right|
        \omega_{ti}
        \\
        &= 
        \sum_{i \in S_t} 
        \left|
            \nabla p_t(i | S_t, \xib)^\top 
             (\wb_1 - \wb_2)
        \right|
        \omega_{ti}
        \\
        &= \sum_{i \in S_t} 
        \left|
            \left(
            p_t(i | S_t, \xib) x_{ti}
            - p_t(i | S_t, \xib)
            \sum_{j \in S_t}
            p_t(j | S_t, \xib) x_{tj}
            \right)^\top (\wb_1 - \wb_2)
        \right|
        \omega_{ti}
        \\
        &\leq 
        \sum_{i \in S_t} 
        p_t(i | S_t, \xib) 
        \left| 
            x_{ti}^\top (\wb_1 - \wb_2)
        \right|
        \omega_{ti}
        +  \sum_{i \in S_t}
         p_t(i | S_t, \xib)
         \omega_{ti}
        \sum_{j \in S_t}
        p_t(j | S_t, \xib)
        \left| 
            x_{tj}^\top (\wb_1 - \wb_2)
        \right|
        \\
        &\leq 
        2\MLEConfRadiusEllip_t(\delta)
        \sum_{i \in S_t} 
        p_t(i | S_t, \xib) 
            \| x_{ti} \|_{\MLEHessian_t(\nub^\star_t)^{-1}}
        \omega_{ti}
        + 
        2\MLEConfRadiusEllip_t(\delta)
        \sum_{i \in S_t}
         p_t(i | S_t, \xib)
         \omega_{ti}
        \sum_{j \in S_t}
        p_t(j | S_t, \xib)
            \|x_{tj}\|_{\MLEHessian_t(\nub^\star_t)^{-1}} 
        \tag{$\wb_1, \wb_2 \in \MLEConfidenceSet_{t}(\delta)$, Lemma~\ref{lemma:MLE_CS}}
        \\
        &\leq 4\MLEConfRadiusEllip_t(\delta) 
        \max_{i \in S_t} \omega_{ti}
        \max_{i \in S_t} \| x_{ti} \|_{\MLEHessian_t(\nub^\star_t)^{-1}},
    \end{align*}
    which concludes the proof.
\end{proof}
\section{Experiment Details and Additional Results}
\label{app_sec:experimat_details}
%
\begin{figure*}[h!]
    \centering
        \includegraphics[clip, trim=0cm 0.0cm 0cm 0.0cm, width=\textwidth]{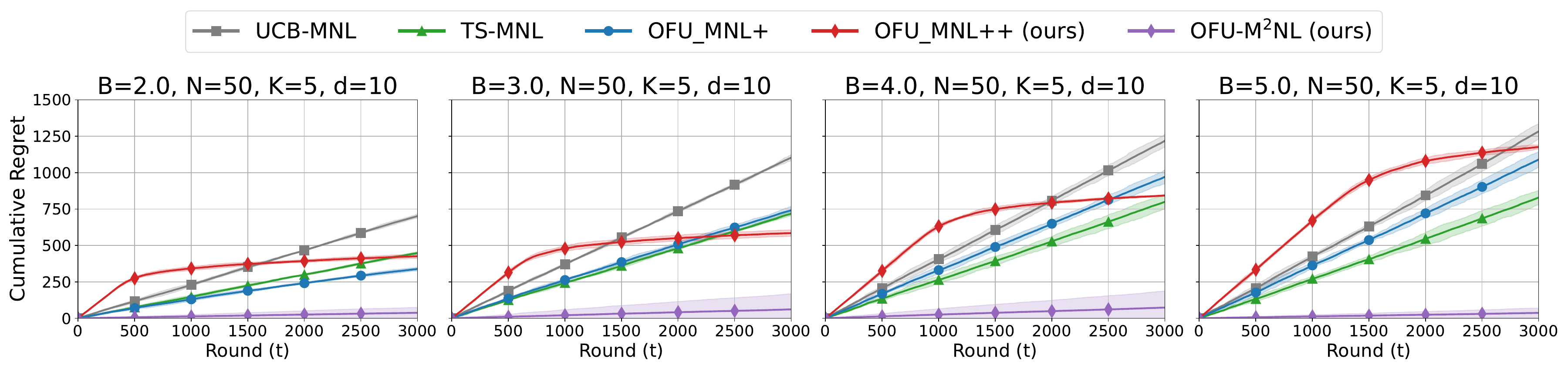}
    \caption{Cumulative regret for different values of $B$ when $d=10$.
    } 
    \label{fig:experiment_d=10}
\end{figure*}
%
For each instance, we sample the true parameter $\wb^\star$ uniformly from a $d$-dimensional Euclidean ball of radius $B$, denoted by $\BB^d(B)$.
Similarly, each context feature $x_{ti}$ is independently and identically distributed (i.i.d.) from a unit ball, denoted as $\BB^d(1)$.
This ensures that $\| \wb^\star \|_2 \leq B$ and $\| x_{ti} \|_2 \leq 1$, satisfying Assumption~\ref{assum:bounded_assumption}.
The rewards are sampled independently in each round from a uniform distribution, i.e., $r_{ti} \sim \operatorname{Unif}(0,1)$. We set the number of items to $N=50$ and the maximum assortment size to $K=5$.
For each instance, we conducted 20 independent runs and reported the average cumulative regret (Figures~\ref{fig:experiment} and~\ref{fig:experiment_d=10}) as well as the average runtime per round (Figure~\ref{fig:experiment_runtime}) for each algorithm.
In our experiments, since the threshold $\Threshold_t$ is too conservative in practice, we empirically tuned the hyperparameter $\Threshold_t$ for~\AlgName{} by searching over a certain range of values while maintaining its inverse relationship with $\alpha$ (i.e., a higher $\Threshold_t$ corresponds to a lower $\alpha$).

As an additional experiment, Figure~\ref{fig:experiment_d=10} presents results for a larger value of $d$, specifically $d=10$. Our algorithms continue to outperform other baselines.
While the performance of~\AlgName{} is somewhat sensitive to the values of $B$ and $d$, primarily due to the adaptive warm-up rounds, its asymptotic performance appears to be the best. Notably, the slope of the regret curve is the smallest for large $t$. Additionally,~\AlgName{} enjoys a constant computational cost, similar to \texttt{OFU-MNL+}.
In contrast, \AlgNameMLE{} is the slowest among the algorithms, as it requires solving a convex optimization problem to compute the optimistic parameter $\MLEUCBParam_{ti}$, as described in Equation~\eqref{eq:MLE_optimization}.

\end{document}